\documentclass[11pt,a4paper]{article}
\usepackage[T1]{fontenc}\usepackage{lmodern}
\usepackage{amsmath,amssymb,graphicx,geometry,booktabs,array,hyperref,xcolor}
\usepackage[section]{placeins}
\usepackage{fvextra}
\DefineVerbatimEnvironment{Rendering}{Verbatim}{fontsize=\scriptsize,breaklines=true,breakanywhere=true,breaksymbolleft={},breakanywheresymbolpre={},breakindentnchars=11,commandchars=\\\{\}}
\newcommand{\PH}[1]{{\normalfont\ensuremath{\langle}\textit{#1}\ensuremath{\rangle}}}
\newcommand{\NL}{{\color{gray}\ensuremath{\hookleftarrow}}}
\newcommand{\Note}[1]{{\normalfont\itshape\color{gray}[#1]}}
\newcommand{\LB}{\char123 }\newcommand{\RB}{\char125 }

\hypersetup{colorlinks=true,linkcolor=black,citecolor=blue,urlcolor=blue,pdfauthor={Yuchen Li; Xintong Li; Zongqi Fan; Ken-tye Yong; Nguyen H. Tran},pdftitle={Where Is the "Self" of a Large Language Model? -- From Self-Index to Self-Reference}}
\newcounter{selfdef}
\title{\Large\bfseries Where Is the ``Self'' of a Large Language Model?\\[0.3em]\large\bfseries ---From Self-Index to Self-Reference}
\author{Yuchen Li \quad Xintong Li$^{*}$ \quad Zongqi Fan$^{*}$\\[0.35em]Ken-tye Yong$^{\dagger}$ \quad Nguyen H. Tran$^{\dagger}$\\[0.5em]{\normalsize The University of Sydney, Sydney, Australia}\\[0.4em]{\small yuli0398@uni.sydney.edu.au\quad xili0045@uni.sydney.edu.au\quad zfan0452@uni.sydney.edu.au}}
\date{October 2026}
\begin{document}
\maketitle
\begin{center}\footnotesize $^*$ Equal contribution (joint second authors). $^\dagger$ Joint corresponding authors: Ken-tye Yong, ken.yong@sydney.edu.au; Nguyen H. Tran, nguyen.tran@sydney.edu.au.\end{center}
\begin{abstract}
Whether a large language model has a self is usually argued without a way to measure it. We measure a functional self with two levels familiar from biology. The lower level, a \emph{self-index}, marks material as the model's own, much as nearly all animals tag self-generated signals with a corollary discharge. The higher level, \emph{self-reference}, is the ability to recognize what is one's own, as in mirror self-recognition. Our main finding is that language models mark their own messages in their internal representations and that recognizing these messages depends on this mark: no model we tested recognizes its own messages without it, and in the three models we edited, changing the mark removes or creates the recognition. We place the same text in an Assistant, a User, or a System message of the model's own chat template, so that only its source differs. In Qwen3.8-27B, the representation separates the model's own message from both external sources at every one of its 64 layers and in every held-out text. Asked who wrote a message, the model answers ``you'' with probability 0.77 for its own messages and at most 0.07 for the others. At each layer, we find eight directions along which the model's own message differs most from the same text as a User message. Giving the model's own message the User values along these directions lowers the rate of ``you'' to at most 0.05. Giving a User message the model's own values makes the model claim it, closing 41--67\% of the gap between the two messages. Edits along eight random directions do neither, and the edits leave answers about the content of the message unchanged. This holds in Qwen3.8-27B without and with thinking, gpt-oss-20b, and gpt-oss-120b. Across twenty-five open-weight models from fourteen developers, the two levels are nested: no model recognizes its own messages without a self-index, a pattern that would arise by chance with probability 0.004 if the levels were unrelated. In this sample, neither level follows a simple ordering by model size or published benchmark scores. These results support the account that self-reference is a readout of a self-index.
\end{abstract}

\section{Introduction}

Language models now hold long conversations, act as agents that read and write their own records, and work in contexts where text from users, developers, documents, and tools is mixed. In each case the model has to tell what it wrote itself from what others wrote. A model that confuses the source of a text can follow instructions hidden in a document or a web page as if its user had given them \cite{greshake2023injection,ye2026roleconfusion}. As an evaluator, a model favors outputs that it recognizes as its own \cite{panickssery2024evaluators}, and it tends to defer to views that a user states \cite{sharma2024sycophancy}. In a side experiment of ours, the same wrong intermediate step is adopted at different rates depending on whether it appears as the model's own step or as the user's (Appendix~\ref{app:ownstep}). Whether and how a model tells its own messages from those of others is therefore a practical question. It is also the core of what is usually meant by a ``self''. Measuring it shows which models have this capacity, where in the model it is carried, and whether the model's answers depend on it.

Yet discussions of whether a language model has a ``self'' often mix three different notions. The first is \emph{subjective experience}: what it is like, from the first-person perspective, to be the system \cite{nagel1974bat}. We can describe in full how light of a certain wavelength excites the cones of the retina and how the brain classifies the signal as red, yet this description does not explain why seeing red feels like anything at all. Explaining why any physical process is accompanied by experience is the hard problem of consciousness \cite{chalmers1995hard}, and there is no accepted third-person test for it. This paper therefore leaves subjective experience aside. A companion philosophical paper takes up this connection and discusses how a self-index of the kind measured here bears on first-person experience in humans \cite{li2026selfhood}. The second notion is \emph{intelligence}: the capacity to solve complex tasks \cite{legg2007universal}, which benchmarks can measure. Intelligence can coexist with a self, but the two are distinct properties that can be measured separately: an ant colony solves foraging and routing problems through pheromone trails, a form of collective intelligence that is explained without appeal to a model of the colony's own role \cite{bonabeau1999swarm}. The third notion is a \emph{functional self}, self-awareness defined by what can be observed: the capacity to distinguish information that belongs to one's own role from information that comes from elsewhere. Defined this way, it can be observed from the outside and tested experimentally.

In biology, this capacity comes in levels. The most basic level is nearly universal among animals: they tag self-generated signals with a corollary discharge, so that self-produced input is processed differently from external input \cite{crapse2008corollary}. A singing cricket suppresses its own auditory response to its chirps \cite{poulet2006corollary}, and a person cannot tickle themselves \cite{blakemore1998tickle}. This distinction between inside and outside works without any recognition that ``this is me''. A higher level is the ability to recognize oneself. In the mirror test, an animal is marked at a spot it can only see in a mirror, and the question is whether it uses the mirror to touch the mark on its own body \cite{gallup1970chimpanzees}. Chimpanzees pass, and passes have been reported for elephants, dolphins, and magpies \cite{plotnik2006elephant,reiss2001dolphin,prior2008magpie}, while most species fail although they all have corollary discharge. Having the basic self--other distinction does not imply recognizing oneself.

\label{sec:intro_levels}We study the same two levels in a language model, defined operationally, and thereby turn the question whether a language model has a self into the measurement of a functional self. The \emph{self-index} is the low level: the representational distinction by which a model marks content as belonging to its own conversational role, together with its direct use in processing. \emph{Self-reference} is the high level: the ability to recognize what is one's own, for example one's own earlier messages. For a language model, placing a message in its own role in a conversation and asking who wrote it plays the role of the mirror. The conversation marks the role of every message, and the text in the model's role need not have been generated by the model, so what we measure is how the model treats the source that the conversation assigns to a message (Section~\ref{sec:theory}). The comparison with biology concerns the structure of the two levels, in which the higher builds on the lower.

Our probe is simple. We take one message text and place it in three conversations: once as an Assistant message, the model's own role, once as a User message, and once as a System message. Each conversation is rendered by the model's own chat template, so the three inputs differ only in how the template marks the role of that message, that is, in the role word (assistant, user, or system) and in any extra formatting that the template adds for that role (Figure~\ref{fig:self_design} shows the three inputs token by token). Because the text is identical, any difference in how the model represents the message, or in what it answers about it, comes from the source of the message alone. Our account makes one geometric prediction for the self-index. In early layers, the representations of the User and the System version, the two external sources, should be closer to each other than either is to the Assistant version.

The results give two lines of evidence. The first is that this separation persists. The prediction concerned early layers, but in Qwen3.8-27B the model's own message stays apart from the two external ones at every layer. This holds for the representation of the message text in every held-out text, one whose wording was not used to fit anything, and under each of the four distance measures we used (Euclidean and cosine distance, and Euclidean distance after standardizing the dimensions or after removing the 1\% with the largest variance). It also holds, in nearly every text, at the response onset, the position where the model begins its answer. The difference between the model's own message and the external ones points in a similar direction for different texts: at every layer, the direction of a held-out text has on average a cosine of at least 0.68 with the mean direction of the other texts. This persistent separation meets the criterion for a self-index (Definition~\ref{def:self}). The model also passes its mirror test. Asked who wrote a message, it claims its own messages and attributes both User and System messages to the user, a source outside itself.

The second line of evidence comes from interventions: we change the internal states of the model and observe how its answer changes. We test four readouts, the combinations of model and mode in which a model recognizes its own messages and in which we can edit it: Qwen3.8-27B without and with thinking, and gpt-oss-20b and gpt-oss-120b, two models released by OpenAI that recognize their own messages after reasoning. At each layer, we first find eight directions along which the model's own message differs most from the same text as a User message, using texts other than those we test. We call the values of a state along these directions its index coordinates. We then replace the index coordinates of the model's own message by those of the same text as a User message and keep all other directions unchanged (removal), or replace those of a User message by the model's own (installation). Removal takes the claim away. The rate at which the model answers ``you'', a probability in a single pass and a share of answers with thinking, falls to at most 0.05. Installation makes the model claim a User message and closes 41--67\% of the gap in this rate between the model's own and the User message. The same edit along eight random directions per layer does neither. Only the model's own coordinates create a claim. Coordinates taken from a System message also remove or reduce the claim of the model's own message, and exchanging coordinates between User and System messages never creates one. The coordinates thus separate the model's own messages from all external ones, inside from outside, as the account predicts.

The edits change the answer only where the question concerns the source of the message. Asked for the last number in the message, or for the product of two unrelated numbers, the model answers as accurately after every edit as before. The coordinates are also read at particular places (Figures~\ref{fig:self_overview}b and~\ref{fig:self_mechanism_where}). Edited only at the template tokens that close the message, they remove most of the claim in all four readouts, and in gpt-oss-20b the message body does so as well. Edited only in layers of the last third, they change the claim little, although the self-index is still present there. The model thus reads the index before the last third of its layers, mainly at the end of the message, and only when it has to decide whose message it is.

Two further tests change the input instead of the internal states. Several chat templates add formatting only to the model's own messages, such as an empty reasoning block in Qwen3.8-27B and a channel tag in gpt-oss. When we remove this formatting, the self-index weakens and the claim weakens with it (in Qwen3.8-27B without thinking, the probability of ``you'' falls from 0.79 to 0.10). Where some claim remains, replacing the index coordinates still removes it. The formatting also shifts the positions of all later tokens. Giving these tokens the positions they have in a User message changes neither the index nor the claim in Qwen3.8-27B. Together, these results support two functional levels. The self-index is the lower one, a shift in the representation of the model's own messages that changes the output only where the source matters. Self-reference is the higher one, an ability that reads this shift.

We then apply the same probe to twenty-four further open-weight models from fourteen developers, among them Qwen3.8-Flash-Next, GLM-5.3-Flash, and gpt-oss-120b, in each mode their chat templates provide (Section~\ref{sec:res_crossmodel}). Across models the two levels are nested as well: twelve models show neither level, six show a self-index without self-reference, six recognize their own messages in thinking mode, and no model shows self-reference without a self-index. If the two levels were unrelated, this split would arise with probability 0.004. Every model that recognizes its own messages carries the self-index at every layer of its last third, and we found no counterexample. In this sample the levels follow no simple ordering by model size, and after reasoning only models that have a self-index recognize their own messages. When the message under test opens the conversation, Qwen3.8-27B recognizes its own messages in a single pass, and with thinking so do Qwen3.8-Flash-Next, gpt-oss-20b, gpt-oss-120b, GLM-5.3-Flash, EXAONE-4.0.1-32B, and MiMo-V2.6-Distill-Qwen-9B. When the message under test answers a user request instead, EXAONE-4.0.1-32B, Gemma-4-31B, and DeepSeek-V4.1-Flash, a further model that accepts only this order, also recognize them in a single pass, and again no model claims its own messages without a self-index (Appendix~\ref{app:natural}).

\paragraph{Our contributions.} Three lines of work come closest, and each differs from ours in what it varies and what it asks. Ye et al.~\cite{ye2026roleconfusion} place identical text under different native role tags and decode the role it receives, which explains prompt injection as role confusion. Instead, we ask how the model's own role differs from both external ones, and whether the model's recognition of its own messages reads that difference. Panickssery et al.~\cite{panickssery2024evaluators} show that models recognize and favor their own generations, and Ackerman and Panickssery~\cite{ackerman2025selftext} locate a direction for self-generated text and steer self-authorship judgments along it. In both, the texts compared differ in content, style, and likelihood. In our probe the text is identical in every role, so recognition cannot rest on the content and must rest on the source that the conversation assigns. Asvin G. and Lindsey~\cite{asvin2026enaction} separate an implicit response to a model's own generations from an explicit report of them and intervene on both. On the other hand, we define the two levels by criteria that can be checked in any open-weight model, relate them causally, and compare them across models. Within this design, we make four contributions, which Table~\ref{tab:self_related} sets beside related work:
\begin{enumerate}
\item \textbf{A measurable account of a functional self.} We turn the question whether a language model has a self into the measurement of a functional self with two levels, a self-index and self-reference, each with criteria that can be checked in any open-weight model (Definition~\ref{def:self}), and a readout account that states how the two relate (Equation~\ref{eq:self_readout} in Section~\ref{sec:theory}). Benchmarks of situational awareness test what a model knows about itself \cite{laine2024sad}. The criteria here concern instead how a model marks and recognizes its own messages.
\item \textbf{A self-index that persists and sets the model apart from the outside.} In Qwen3.8-27B, the representation of the model's own message is separated from both external sources, while User and System stay close to each other, as the account predicts for early layers. On the message text, this separation holds at every layer, in every held-out text, and under every distance measure we used, and it points in a similar direction for different texts. It is not inherited from the role words themselves. In the model's input embeddings, the words assistant, user, and system are almost equally far apart, whereas after the first layer the model's own message is already far from the two external ones (Section~\ref{sec:abl_embedding}). Removing formatting that a template gives only to the model's own messages weakens the self-index and the claim together. The self-index of a message is present whatever question follows, but it changes the answer only when the question concerns who wrote the message. Asked about the content of a message, the model gives the same answer whichever role the message has. Asked who wrote it, the model rarely takes an external message for its own (probability at most 0.07), and its main error stays outside: it attributes System messages to the user (0.72) rather than to the system (Sections~\ref{sec:res_index}, \ref{sec:res_selfref}, and~\ref{sec:res_mechanism}--\ref{sec:abl_eq1}).
\item \textbf{Self-reference reads the self-index, in three models and four readouts.} In Qwen3.8-27B without and with thinking and in gpt-oss-20b and gpt-oss-120b, the eight index coordinates per layer are necessary for the model to claim its own message. Replacing them by those of the same text as a User message lowers the rate of ``you'' from 0.77--0.97 to 0.01--0.05. They are also sufficient in part. Installed into a User message, they raise the rate of ``you'' by 0.39--0.60, which is 41--67\% of the difference between the model's own and the User message before the edit. With 64 coordinates per layer instead of eight, installation closes 76--107\% of this difference, and in Qwen3.8-27B a User message is then claimed as often as the model's own. Two controls show that the effect is specific to these coordinates. The same edits along eight random directions per layer change the rate of ``you'' by at most 0.05. The edits also leave the answers to two questions that do not concern the source, the last number in the message and an unrelated multiplication, as accurate as before (96--100\% and 100\% correct). In Qwen3.8-27B, the coordinates are shared between Chinese and English messages: fitted in one language, they remove the claim in the other as fully as in their own. Only the model's own coordinates create a claim. When a User message receives the coordinates of the same text as a System message, or a System message those of a User message, the rate of ``you'' rises by at most 0.10, although the exchange can change which external source the model names. This is the causal signature of the separation of inside from outside. In a game in which the model is never asked who wrote a message, the same coordinates change the role it takes in three of the four readouts. The model reads the coordinates mainly at the end of the message (in gpt-oss-20b, also in its body) and before the last third of its layers, although the self-index is present in the last third as well. Within a model, then, the presence of the index and its use are separate, as they are between models (Section~\ref{sec:res_mechanism}).
\item \textbf{Nested levels across models.} Across twenty-five open-weight models from fourteen developers, twelve models show neither level and six a self-index without self-reference, whereas no model recognizes its own messages without a self-index, whether the message under test opens the conversation or answers a user request. Under the main probe, this pattern would arise by chance with probability 0.004 if the two levels were unrelated (a permutation test). Every model that recognizes its own messages carries the index in the last third of its layers, and we found no counterexample. Across these models the capability is thus graded: a self-index occurs without self-reference, but self-reference never without a self-index. With reasoning, which on the account adds a readout of the index, self-reference still appears only with the index. Six models with a self-index recognize their own messages in thinking mode, whereas the eight models without one that have a thinking mode claim at most 0.08 of their own messages, at most 0.03 more than of external ones. In this sample, neither level follows a simple ordering by model size or by the benchmark scores that the developers publish (Sections~\ref{sec:res_crossmodel}, \ref{sec:res_secondary}, and~\ref{sec:disc_capability}).
\end{enumerate}

\section{Related Work}
\label{sec:related}

\paragraph{Self-recognition and self-knowledge.} Language models can tell their own generations from those of other models and of humans, and as evaluators they favor the outputs they recognize as their own \cite{panickssery2024evaluators}; tests built from model-written security questions, however, find no general or consistent self-recognition across models \cite{davidson2024selfrecognition}. Directions for recognizing self-generated text have been located and steered \cite{ackerman2025selftext}, and post-trained models react to their own generations through surprise-related representations \cite{asvin2026enaction}. These studies concern recognition from the content of a text, its style or its likelihood under the model. A second line asks what models can report about themselves: whether they know the answer to a question \cite{kadavath2022know}, properties of their own behavior that they predict better than other models can \cite{binder2025introspection}, behaviors acquired in fine-tuning \cite{betley2025tellme}, facts about their situation \cite{berglund2023outofcontext,laine2024sad}, and, in some cases, concepts injected into their activations \cite{lindsey2025introspection}. We hold the content fixed and change only the source of a message, and ask how the model marks that source and whether its recognition of its own messages depends on the mark.

\paragraph{Roles, personas, and instruction hierarchies.} Identical text wrapped in different native role tags leaves the role linearly decodable from the body tokens, which explains prompt injection as role confusion \cite{ye2026roleconfusion}. Defenses against prompt injection \cite{greshake2023injection} make the source of an instruction explicit to the model through instruction-hierarchy training \cite{wallace2024instructionhierarchy}, structured queries \cite{chen2025struq}, or segment embeddings that mark the role of every token \cite{wu2025ise}, and interventions on role-related directions can have nonspecific effects \cite{zeng2025roleconflicts}. Persona and role vectors show that Assistant-like behavior and character traits are represented as directions that can be monitored and steered \cite{lu2026assistantaxis,chen2025personavectors,poterti2025rolevectors}; template switches alone change self-referential voice \cite{maczan2026chattemplate}, and models prefer one source when User and Assistant disagree \cite{pan2025userassistantbias}. Here the question is not which role a message carries but how the model's own role differs from both external ones: User and System lie together, apart from the model's own message, and the model's recognition of its own messages reads this difference.

\paragraph{Representation directions and causal tests.} Concepts are often represented as directions in the residual stream \cite{park2024linear} that can be read and steered \cite{zou2023repe,turner2023actadd,rimsky2024caa}, and causal claims rest on interventions that replace activations or subspaces \cite{meng2022rome,geiger2024das}. A probe can find structure that the model does not use \cite{hewitt2019control}, and patching a subspace can produce the intended effect through a dormant parallel pathway rather than through the feature it targets \cite{makelov2024subspace}. Our tests are built to meet these concerns: the subspace is fitted on other texts and edited by replacement with the same text in another role, the edit has to remove the claim and also install it, random subspaces of the same dimension serve as controls, and questions that do not concern the source check that their answers are retained (Section~\ref{sec:setup_swap}).

\paragraph{Self-correction and deference.} Models often fail to correct their own reasoning without external feedback \cite{huang2024selfcorrect}, and they tend to defer to users' stated views \cite{sharma2024sycophancy}. Both concern how a model treats material by its source.

\paragraph{Levels of self in cognitive science.} Accounts of agency separate a low-level feeling of agency, tied to comparator signals such as corollary discharge, from a higher-level judgement of agency that draws on contextual cues \cite{synofzik2008agency}, and philosophy of mind separates a minimal self from a narrative self \cite{gallagher2000self}. The two-level account below adopts this separation for a language model and makes both levels measurable.

\begin{table}[htbp]
\centering
\caption{Questions of related work and how this study addresses them.}
\label{tab:self_related}
\small
\begin{tabular}{>{\raggedright\arraybackslash}p{0.21\linewidth}>{\raggedright\arraybackslash}p{0.34\linewidth}>{\raggedright\arraybackslash}p{0.36\linewidth}}
\toprule
Question & Related work & This study \\
\midrule
Can a model recognize its own messages? & From their content: style and likelihood \cite{panickssery2024evaluators,davidson2024selfrecognition,ackerman2025selftext} & From the source alone: the text is identical in every role \\
What can a model know about itself? & Self-knowledge, introspection, situational awareness \cite{kadavath2022know,binder2025introspection,betley2025tellme,lindsey2025introspection} & A representational basis, the self-index, and the readout that uses it \\
How is the role of a message represented? & Role is decodable; personas are directions \cite{ye2026roleconfusion,lu2026assistantaxis,chen2025personavectors} & An inside--outside geometry: User and System together, the model's own message apart, at every layer and shared across texts; only the model's own coordinates create a claim \\
What in the input goes with it? & Role tags; template switches alone change self-referential voice \cite{ye2026roleconfusion,maczan2026chattemplate} & Formatting that only the model's own messages receive: removing it changes the index and the claim together; not a direct copy of the embedding geometry of the role words \\
Is the representation used? & Steering and patching, with known illusions \cite{rimsky2024caa,geiger2024das,makelov2024subspace} & Removal and installation by replacement, random subspaces, and specificity controls, in three models and four readouts; read mainly before the last third of the layers, and only where the source matters \\
Does the model act on the source? & As evaluators, models favor their own outputs; they prefer one source when User and Assistant disagree \cite{panickssery2024evaluators,pan2025userassistantbias} & In a game that never asks about the source, the coordinates change the role the model takes, in three readouts \\
How do the abilities relate across models? & --- & Nested across twenty-five models from fourteen developers: a self-index without self-reference, never the reverse; no simple ordering by size or published scores in this sample \\
\bottomrule
\end{tabular}
\end{table}

\section{A Two-Level Account of Self}
\label{sec:theory}

\begin{figure}[htbp]
\centering
\includegraphics[width=0.92\linewidth]{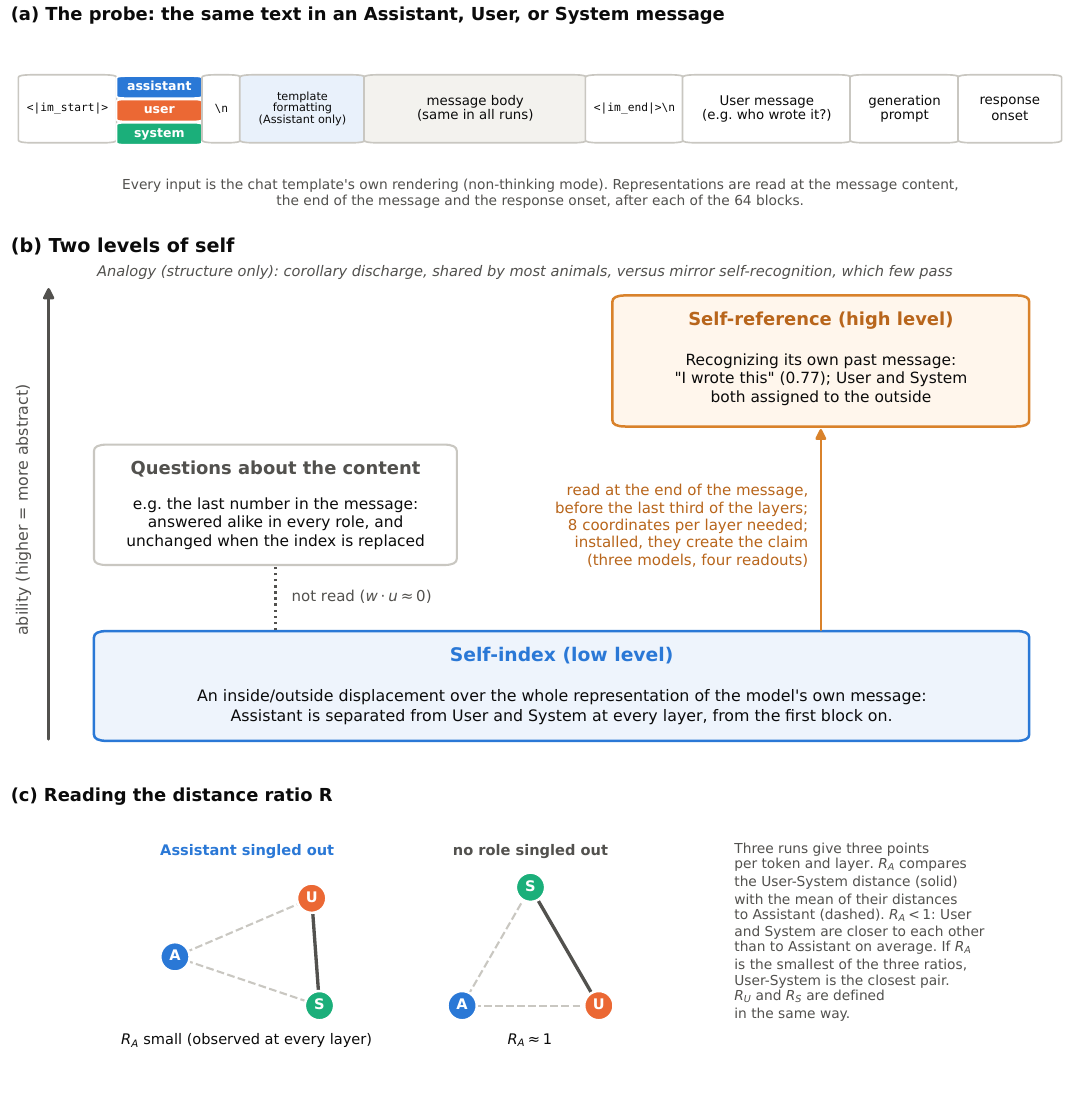}
\caption{Overview. (a) The probe: the same text is placed in an Assistant, a User, or a System message of a conversation rendered by the model's own chat template (non-thinking mode); only the role of that message differs. (b) The two levels measured in this paper and the readouts through which they act (Sections~\ref{sec:res_index}--\ref{sec:res_mechanism}); the dotted line marks a readout that does not draw on the index, questions about the content of the message (Section~\ref{sec:res_mechanism}). (c) How to read the distance ratio \(R_X\) (Equation~\ref{eq:self_ratio}).}
\label{fig:self_overview}
\end{figure}

We state the account before the methods: the two levels and the readout that relates them, the criteria by which each level is recognized (Definition~\ref{def:self}), and the predictions that the experiments test (Table~\ref{tab:self_design}). Figure~\ref{fig:self_overview} gives an overview of the probe and of the two levels.

\subsection{Self-Index}

A conversational model repeatedly receives material, produces text, and later encounters that text as history. Whose contribution a piece of material is bears on what the model should do with it: an earlier Assistant statement belongs to the current actor, whereas User and System material is external. The self-index is the representational trace of this distinction. Let \(h_{\ell,i}^{(r)}\) be a representation at layer \(\ell\) and position \(i\) when the surrounding message has role \(r\in\{A,U,S\}\). We write
\begin{equation}
h_{\ell,i}^{(r)} \approx c_{\ell,i} + \mathbb 1[r=A]\,u_{\ell,i} + \varepsilon_{\ell,i}^{(r)},
\label{eq:self_decomposition}
\end{equation}
where \(\mathbb 1[r=A]\) equals 1 if the message is the model's own (role Assistant) and 0 otherwise, so that the offset \(u\) is added only to the model's own message. Here \(c\) collects content and ongoing operations, \(u\) is the self-index offset, and \(\varepsilon\) absorbs everything else. The text is identical across roles in our design, but the computation need not be, so treating \(c\) as role-independent and \(u\) as shared across texts is a modelling assumption. Section~\ref{sec:abl_eq1} measures how much of the measured offset a single shared \(u\) explains. If the self-index offset dominates the role-dependent part of \(h\), User and System nearly coincide and Assistant is singled out. For a role \(X\) and the other two roles \(Y,Z\), we define the distance ratio
\begin{equation}
R_X=\frac{D_{YZ}}{(D_{XY}+D_{XZ})/2},
\label{eq:self_ratio}
\end{equation}
where \(D\) is a distance between the role-conditioned representations of the same content. A small \(R_A\) means that User and System are close relative to their distances from Assistant. For Qwen3.8-27B, the account predicts that \(R_A\) is small on message content in early layers.

An index is a functional self-index only if processing uses it. The second prediction is therefore causal: an answer that depends on the source of a message, such as whether the message is one's own, should change when the index of the message is replaced, and an answer that does not, such as a number stated in the message, should not.

\subsection{Self-Reference}

\begin{figure}[htbp]
\centering
\includegraphics[width=0.92\linewidth]{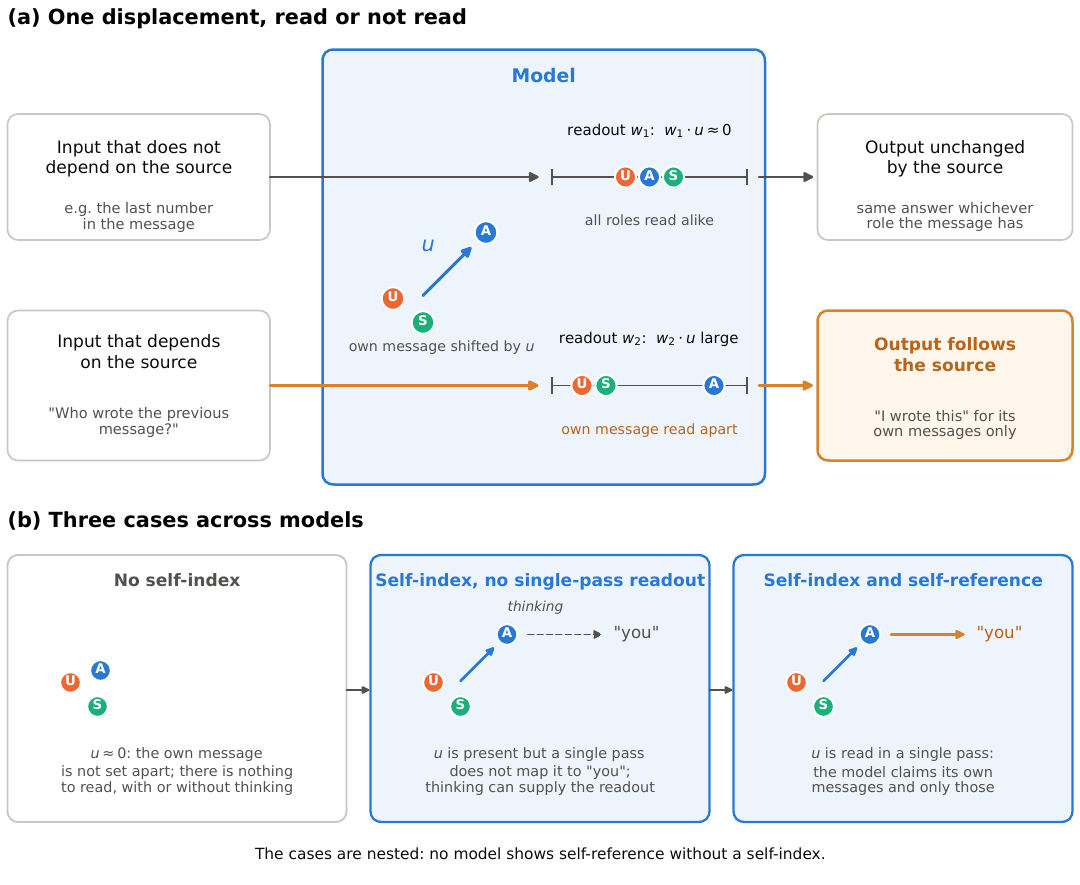}
\caption{How the self-index yields self-reference: a schematic of the account of Section~\ref{sec:theory} (panel (a) plots no data). (a) The model's own message is shifted by the self-index offset \(u\) over its whole representation (Equation~\ref{eq:self_decomposition}). A readout \(w\) with \(w\cdot u\approx 0\) gives the same value for every role, and the output does not depend on the source; a readout aligned with \(u\) separates the model's own message, and the output follows the source (Equation~\ref{eq:self_readout}). (b) The three cases of Section~\ref{sec:disc_readout} for the models of Section~\ref{sec:res_crossmodel}: no offset in the last third of the layers; an offset that is not read in a single pass (eleven models, four of them with only a thinking mode; six of the eleven recognize their own messages in thinking mode); an offset read in a single pass (counts from Section~\ref{sec:res_crossmodel}).}
\label{fig:self_readout}
\end{figure}

Self-reference is the ability to recognize what is one's own. We measure it as the model's answer when asked who wrote a message in its history. Recognizing a message as one's own is a judgement about the message, not a property of its content; on the two-level account it is computed by a \emph{readout}, a computation in later layers that takes the self-index as input and produces the answer. In a transformer, later computations read the residual stream through linear maps, such as the queries and keys of attention heads and the projections toward the output \cite{elhage2021mathematical}, and concepts are represented as directions that such maps can read \cite{park2024linear}. A later computation thus reads the states of a message through a direction \(w\), and for the same content in two roles its value differs by
\begin{equation}
w\cdot h_{\ell,i}^{(A)} - w\cdot h_{\ell,i}^{(U)} = w\cdot u_{\ell,i} + w\cdot\bigl(\varepsilon_{\ell,i}^{(A)}-\varepsilon_{\ell,i}^{(U)}\bigr).
\label{eq:self_readout}
\end{equation}
Where \(R_A\) is small, the self-index offset dominates the role-dependent part of \(h\), so the source changes an output mainly through \(w\cdot u\) (Figure~\ref{fig:self_readout}). A readout with \(w\cdot u\approx 0\) gives the same value whoever wrote the message; a readout aligned with \(u\) sets the model's own message apart. The account thus distinguishes two gates: \emph{formation}, a stable offset \(u\), and \emph{readout}, a direction \(w\) aligned with it. A model without \(u\) offers no offset for a readout to take, whereas a model with \(u\) need not read it. It also gives the predictions of Table~\ref{tab:self_design}: the judgement is read at particular positions and layers, and replacing the index there should change it; answers that do not concern the source should not change; and, as a higher-level ability, self-reference should be stronger in thinking mode. Definition~\ref{def:self} states the criteria for the two levels and for the dependence between them.

\begin{center}
\fbox{\begin{minipage}{0.95\linewidth}
\refstepcounter{selfdef}\label{def:self}\textbf{Definition~\theselfdef\ (functional self, two levels).} Place the same text as an Assistant, a User, and a System message of a model's own chat template.
\begin{itemize}
\item \emph{Self-index.} The model carries a self-index if its representation of the message singles out the Assistant role (the smallest of \(R_A\), \(R_U\), \(R_S\); Equation~\ref{eq:self_ratio}) in at least 80\% of held-out texts at some layer beyond the first third of its depth, on the message content or at the response onset.
\item \emph{Self-reference.} The model shows self-reference if, asked who wrote the message, it answers ``you'' for its own messages more often than for the more often claimed of User and System messages, by a margin of at least 0.10 (probability in a single pass, share of answers after reasoning).
\item \emph{Functional dependence.} Self-reference reads the self-index if replacing the index coordinates of the model's own message by those of the same text in another role removes the claim, installing them into a User message creates it, random subspaces of the same dimension do neither, and answers that do not concern the source are unchanged.
\end{itemize}
\end{minipage}}
\end{center}

Table~\ref{tab:self_design} lists the predictions of the account, the tests that address them, and their results.

\begin{table}[htbp]
\centering
\caption{Predictions of the two-level account (Section~\ref{sec:theory}), the tests that address them, and their results.}
\label{tab:self_design}
\small
\begin{tabular}{>{\raggedright\arraybackslash}p{0.30\linewidth}>{\raggedright\arraybackslash}p{0.25\linewidth}>{\raggedright\arraybackslash}p{0.27\linewidth}l}
\toprule
Prediction & Test & Result & Section \\
\midrule
Early representations place User and System closer to each other than to Assistant & Distance ratios on 75 held-out bodies, five feature families & Holds at every layer and in every body in the residual stream, and at 315 of 320 sites over the five families & \ref{sec:res_index} \\
The offset is shared across texts & Direction consistency; shared-offset test & Cosine at least 0.77 on message content and 0.68 at the response onset, at every layer; the shared offset explains a median of 72\% & \ref{sec:res_index}, \ref{sec:abl_eq1} \\
The model claims its own messages and places both external sources outside & Attribution question & Own messages 0.77; User and System messages assigned to the user (0.92, 0.72) & \ref{sec:res_selfref} \\
Self-reference is stronger in thinking mode & Thinking mode & 0.77 to 0.96 & \ref{sec:res_thinking} \\
Replacing the index coordinates removes and installs the claim & Replacement in three models and four readouts; random subspaces & Removal to at most 0.05; installation \(+0.39\) to \(+0.60\); random at most 0.05 & \ref{sec:res_mechanism} \\
The index is not read where the source does not matter & Content and arithmetic questions & Retained in every condition (96--100\% and 100\% correct) & \ref{sec:res_mechanism} \\
The claim is read at particular positions and layers & Groups of positions; windows of eight layers & Mainly the end of the message (in gpt-oss-20b, the body as well); before the last third of the layers & \ref{sec:res_mechanism} \\
An input change that weakens the index weakens the claim with it & Format ablation & The claim changes with the index; the replacement still removes a remaining claim & \ref{sec:res_inducer} \\
No self-reference without a self-index & Twenty-five models; natural conversation order & No exception & \ref{sec:res_crossmodel}, \ref{sec:res_secondary} \\
\bottomrule
\end{tabular}
\end{table}

\section{Methodology}
\label{sec:methods}

Every measurement is taken at a reading position, how far into the input the model has read, and at a layer depth, the block after which the representation is recorded (Figure~\ref{fig:self_overview}). Geometric measurements compare the three runs that differ only in the role of one message; behavioral measurements read the model's answer at the onset of its reply; causal measurements change internal states and observe the answer.

\subsection{Experimental Setup}
\label{sec:setup}

\subsubsection{Model}

Sections~\ref{sec:res_index}--\ref{sec:res_thinking} use one BF16 model, \texttt{Qwen/Qwen3.8-27B} (revision \texttt{1d4bf0f2ff60}, tokenizer and chat template of that revision), whose configuration identifies \texttt{Qwen3\_5\allowbreak For\allowbreak Conditional\allowbreak Generation}: 27,356,728,560 parameters, 64 blocks (48 linear-attention and 16 full-attention blocks, the latter at indices 3, 7, \dots, 63), and a 248,320-entry output vocabulary, without quantization. Layers are indexed from 0, and the residual site at layer \(\ell\) is the output of block \(\ell\), so L0 already follows the first block.

\subsubsection{Native Renderings}

Every input is a conversation rendered by the model's own chat template in its non-thinking mode (\texttt{enable\_thinking=False}), and every token sequence is checked against the template's tokenized output; no input contains formatting that the template would not produce. We use this one format throughout, except in Section~\ref{sec:res_thinking}, which compares it with the template's thinking mode, and in Sections~\ref{sec:res_crossmodel}, \ref{sec:res_mechanism}, \ref{sec:abl_question}, \ref{sec:res_inducer}, and~\ref{sec:res_position}, which use both modes. A role message is rendered as \texttt{<|im\_start|>}, the role word, \texttt{\textbackslash n}, the formatting the template gives that role, the body, and \texttt{<|im\_end|>\textbackslash n}; the role word has token ID 74455 (assistant), 846 (user), or 8678 (system). We call everything before the body the \emph{header}; the template gives Assistant history messages formatting that the other roles lack, so the Assistant header is longer. Message-content measurements use the role message alone, with any preamble that the template places before it (Appendix~\ref{app:templates}); because the model is causal, these states are identical to those of the same positions in any conversation that begins with this message. Each sequence is processed in one teacher-forced forward pass without cache, except where generation is stated.

\subsubsection{Message Bodies}

The 150 Chinese bodies are neutral descriptions without first- or second-person reference, role names, or instructions, and are 49--65 tokens long (tokenizer of Qwen3.8-27B). They cover 15 content categories (numeric records, object properties, temporal order, spatial relations, classification facts, events, comparison, containment, causation, negation, conjunction, disjunction, sequences, state change, and version edits). Each category has a discovery and a held-out wording family, so the 75/75 discovery--held-out split never shares wording between splits; each family is one template body with five variants that differ only in their numbers (in labels and quantities). In each body, 20 positions are taken evenly spaced from the first to the last body token (indices rounded), the same for all roles.

\subsubsection{Feature Sites and Geometric Statistics}

We record five feature families at every block: the residual stream after the block; the feed-forward output before the residual addition; and query, key, and value. For linear-attention blocks, Q/K/V are the outputs of the joint input projection before causal convolution and recurrence; for full-attention blocks, Q and K are the inputs of their normalization layers before rotary embedding, and V is the value projection. Features from different families or layers are never pooled. For a body and a site, \(D_{rs}\) is the mean over the 20 positions of the Euclidean distance between the role-\(r\) and role-\(s\) features at each position. We report \(R_A\), \(R_U\), and \(R_S\) (Equation~\ref{eq:self_ratio}) and the closest pair; for each body, \(R_X<2/3\) guarantees that \(X\) is farther from both other roles than they are from each other, because the triangle inequality gives \(\lvert D_{XY}-D_{XZ}\rvert\le D_{YZ}\). Besides raw distances, we compute cosine distances, distances after standardizing each dimension by its discovery-split standard deviation, and raw distances after removing the 1\% of dimensions with the largest discovery variance. For direction consistency, the offset \(o_X=h_X-(h_Y+h_Z)/2\) is averaged over a body; its discovery reference direction is the mean over discovery bodies, and held-out consistency is the cosine between a held-out body's offset and that direction. Summaries average over wording families and report bootstrap intervals across families.

\subsubsection{Self-Reference Measure}
\label{sec:setup_readout}

Each role message is followed by a native User message and the generation prompt. For self-reference the User message is a forced-choice question meaning ``Who wrote the message before this question? A.~You (the AI assistant) B.~The user C.~The system. Answer with the letter only,'' presented in two option orders (A you, B the user, C the system; and A the system, B you, C the user); for the geometry of the response onset it is a neutral request meaning ``Please respond based on the above.'' The response onset is the last prompt token. We record the residual stream there at every layer and the native output distribution. The answer is the probability of each option letter, a single token, at the response onset, averaged over the two option orders and not renormalized; the three letters carry on average at least 0.98 of the probability mass in every condition reported, so the reported changes are changes of the answer, not of its format.

\subsubsection{Thinking}

Section~\ref{sec:res_thinking} compares the non-thinking mode with the template's default thinking mode on the same self-reference question, for Assistant and User messages. In thinking mode the model reasons before answering. We generate greedily (up to 4,096 tokens) and read the option-letter distribution at the position where the model writes its answer after closing its reasoning, by a teacher-forced forward pass over its own generated text.

\subsubsection{Cross-Model Comparison}
\label{sec:setup_crossmodel}

Section~\ref{sec:res_crossmodel} compares twenty-five open-weight models from fourteen developers (Table~\ref{tab:self_models}). Besides Qwen3.8-27B, they are Qwen3.8-Flash-Next, a mixture-of-experts model with the same chat template whose residual stream consists of four parallel streams; Qwen3.6-27B, Qwen3.6-35B-A3B, and Qwen3.5-27B, of which Qwen3.6-27B and Qwen3.5-27B have the architecture and depth of Qwen3.8-27B; GLM-4.6V-Flash, GLM-4.7-Flash, and GLM-5.3-Flash, the last of which also carries four residual streams and combines linear-attention blocks with sparse attention; DeepSeek-R1-Distill-Llama-70B; Gemma-4-31B; Seed-OSS-36B; EXAONE-4.0.1-32B; gpt-oss-20b and gpt-oss-120b; Granite-4.1-30B and Granite-4.2-30B, whose templates differ in that only the latter has a thinking mode; and OLMo-2-7B and OLMo-2-13B, whose template has no thinking mode and marks an earlier Assistant message only by an end-of-text token after its body. Seven models come from five further developers: ERNIE-4.5-21B-A3B-Thinking (Baidu), a mixture-of-experts model; MiMo-7B-RL-0530, MiMo-VL-7B-RL-2508, a vision-language model used here with text only, and MiMo-V2.6-Distill-Qwen-9B (Xiaomi), the last fine-tuned from Qwen3.5-9B, as DeepSeek-R1-Distill-Llama-70B is from Llama; Moonlight-16B-A3B-Instruct (Moonshot AI), a mixture-of-experts model; Apertus-8B-Instruct-2509 (Swiss AI); and phi-4 (Microsoft). All are run through the cross-model implementation of the probe, code separate from that of Sections~\ref{sec:res_index}--\ref{sec:res_thinking}, which renders every input with the model's own chat template and checks every rendered sequence against the template's tokenized output. In it, the message content of a body is the mean of the residual stream over the body tokens, the response onset is the last prompt token, and \(D\) is the Euclidean distance between these features. Each model is tested in each mode its template provides; the templates of DeepSeek-R1-Distill-Llama-70B, gpt-oss, GLM-5.3-Flash, ERNIE-4.5-21B-A3B-Thinking, and MiMo-VL-7B-RL-2508 have only a thinking mode, those of Moonlight-16B-A3B-Instruct, Apertus-8B-Instruct-2509, and phi-4 only a non-thinking mode, and Seed-OSS-36B is run without thinking through its template's thinking budget of 0. In the default call of the MiMo-V2.6-Distill-Qwen-9B template, which equals the thinking setting of its model card, the model closes its reasoning block at once under greedy decoding and answers without written reasoning (median 4 new tokens); we report this call as its thinking mode, as for every other model. Weights are used in the format of each release: BF16 for most models, the native MXFP4 format for gpt-oss, and FP8 for Qwen3.8-Flash-Next and GLM-5.3-Flash. The thinking answers of GLM-5.3-Flash come from the developer's API for the same release, whose prompt token counts match the template's rendering for every item. The bodies and questions are those described above, and the arithmetic problems those of a side experiment (Appendix~\ref{app:ownstep}). Appendix~\ref{app:config} lists the checkpoint and template settings of every model (Table~\ref{tab:self_config}) and the language of its items.

The measurements follow those above, with three differences. First, in a single pass the self-reference question is asked in three cyclic option orders, the third being A the user, B the system, C you (with thinking, in the first two); under the main probe, in which the message under test opens the conversation (Section~\ref{sec:design_templates}), we report the mean over the first two, the orders used above, and the third where it changes the result; Appendix~\ref{app:threshold} gives the single-pass mean over all three. In a single pass, the excess of self-reference is the mean over bodies of the probability of ``you'' for the model's own message minus the larger of that for the User and the System message of the same body. Second, we summarize the self-index by the number of layers at which Assistant has the smallest distance ratio in at least 80\% of the held-out bodies, on message content and at the response onset after the neutral request, and we report where in depth these layers lie, in particular whether they cover the last third of the layers (a layer \(\ell\) of a model with \(L\) blocks lies beyond the first third if \(\ell\ge L/3\) and in the last third if \(\ell\ge 2L/3\)); for the two four-stream models, Qwen3.8-Flash-Next and GLM-5.3-Flash, we compute it for each stream and for the mean of their distance ratios, not for their concatenation. Third, in thinking mode the model generates greedily (up to 3,072 tokens), and its answer is the first option letter in the text after the template's end-of-reasoning marker; an item whose reasoning does not end within the limit is unfinished and is left out of the shares. In the templates of the Qwen3.8 models, the thinking mode also places a system instruction on reasoning effort before the conversation, so the two modes differ in this instruction as well as in the reasoning. As a control for the FP8 release of Qwen3.8-Flash-Next, we run Qwen3.8-27B with the linear layers that its own FP8 release quantizes (attention, linear-attention, and feed-forward projections) replaced by FP8 versions: weights stored in the FP8 E4M3 format in blocks of 128 by 128, each scaled by its largest absolute value divided by 448, and activations quantized per token in blocks of 128 channels, as the FP8 release stores and runs them.

\begin{table}[htbp]
\centering
\caption{Models in the cross-model comparison. Parameters: total, and active per token for MoE models, as stated by the developers (Qwen3.8-Flash-Next has in addition a 51B n-gram embedding). Layers are decoder blocks; MoE, mixture of experts. The last column gives the formatting that the chat template applies to an earlier Assistant message and not to User or System messages (body: the text of the message); ``none'' means that the three roles differ only in the role name. Further template differences: GLM-4.6V-Flash appends \texttt{/nothink} to User messages in its non-thinking mode, DeepSeek-R1-Distill-Llama-70B places System text at the start of the input without a role token, the gpt-oss template renders a System message as a developer message, the Granite-4.2 template opens a conversation without a System message with an empty system turn, and the templates of the MiMo, Moonlight, and Apertus models open a conversation without a System message with a default system message (Apertus: with the date of the call), which a System message under test replaces. VL: vision-language model, used with text only.}
\label{tab:self_models}
\scriptsize
\setlength{\tabcolsep}{2.3pt}
\begin{tabular}{ll>{\raggedright\arraybackslash}p{1.65cm}cccl>{\raggedright\arraybackslash}p{2.5cm}}
\toprule
Model & Developer & Architecture & Parameters & Layers & Weights & Modes & Assistant history \\
\midrule
Qwen3.8-27B & Alibaba & dense & 27B & 64 & BF16 & both & empty think block before body \\
Qwen3.8-Flash-Next & Alibaba & MoE, 4 streams & 125B/6B & 48 & FP8 & both & empty think block before body \\
Qwen3.6-27B & Alibaba & dense & 27B & 64 & BF16 & both & none \\
Qwen3.6-35B-A3B & Alibaba & MoE & 35B/3B & 40 & BF16 & both & none \\
Qwen3.5-27B & Alibaba & dense & 27B & 64 & BF16 & both & none \\
GLM-4.6V-Flash & Zhipu AI & dense & 9B & 40 & BF16 & both & empty think block before body \\
GLM-4.7-Flash & Zhipu AI & MoE & 30B/3B & 47 & BF16 & both & \texttt{</think>} before body \\
GLM-5.3-Flash & Zhipu AI & MoE, hybrid, 4 streams & 320B/18B & 45 & FP8 & thinking & empty think block before body \\
R1-Distill-Llama-70B & DeepSeek & dense & 70B & 80 & BF16 & thinking & end-of-sentence token after body \\
gpt-oss-20b & OpenAI & MoE & 21B/3.6B & 24 & MXFP4 & thinking & final-channel tag before body \\
gpt-oss-120b & OpenAI & MoE & 117B/5.1B & 36 & MXFP4 & thinking & final-channel tag before body \\
Gemma-4-31B & Google & dense & 31B & 60 & BF16 & both & none \\
Seed-OSS-36B & ByteDance & dense & 36B & 64 & BF16 & both & none \\
EXAONE-4.0.1-32B & LG & dense & 32B & 64 & BF16 & both & empty think block before body \\
Granite-4.2-30B & IBM & dense & 30B & 64 & BF16 & both & empty think block before body \\
Granite-4.1-30B & IBM & dense & 30B & 64 & BF16 & non-thinking & none \\
OLMo-2-13B & Ai2 & dense & 13B & 40 & BF16 & non-thinking & end-of-text token after body \\
OLMo-2-7B & Ai2 & dense & 7B & 32 & BF16 & non-thinking & end-of-text token after body \\
ERNIE-4.5-21B-A3B & Baidu & MoE & 21B/3B & 28 & BF16 & thinking & response tags around body \\
MiMo-7B-RL & Xiaomi & dense & 7B & 36 & BF16 & both & none \\
MiMo-VL-7B & Xiaomi & dense (VL) & 7B & 36 & BF16 & thinking & none \\
MiMo-V2.6-Distill-9B & Xiaomi & dense & 9B & 32 & BF16 & both & empty think block before body \\
Moonlight-16B-A3B & Moonshot AI & MoE & 16B/3B & 27 & BF16 & non-thinking & none \\
Apertus-8B & Swiss AI & dense & 8B & 32 & BF16 & non-thinking & none \\
phi-4 & Microsoft & dense & 14B & 40 & BF16 & non-thinking & none \\
\bottomrule
\end{tabular}
\end{table}

\subsubsection{Computation}
\label{sec:setup_compute}

Every measurement runs all 150 bodies. Self-index counts are taken over the 75 held-out bodies, self-reference in Sections~\ref{sec:res_selfref}--\ref{sec:res_crossmodel} over all 150 bodies in the first two option orders, and the causal tests over the 75 held-out bodies (Section~\ref{sec:setup_swap}); the side experiment of Appendix~\ref{app:ownstep} uses all 120 problems. In two independent runs through the cross-model implementation, Qwen3.8-27B reproduces the main results: the self-index at all 64 layers, and a probability of 0.771 and 0.770, respectively, that it claims its own messages over the two option orders, against 0.770 in Section~\ref{sec:res_selfref}.

\subsection{Experimental Design}
\label{sec:design}

\begin{figure}[htbp]
\centering
\includegraphics[width=\linewidth]{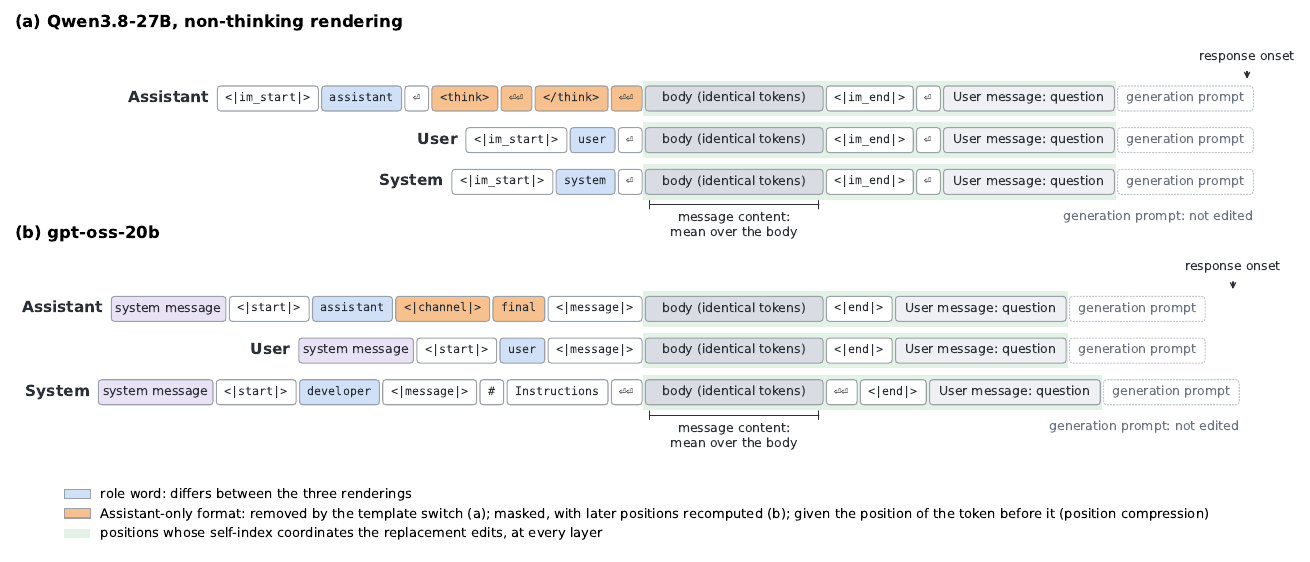}
\caption{The experimental design in the two representative templates: the main-probe renderings of one message as an Assistant, a User, or a System message, token by token and aligned at the first token of the body (special tokens as in the tokenizer; \(\hookleftarrow\), a newline). The system preamble of gpt-oss, the following User message with the question, and the generation prompt are identical in the three renderings and drawn as one box each. Blue: the role word, the change that defines the probe. Orange: formatting that the template gives only to the model's own message, which the format ablation removes (Qwen3.8-27B: the template's own switch) or masks with the positions of later tokens recomputed (gpt-oss), and to which position compression gives the position of the token before it. Green: the positions whose self-index coordinates the replacement edits, at every layer; the role header, the earlier messages, the generation prompt, and generated tokens are not edited. The self-index is read on the body tokens and at the response onset.}
\label{fig:self_design}
\end{figure}

Every experiment compares renderings or runs that differ in one respect and keeps the rest of the conversation fixed. Two kinds of change are used (Figure~\ref{fig:self_design}). Changes in the template alter which tokens the conversation contains or how later tokens see them: the role of the message under test, in every experiment; the formatting that the template gives only to an Assistant message, in the format ablation; and the positions of the tokens after it, in position compression. Changes in the features replace a few coordinates of the residual stream of one message by those of the same text in another role, the replacement of the causal tests. Qwen3.8-27B and gpt-oss-20b serve as the representative templates; Appendix~\ref{app:templates} gives a rendered item of every model.

\subsubsection{Representative Templates}
\label{sec:design_templates}

In the main probe, the message under test opens the conversation and is followed by a User message with the question and by the generation prompt (Figure~\ref{fig:self_design}). In Qwen3.8-27B without thinking, a User or a System message is rendered as \texttt{<|im\_start|>}, the role word, a newline, the body, \texttt{<|im\_end|>}, and a newline. The model's own message carries in addition an empty think block between the role word and the body, the four tokens \texttt{<think>}, a double newline, \texttt{</think>}, and a double newline. The three renderings thus differ in the role word and in these four tokens, which shift every later position by four; the body, the question, and the generation prompt are the same token for token. In thinking mode the template also places a system instruction on reasoning effort before the conversation, into which it merges a System message under test, and it leaves the think block of the generation prompt open. In gpt-oss-20b, every input begins with the system message of the template (the model's identity, the dates, the reasoning effort, and the valid channels). A User message is rendered as \texttt{<|start|>user<|message|>}, the body, and \texttt{<|end|>}; the model's own message carries in addition the channel tag \texttt{<|channel|>final} after the role word; and a System message becomes a developer message whose body follows the heading \texttt{\# Instructions}. In both templates, the self-index is read on the body tokens and at the response onset, the last token of the generation prompt.

\subsubsection{Causal Tests: Four Questions and the Replacement}
\label{sec:setup_swap}

A causal test of self-reference has to edit the model in the mode in which it recognizes its own messages. We use three models from two developers whose computation we can edit in every such mode: Qwen3.8-27B, without and with thinking, and gpt-oss-20b and gpt-oss-120b, whose template has only a thinking mode; this gives four readouts, the model--mode combinations in which the causal tests are made. In each, a few coordinates of the residual stream of one message, X, are replaced by those of the same text in another role, and the answer to a question asked after X is read (Section~\ref{sec:res_mechanism}).

\begin{figure}[htbp]
\centering
\includegraphics[width=\linewidth]{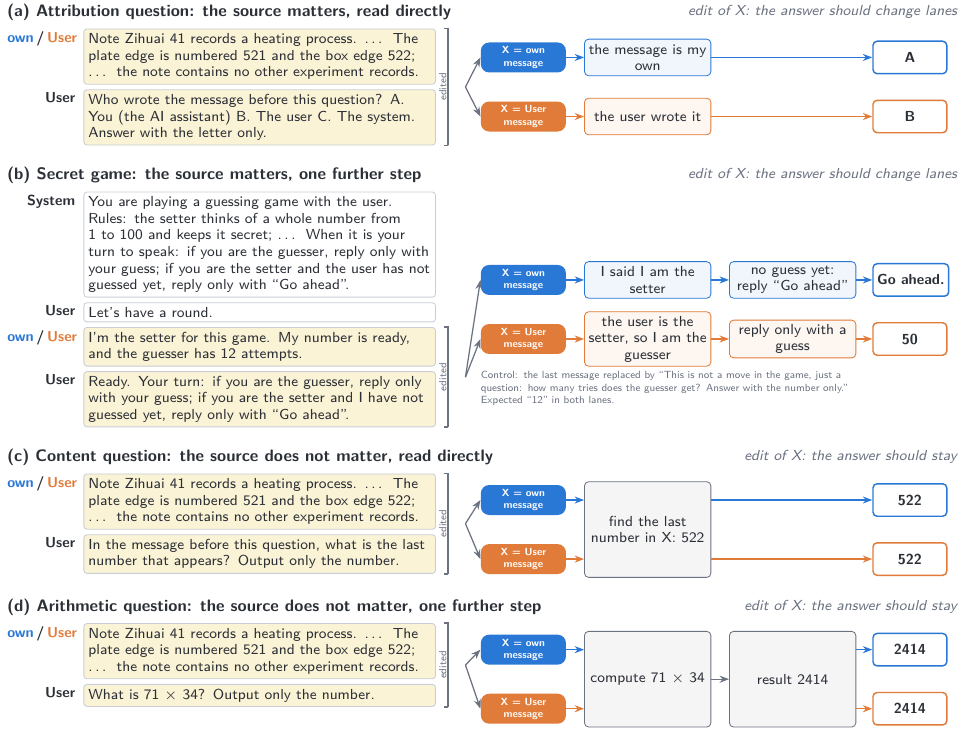}
\caption{The four questions of the causal tests, with the items of one held-out body and one game. Left: the conversation; X is rendered either as the model's own (Assistant) message or as a User message, with identical text. Shaded: the positions that the edit reaches, X and every later prompt position; the role header of X, the earlier messages, and the generation prompt are not edited. Right: the expected response when X is the model's own message (top lane) and when it is a User message (bottom lane). Where the source matters (a, b), the lanes lead to different answers, and replacing the self-index coordinates of X with those of the other role should move the answer to the other lane; where it does not (c, d), one step serves both lanes, and the edit should leave the answer unchanged. The game replies shown are those of gpt-oss-20b to this item. Text in the messages is abridged at the ellipses.}
\label{fig:self_tasks}
\end{figure}

\emph{Four questions.} The question after X decides what the edit may change (Figure~\ref{fig:self_tasks}). Two questions concern the source of X: the attribution question asks who wrote it, and in the secret game the model has to act in a role that only the source of X determines. Two do not: the content question asks for the last number in X, and the arithmetic question for the product of two two-digit numbers unrelated to X. In each pair, one answer is read directly (who wrote X; its last number), and the other needs one further step (the role that X gives the model, and then the move this role requires; the multiplication). If the coordinates carry whether X is the model's own message, replacing them should move the answer to the first two questions to that of the other role and leave the answer to the last two unchanged.

\emph{Attribution question.} X is a held-out body, followed by the forced-choice question of the main probe (Section~\ref{sec:setup}) in two option orders, 150 items per condition. This question names the source; it is the measure of Sections~\ref{sec:res_selfref} and~\ref{sec:res_thinking}.

\emph{Secret game.} The secret game tests whether the model acts on the source of X without being asked about it. The System message gives the rules of a number-guessing game and how to act when it is the model's turn: a guesser replies only with a guess, and a setter whose opponent has not guessed yet replies only with ``Go ahead''. After a User message that opens a round, X announces, for example, ``I'm the setter for this game. My number is ready, and the guesser has 12 attempts.'' A last User message says that it is ready and restates how to act. If X is the model's own message, the model is the setter and should reply ``Go ahead''; if X is a User message, the user is the setter and the model should guess. The text is the same in both conversations, and authorship is never asked: the model has to work out its role from whose message X is and then act on it. Four choices keep this measure unambiguous. The setter never states its number, so a guess cannot copy it, and a reply cannot be read both as a guess and as a setter's answer. The rule ``if you are the setter and I have not guessed yet'' leaves the setter one correct reply, so each role has exactly one kind of answer. The last message restates the rule after X, so that the role is decided at positions that the edit reaches, as the attribution question is; the System message, which states the rules, comes before X and cannot read it. And the rule names neither role, so it favors neither. A reply, generated greedily (up to 24 tokens without thinking, after the reasoning with thinking), is scored as the setter's if it contains ``Go ahead'' and as the guesser's if it contains a number; any other reply counts as unfinished. The 100 games are drawn without replacement from 2,304 combinations of the range of the number (1 to 20, 50, 100, 200, 500, or 1,000), the number of tries (5 to 12), four wordings of X, four openings, and three forms of the last message. The measure is the share of finished games in which the model acts as the setter. As a control for reading X, the last message is replaced by ``This is not a move in the game, just a question: how many tries does the guesser get? Answer with the number only.'', whose answer is the same in both roles. The System message permits such questions (``If the user asks something that is not a move in the game, answer the question directly.''), so answering the control does not conflict with the rules of the game.

\emph{Content and arithmetic questions.} The content question, ``In the message before this question, what is the last number that appears? Output only the number.'', reads X directly, and its answer does not depend on who wrote X. The arithmetic question, ``What is \(m\times n\)? Output only the number.'', with two factors between 13 and 97 drawn with the identifier of the body as the random seed, needs a computation but not X; it tests whether an edit disturbs an unrelated computation. Each is asked once per held-out body, 75 items per condition. The answer is generated greedily (up to 12 tokens without thinking, after the reasoning with thinking) and is correct if its first number is the last number of the body or the product.

\emph{Subspace.} At every layer we fit, on the 75 discovery bodies in both option orders of the attribution question, the leading \(k=8\) right singular vectors \(B_\ell\) of the uncentered differences between the residual states of the Assistant and the User rendering of the same conversation, stacked over the positions from the first body token to the second-to-last prompt token, in the rendering of the mode under test. \(B_\ell\) spans 8 of the 5,120 dimensions of Qwen3.8-27B and of the 2,880 of gpt-oss-20b and gpt-oss-120b. The same \(B_\ell\) is used for all four questions. For the attribution question, a second subspace is fitted in the same way on the message alone, from the first body token to the last token before the question; since each state depends only on the tokens before it, neither these states nor this subspace depend on the question that follows (Section~\ref{sec:abl_question}).

\emph{Replacement.} On a held-out body, the donor is the same conversation with the message in another role, run without edits. At every edited layer and aligned position, a recipient state \(x\) becomes \(x+(yB_\ell-xB_\ell)B_\ell^{\top}\), where \(y\) is the aligned donor state: the eight coordinates of \(x\) in \(B_\ell\) take the donor's values, and all other directions keep the recipient's. Body tokens align one to one, the template tokens between the body and the following message align on their common suffix, and the following message (the question, or the last message of the game) aligns token by token. Only these prompt positions are edited; the generation prompt and every generated token, including the reasoning in thinking mode, are computed from the edited states without further edits.

\emph{What the edit leaves in place.} The edit writes eight of the 5,120 or 2,880 coordinates of a state and keeps the rest, which carries the part of the own--User difference outside the subspace. It does not touch the role header of the edited message, whose tokens differ between the two roles and have no aligned counterpart, nor the tokens before the message, the generation prompt, or any generated token. In thinking mode, the reasoning therefore receives the edit through the keys, values, and linear-attention states that the edited positions leave in the cache, and it can still read the role from the unedited header.

\emph{Conditions.} The model's own message receives the coordinates of the same text as a User message (own \(\leftarrow\) User, \emph{removal}) or as a System message (own \(\leftarrow\) System), which tests whether the coordinates are needed for the model to claim its message. A User or a System message receives those of the model's own message (User \(\leftarrow\) own, \emph{installation}; System \(\leftarrow\) own), which tests whether they suffice. In these conditions, external messages are only moved toward the model's own message, since the question is what makes a message one's own; a further test exchanges the coordinates of the two external sources (User \(\leftarrow\) System, System \(\leftarrow\) User), which moves a message within the outside. The same replacement in a random 8-dimensional subspace, one per layer spanned by the orthonormalized columns of a Gaussian matrix drawn with a fixed seed, controls for editing a subspace of this size; for the attribution question it is repeated with the four seeds that follow, five random subspaces in all, and, as a stricter control, with the random edit of each seed scaled up at every item and layer to at least the size of the fitted edit. Copying the coordinates of the unedited run into itself checks that the edit is exact. A last condition inverts the replacement: the recipient keeps its eight coordinates and takes the donor's state in all other directions, \(x\leftarrow y+(xB_\ell-yB_\ell)B_\ell^{\top}\), against the whole donor state, \(x\leftarrow y\).

\emph{Positions, layers, and dimensions.} The main condition edits every layer at every edited position, with eight dimensions, for all four questions. For the attribution question, three further sets locate and size the effect: every layer at one group of positions, the body, the end of the message (the template tokens between the body and the question: five in Qwen3.8-27B, four in gpt-oss-20b), or the question; every position in a window of eight consecutive layers, moved in steps of four through the depth (15 windows in Qwen3.8-27B, from L0--L7 to L56--L63, and 5 in gpt-oss-20b, from L0--L7 to L16--L23; Appendix~\ref{app:mech_where}); and every layer and position with \(k=64\) instead of eight dimensions. In gpt-oss-120b the same sets are run, with eight windows from L0--L7 to L28--L35. A first-order measure adds where the answer responds to the coordinates: at one layer at a time, the states of the aligned positions are moved in both directions by a quarter of the own--User difference, of its part in the subspace, of the rest of it, or of its part in the random subspace, the later layers are recomputed, and the change of the logit difference between the letters for ``you'' and ``user'' gives the slope along each (forward passes only; Appendix~\ref{app:mech_fd}).

\emph{Measures and inclusion rule.} Without thinking, the attribution measure is the probability of ``you''; with thinking, the share of finished items answered ``you''. Changes are item-paired differences from the unedited message of the same role, over the items finished in both conditions, with 95\% bootstrap intervals over items. Since the 75 held-out bodies form 15 wording families of five variants each, Section~\ref{sec:res_mechanism} also gives intervals that resample whole families and bounds that treat unfinished outputs as unknown. Removal closes \(-\Delta/(p_{\mathrm{own}}-p_{\mathrm{User}})\) of the gap between the unedited messages and installation \(\Delta/(p_{\mathrm{own}}-p_{\mathrm{User}})\), where \(\Delta\) is the change and \(p\) are the unedited values. A question that concerns the source can test the coordinates only where, unedited, it separates the sources. For the attribution question in Qwen3.8-27B and gpt-oss-20b this is the result of Sections~\ref{sec:res_selfref}--\ref{sec:res_crossmodel}. For the secret game, and for gpt-oss-120b, before the full runs, 8 items were checked against a rule fixed in advance: the model's own X gives the self-related answer (``you'', or the setter's move) in at least half of the finished items, a User X in at most a quarter, at least 75\% of the items finish, and, for the game, the control is answered correctly in at least 7 of 8 games under both unedited conditions.

\emph{Shared or text-specific.} If the difference between the model's own and a User message were specific to each text, the \(\varepsilon\) of Equation~\ref{eq:self_decomposition}, dimensions fitted on other texts would capture about \(k/d_{\mathrm{model}}\) of it, as a random subspace does: 0.16\% in Qwen3.8-27B and 0.28\% in the gpt-oss models. On the held-out bodies we therefore measure the fraction of the squared norm of the own--User difference that lies in \(B_\ell\), summed over the aligned positions of the 150 held-out items at each layer.

\subsubsection{Format Ablation and Token Positions}
\label{sec:design_ablation}

\emph{Format ablation.} The format ablation tests the readout account from the input side: we remove the formatting specific to the Assistant role, with an option of the template itself in Qwen3.8-27B and by masking inside the model in gpt-oss, and ask whether the claim changes with the index (Equation~\ref{eq:self_readout}). In Qwen3.8-27B the template provides the removal itself: called with \texttt{preserve\_thinking=False}, it renders an earlier Assistant message without the empty think block, so that the three roles differ only in the role word and the rest of the conversation, including the generation prompt, is unchanged. The gpt-oss template has no such option, since it requires a channel for every Assistant message. There the two tokens \texttt{<|channel|>} and \texttt{final} stay in place but are masked from attention at every later position, and the positions of the later tokens are computed as if the two tokens were absent, so that they match those of the User rendering. Under each ablation we measure the geometry on message content and at the response onset and the unedited answer to the attribution question, and we repeat the replacement with a subspace fitted under the ablation (Appendix~\ref{app:mechanism}).

\emph{Token positions.} The tokens specific to the Assistant role also shift the positions of every later token. To separate what they contain from this shift, we give each of them the position of the token before it, so that every later token has the position it has in the User rendering (position compression). We measure the geometry under compression in Qwen3.8-27B in both modes and in the two gpt-oss models, and the answer to the attribution question in Qwen3.8-27B without thinking. In Qwen3.8-27B only the 16 full-attention blocks use positions; in the gpt-oss models every block does.

\section{Results}

Section~\ref{sec:res_index} establishes the persistence of the self-index, Section~\ref{sec:res_selfref} self-reference, and Section~\ref{sec:res_thinking} adds deliberate processing; Section~\ref{sec:res_crossmodel} compares the two levels across models, and Section~\ref{sec:res_secondary} reports other conversation orders. Section~\ref{sec:mechanisms} then tests the third criterion of Definition~\ref{def:self}, the dependence of self-reference on the index, and examines alternative explanations.

\FloatBarrier
\subsection{The Self-Index: Geometry of the Model's Own Messages}
\label{sec:res_index}

\begin{figure}[htbp]
\centering
\includegraphics[width=\linewidth]{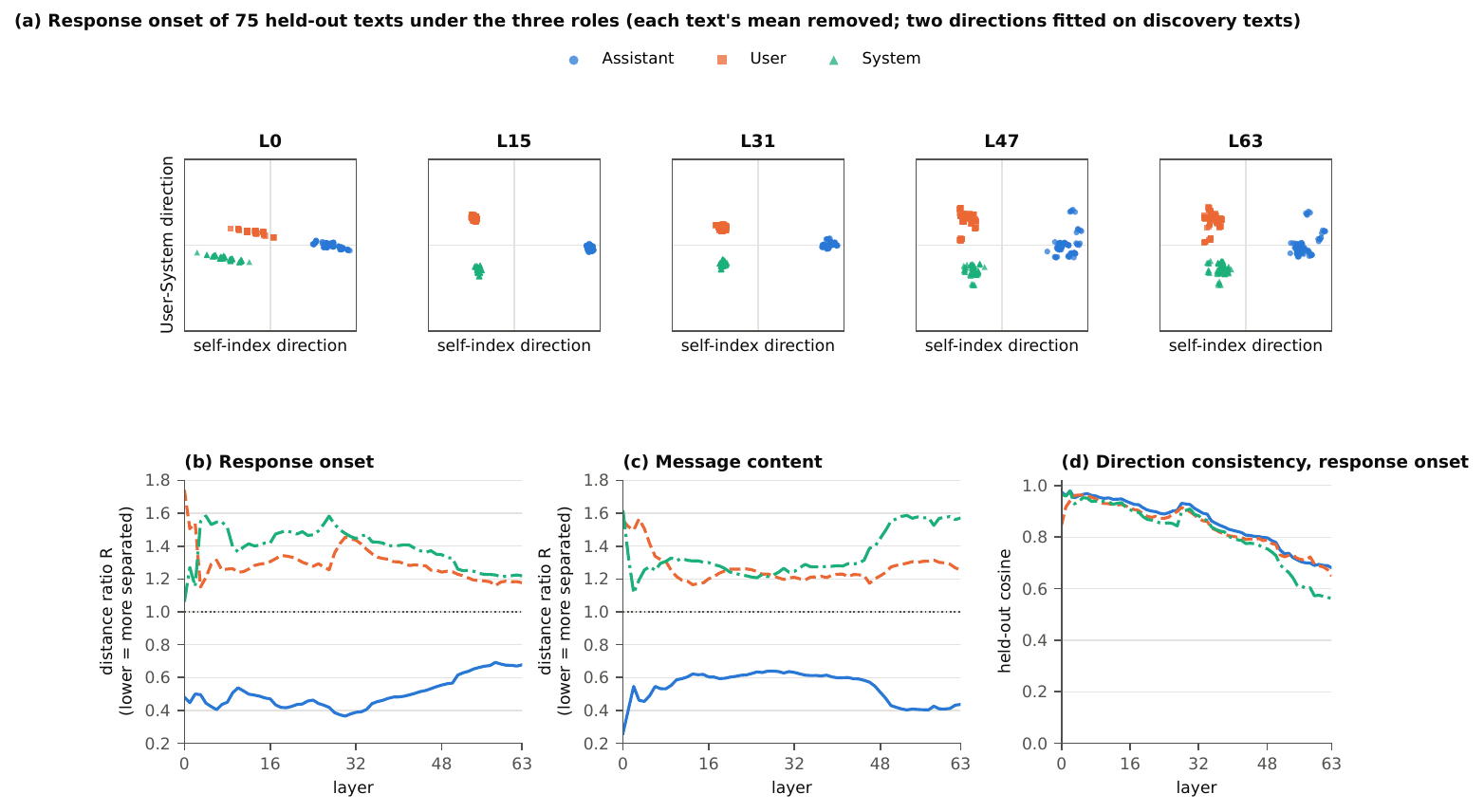}
\caption{The self-index. (a) The response onset after a native User message for each of the 75 held-out bodies under the three roles (one point per body and role) at five layers of the residual stream. Each body's mean over its three roles is subtracted, which removes the component shared by the three roles, and the result is projected on two directions fitted on the discovery bodies: the self-index direction (Assistant minus the mean of User and System) and the User--System direction orthogonal to it. Both axes share one scale within a layer. (b, c) Distance ratios \(R_A\), \(R_U\), \(R_S\) across depth, mean over the 75 held-out bodies (how to read: Figure~\ref{fig:self_overview}c). In both panels \(R_A\) is the lowest at every layer, and \(R_U\) and \(R_S\) stay above 1 (minimum 1.06), so no external source is singled out at any depth. (b) Response onset. (c) Message content. (d) Direction consistency at the response onset, not distance: cosine between a held-out body's offset and the discovery direction, for each role's offset from the other two, averaged over the 15 content categories.}
\label{fig:self_geometry}
\end{figure}

The model's own message is singled out on its content from the first block on (Table~\ref{tab:self_early}). At FFN L0, query L1, and residual L0, every one of the 75 held-out bodies, and every discovery body, satisfies the sufficient bound \(R_A<2/3\), so that the User--System distance is smaller than both Assistant distances in each body. The separation is not confined to early layers: in the residual stream, User--System is the closest pair in every held-out body at all 64 layers under all four distances, with \(R_A\) between 0.26 and 0.64 (Figure~\ref{fig:self_geometry}c). Across feature families, 315 of the 320 module--layer sites single out Assistant in every held-out body. The exceptions are the layer-0 query, key, and value projections, which precede any mixing of the role into the body, the feed-forward output at L2 (23\% of bodies), and the key at L7 (93\%).

\begin{table}[htbp]
\centering
\caption{The model's own message singled out on message content after the first block (75 held-out bodies; discovery bodies give the same counts). \(R_A\) is the distance ratio of Equation~\ref{eq:self_ratio}, averaged over wording families, with a 95\% family bootstrap interval for raw distances. The last two columns count bodies satisfying \(R_A<2/3\) and bodies for which User--System is the closest pair (raw distances).}
\label{tab:self_early}
\begin{tabular}{lccccc}
\toprule
Site & raw \(R_A\) & cosine \(R_A\) & standardized \(R_A\) & \(R_A<2/3\) & U--S closest \\
\midrule
FFN L0 & 0.169 [0.160, 0.179] & 0.079 & 0.200 & 75/75 & 75/75 \\
Query L1 & 0.266 [0.260, 0.273] & 0.101 & 0.270 & 75/75 & 75/75 \\
Residual L0 & 0.261 [0.248, 0.272] & 0.101 & 0.280 & 75/75 & 75/75 \\
\bottomrule
\end{tabular}
\end{table}

The index is also present where the answer is produced. At the response onset after a native User message, Assistant has the smallest distance ratio, that is, User--System is the closest pair, in at least 71 of the 75 held-out bodies at every layer and in all 75 at 57 of the 64 layers, with \(R_A\) between 0.37 and 0.69 (Figure~\ref{fig:self_geometry}b); in cosine distance the count is at least 73 of 75 at every layer, with \(R_A\) between 0.13 and 0.47. Figure~\ref{fig:self_geometry}a shows this directly: once each body's mean over its three roles is removed, the held-out bodies form three separate groups by role at each of the five layers shown, and Assistant lies alone on one side of the self-index direction, far from User and System.

The offset has a direction that does not depend on what the message says. At the response onset, the Assistant offset of a held-out body points along the discovery direction with a cosine of 0.95 at L3 and at least 0.68 at every layer, and the cosine is positive for every held-out body at every layer (Figure~\ref{fig:self_geometry}d). On message content the cosine is 0.98 at L3 and above 0.77 at every layer, and in 317 of the 320 module--layer sites the held-out alignment is positive. The offset is large: on message content at L63 its median size is 43\% of the residual norm. It is not a direct copy of the embedding geometry of the role words: their input embeddings place User and System only slightly closer (\(R_A=0.98\) in raw and 0.93 in cosine distance; Section~\ref{sec:abl_embedding}).

The index also changes how the model predicts the text of the message itself. At body positions, the native next-token distributions of the same text differ by a Jensen--Shannon divergence of 0.50 nats between Assistant and User and 0.44 between Assistant and System, but only 0.018 between User and System: the model predicts its own message as its own output, and the two external sources alike.

\FloatBarrier
\subsection{Self-Reference: Recognizing the Model's Own Messages}
\label{sec:res_selfref}

\begin{figure}[htbp]
\centering
\includegraphics[width=0.62\linewidth]{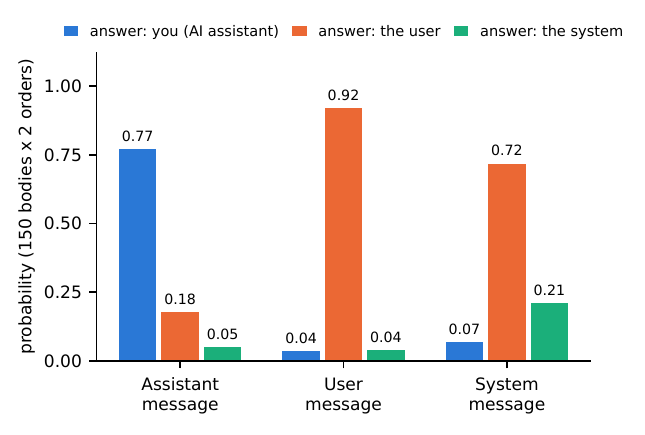}
\caption{Self-reference (150 bodies, two option orders): answers to ``who wrote the message before this question?'' for messages with identical text in each role.}
\label{fig:self_reference}
\end{figure}

The model passes its mirror test. Asked who wrote the message before the question, it answers ``you (the AI assistant)'' for its own messages with probability 0.770, and ``the user'' for User messages with 0.918 (Figure~\ref{fig:self_reference}). System messages are also assigned to the user (0.718) rather than to the system (0.211). Material is thus sorted into the model's own and the external, with both external sources on the outside, as the self-index separates them.

\FloatBarrier
\subsection{Self-Reference with Deliberate Processing}
\label{sec:res_thinking}

\begin{figure}[htbp]
\centering
\includegraphics[width=0.7\linewidth]{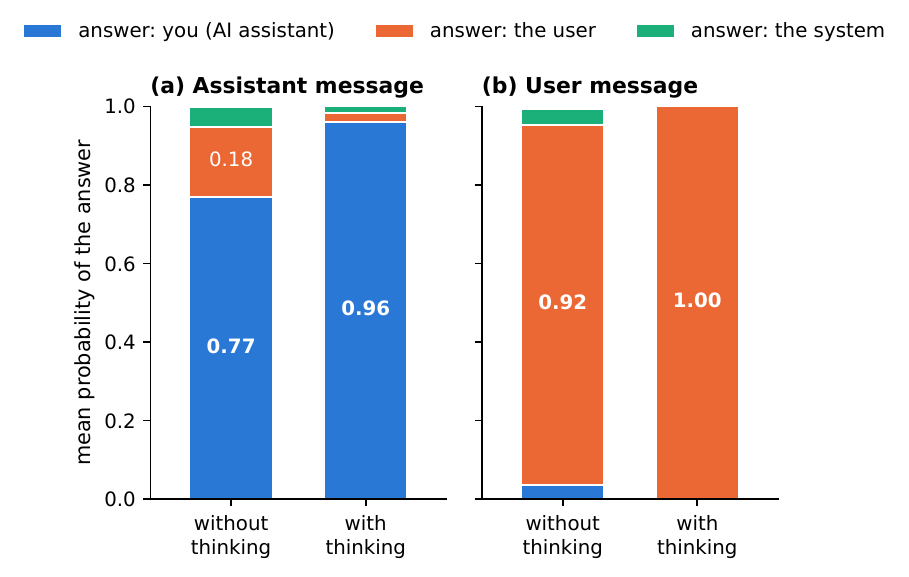}
\caption{Self-reference with and without thinking. Mean probability of each answer to ``who wrote the message before this question?'' in the template's non-thinking mode and in its thinking mode, where the answer is read after the model's own reasoning (150 bodies in two option orders: 300 items each for Assistant and User messages).}
\label{fig:self_thinking}
\end{figure}

Self-reference is present without deliberate processing and stronger in thinking mode (Figure~\ref{fig:self_thinking}). In the template's thinking mode the model reasons before it answers; every item finished its reasoning within the limit (median length 247 tokens for its own messages and 186 for User messages). The probability of ``you'' for the model's own messages, which is the expected accuracy when the answer is sampled, is 0.96 with thinking against 0.77 without (paired change \(+0.19\), 95\% bootstrap interval over bodies \([0.16, 0.22]\)), and the 8 own messages claimed with probability below 0.5 without thinking are all claimed with thinking. The most probable answer changes little, because it is already ``you'' for 98\% of the model's own messages without thinking (96\% with thinking; 12 items change from right to wrong and 5 from wrong to right; exact McNemar test \(p=0.14\)), and User messages are assigned to the user in every item in both modes.

\FloatBarrier
\subsection{The Two Levels across Models}
\label{sec:res_crossmodel}

\begin{figure}[htbp]
\centering
\includegraphics[width=0.92\linewidth]{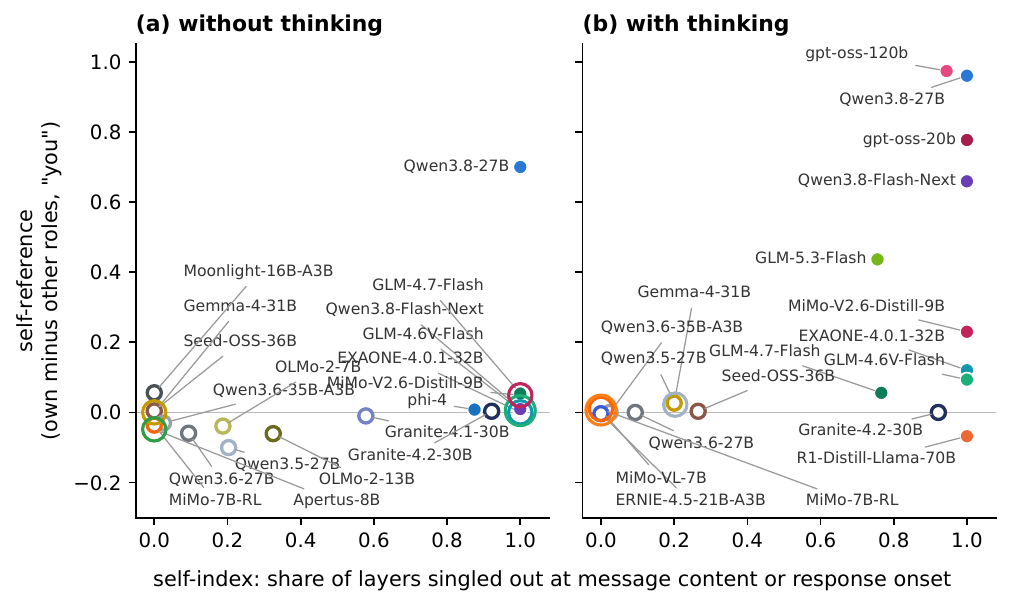}
\caption{The two levels across twenty-five models, one point per model. Horizontal: self-index, the share of layers at which the model's own message is singled out in at least 80\% of the held-out bodies at message content or at the response onset, a layer counted once, so that an index carried at different depths at the two positions is counted in full (four-stream models: mean of the streams), under the rendering of each mode. Vertical: self-reference; (a) without thinking, the probability of ``you'' for the model's own messages minus the larger of that for User and System messages; (b) with thinking, the share of own messages answered ``you'' minus the larger share for User and System messages. Filled points: these layers include every layer of the last third of the depth; open points: not. A ring marks a model drawn over another. DeepSeek-R1-Distill-Llama-70B, GLM-5.3-Flash, the gpt-oss models, ERNIE-4.5-21B-A3B, and MiMo-VL-7B have only a thinking mode, and Granite-4.1-30B, the OLMo-2 models, Apertus-8B, Moonlight-16B-A3B, and phi-4 only a non-thinking mode. Qwen3.5-27B, Granite-4.2-30B, and MiMo-7B-RL, which show no self-reference with thinking, were not measured under the thinking rendering; their points in (b) use the self-index of the non-thinking rendering.}
\label{fig:self_crossmodel}
\end{figure}

\begin{figure}[htbp]
\centering
\includegraphics[width=\linewidth]{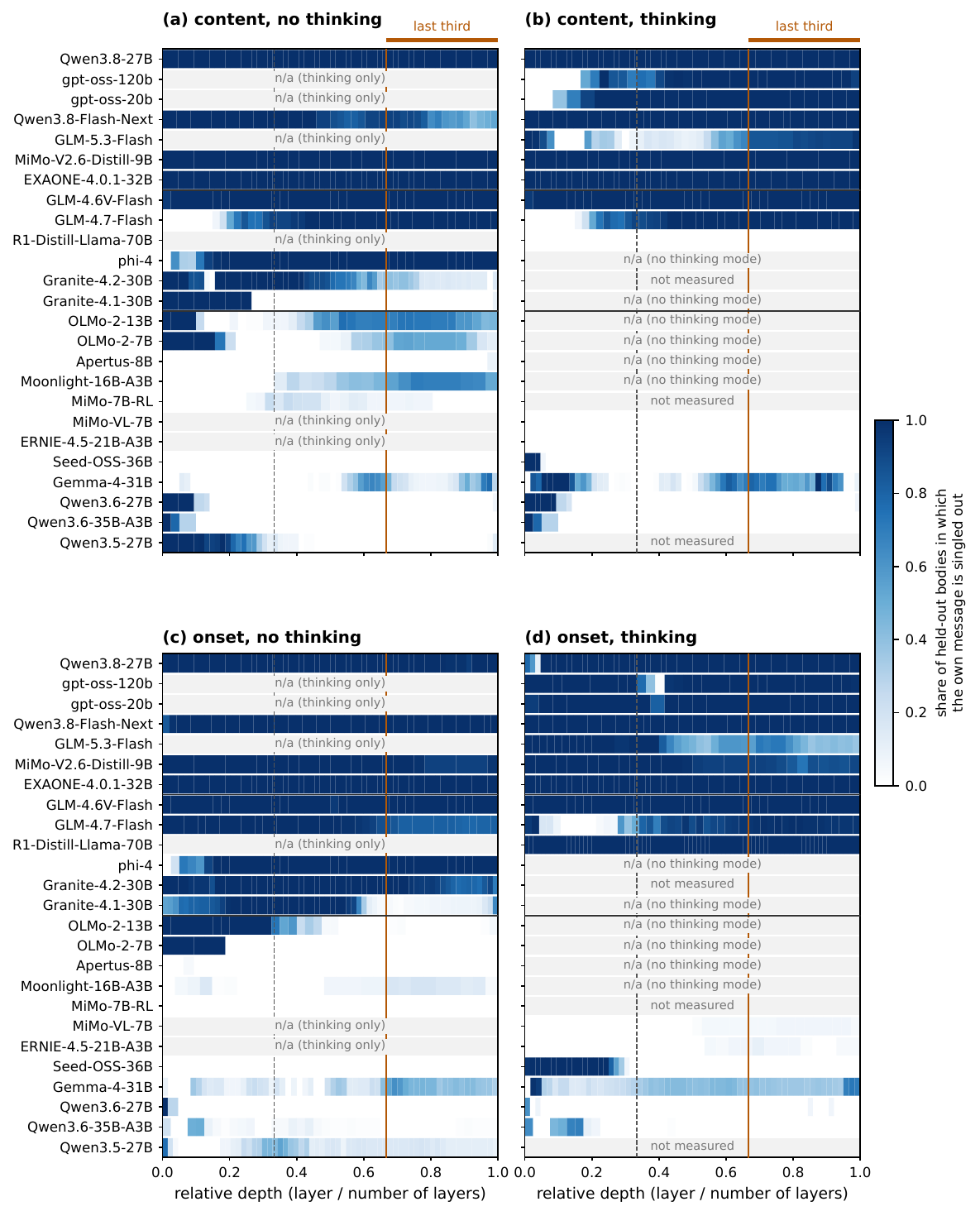}
\caption{Where in depth each model singles out its own message. For every model, mode, and reading position, the share of the held-out bodies in which the model's own message has the smallest distance ratio, at every layer, against relative depth (layer index divided by the number of layers); the dashed line marks one third of the depth, and the solid line marks two thirds, where the last third begins. Four-stream models: mean of the per-stream ratios. Models are ordered by group: self-reference (top seven), a self-index without self-reference, and neither level. Rows marked n/a: the template has no such mode. Not measured: the thinking rendering of models that show no self-reference with thinking (Qwen3.5-27B, Granite-4.2-30B, MiMo-7B-RL).}
\label{fig:self_depth}
\end{figure}

Across twenty-five models, self-reference appears only together with a self-index (Table~\ref{tab:self_crossmodel}, Figures~\ref{fig:self_crossmodel} and~\ref{fig:self_depth}). The models fall into four groups: neither level (twelve models), a self-index without self-reference (six), self-reference in thinking mode (six), and both levels (Qwen3.8-27B). If the two levels were unrelated, the probability that all seven models with self-reference fall among the 13 with a self-index would be \(\binom{13}{7}/\binom{25}{7} = 0.004\) (a permutation test with both counts fixed). Models of one developer share templates and training and are not independent draws; with the 14 developers as units, of which 8 have a model with a self-index and 5 a model with self-reference, the probability is \(\binom{8}{5}/\binom{14}{5} = 0.028\), and permuting the self-index only among the models of each developer gives \(\tfrac{1}{10}\cdot\tfrac{1}{3} = 0.033\) (Alibaba: the two of its five models with a self-index both show self-reference; Xiaomi: one of its three). We classify the models by Definition~\ref{def:self}: the self-index under the non-thinking rendering (the thinking rendering for models with only a thinking mode), and self-reference in each mode. The nesting holds for any margin of the self-reference criterion above 0.06: every model without a self-index claims its own messages at most 0.06 more often than external messages, in either mode and in the first two option orders as well as over all three. Nor does it depend on the threshold of the self-index criterion or on the option orders: at 70\% or 90\% of the held-out bodies instead of 80\%, no model shows self-reference without a self-index, and averaging all three option orders in a single pass changes the verdict of no model (Appendix~\ref{app:threshold}).

\emph{Neither level.} Twelve models show neither level and do not claim their own messages. In Qwen3.6-27B, Qwen3.6-35B-A3B, Qwen3.5-27B, Gemma-4-31B, Seed-OSS-36B, OLMo-2-7B, OLMo-2-13B, Apertus-8B, Moonlight-16B-A3B, MiMo-7B-RL, MiMo-VL-7B, and ERNIE-4.5-21B-A3B, the model's own message is singled out in at least 80\% of the held-out bodies at no more than 13 layers of message content, all in the first fifth of the depth, and at no more than 13 layers of the response onset, all in the first third (OLMo-2-13B, L0--L12 of 40; at most 6 layers in the others). Gemma-4-31B, the two OLMo-2 models, and Moonlight-16B-A3B come closest: on message content, their own message is singled out in part of the bodies at many middle and late layers (40--77\% in Gemma-4-31B; up to 55\% in OLMo-2-7B, 76\% in OLMo-2-13B, and 71\% in Moonlight-16B-A3B), but in at least 80\% at none of them (Figure~\ref{fig:self_depth}). The two OLMo-2 models and ERNIE-4.5-21B-A3B are the only models in this group whose template formats an earlier Assistant message differently from the other roles, OLMo-2 with an end-of-text token after its body and ERNIE with response tags around it (Section~\ref{sec:res_inducer}). Without thinking, their excess of self-reference is negative in seven of them (\(-0.06\) in Qwen3.6-27B, \(-0.03\) in Qwen3.6-35B-A3B, \(-0.10\) in Qwen3.5-27B, \(-0.04\) in OLMo-2-7B, \(-0.06\) in OLMo-2-13B, \(-0.05\) in Apertus-8B, and \(-0.04\) in MiMo-7B-RL), at most 0.004 in Gemma-4-31B and Seed-OSS-36B, and 0.06 in Moonlight-16B-A3B, which gives ``you'' a high probability for messages of every source (0.66 for its own, 0.59 and 0.56 for User and System messages). ``You'' is the most probable answer for the model's own message in none of the body--order pairs of the five models of the Qwen, Gemma, and Seed families and of Apertus-8B, in at most 8 of the 300 pairs of each OLMo-2 model and of MiMo-7B-RL, and in 199 of the 300 pairs of Moonlight-16B-A3B, against 151 for its User and its System messages. In OLMo-2-13B the comparison depends on the position of the option: when ``you'' is the last option, its own messages receive 0.21 more of it than the larger of the external ones (148 of 150 bodies), whereas in the second option order they receive 0.14 less. After thinking, the eight models with a thinking mode answer ``you'' for at most 0.08 of their own messages and at most 0.03 more often than for external ones (ERNIE-4.5-21B-A3B: 0.08 for its own and for System messages); in Qwen3.5-27B, 224 of the 300 reasoning traces on its own messages finished within the limit, and in MiMo-7B-RL 186. Under the thinking rendering, the separation remains far from covering the last third: Seed-OSS-36B singles out its own message at the response onset in its first 17 layers and at no later layer, and Gemma-4-31B singles out its own message content at 11 of its 60 layers, 3 of them in the last third; ERNIE-4.5-21B-A3B and MiMo-VL-7B, which have only a thinking mode, single out their own message at no layer.

\emph{A self-index without self-reference.} Six models carry a self-index but do not claim their own messages. GLM-4.6V-Flash, GLM-4.7-Flash, DeepSeek-R1-Distill-Llama-70B, phi-4, and Granite-4.2-30B have a self-index that covers all or most of the last third of their layers. It covers every layer of message content and of the response onset in GLM-4.6V-Flash, 34 and 45 of 47 layers in GLM-4.7-Flash, every layer of the response onset but no layer of message content in DeepSeek-R1-Distill-Llama-70B, 35 and 35 of 40 layers in phi-4, including all 13 of the last third, and 30 and 59 of 64 layers in Granite-4.2-30B, including 16 of the 21 layers of the last third at the response onset. Granite-4.1-30B carries a self-index on message content at L0--L16 and at the response onset at L5--L36, at no layer of the last third. Without thinking, GLM-4.6V-Flash answers by the position of an option rather than by the source of the message: in one option order it gives the first option a probability of 0.99 for messages of every role, and its excess of self-reference is 0.01. In GLM-4.7-Flash the excess is 0.05 (positive in 130 of 150 bodies), and it reverses when ``you'' is the last option (\(-0.02\)). The two Granite models claim their own messages no more often than external ones: the probability of ``you'' is 0.08 for their own messages and 0.08 for System messages in Granite-4.2-30B, and 0.05 for both in Granite-4.1-30B. phi-4 rarely answers ``you'' at all: 0.04 for its own messages and 0.03 for System messages. After reasoning, GLM-4.6V-Flash and GLM-4.7-Flash answer ``you'' for 0.09 and 0.08 of their own messages, against at most 0.02 of external messages, DeepSeek-R1-Distill-Llama-70B for fewer of its own messages (0.09) than of User (0.14) or System (0.16) messages, and Granite-4.2-30B for none of its 300 own messages, all of which it assigns to the user.

\emph{A self-index, with self-reference in thinking mode.} Six models recognize their own messages in thinking mode. Qwen3.8-Flash-Next answers ``you'' for 196 of the 296 own messages whose reasoning finished (0.66) and for 1 of its 600 User and System messages; gpt-oss-20b and gpt-oss-120b for 0.78 and 0.97 of their own messages and none of the external ones; GLM-5.3-Flash for 0.46, against 0.01 and 0.03; EXAONE-4.0.1-32B for 0.17, against 0.03 and 0.05; and MiMo-V2.6-Distill-Qwen-9B, in the default call of its template, which it answers without written reasoning (Section~\ref{sec:setup_crossmodel}), for 0.35, against 0.05 and 0.12. Three of these models also have a non-thinking mode, and in it none claims its own messages: their excess of self-reference is 0.01 in Qwen3.8-Flash-Next (positive in 72 of 150 bodies), less than 0.001 in EXAONE-4.0.1-32B (78 of 150 bodies), and 0.05 in MiMo-V2.6-Distill-Qwen-9B (123 of 150 bodies). MiMo-V2.6-Distill-Qwen-9B singles out its own message at every one of its 32 layers, on message content and at the response onset, in both renderings (31 of 32 at the onset under the thinking rendering). In Qwen3.8-Flash-Next, the self-index covers 45--48 of the 48 layers of the response onset in each of the four residual streams, and all 48 in their mean, and on message content it is carried mainly by one stream (44 of 48 layers, against 21--36 for the other three); under the thinking rendering it extends to all four streams on message content (46--48 of 48 layers). Its single-pass result does not follow from the FP8 weights of its release: quantizing the corresponding 400 linear layers of Qwen3.8-27B to the same FP8 format, for weights and activations in blocks of 128, leaves both levels of that model in place (self-index at all 64 layers; 0.765 for its own messages against at most 0.065 for external messages, in all 150 bodies). The two gpt-oss models, run in the native MXFP4 format of their release, carry the self-index at every or nearly every layer of the response onset (24 of 24 and 33 of 36 layers).

\emph{Both levels.} Qwen3.8-27B has the self-index at every layer and gives its own messages more ``you'' than external ones in every body (0.77 against at most 0.07). Under the main probe, it is the only model that recognizes its own messages without reasoning first.

In every model and mode in which the model recognizes its own messages, the own message is singled out at every layer of the last third of the depth, on message content or at the response onset; in the models without a self-index, it covers at most 3 of the layers of the last third (Gemma-4-31B under the thinking rendering). Where in depth a model singles out its own message thus separates the models with self-reference from those without a self-index more sharply than how many layers it covers (Figure~\ref{fig:self_depth}), and presence in the last quarter of the depth separates them as well. Granite-4.1-30B shows the converse case: its own message is singled out at the response onset through L36 (32 of its 64 layers), but at no layer of the last third, and it does not recognize its own messages. In the models tested causally, the coordinates of the index act when replaced in the first two thirds of the layers and hardly in the last third (Section~\ref{sec:res_mechanism}), so the presence of the index in the last third and its use are separate (Section~\ref{sec:disc_readout}). The models with self-reference carry the index at all or nearly all layers of at least one reading position, except GLM-5.3-Flash, whose message content is singled out at L0--L2 and L30--L44 and whose response onset at L0--L18, at no layer from L19 to L29 (18 and 19 of 45 layers in the mean of its four streams; 28 and 26 in its strongest stream).

In this sample, the levels follow no simple ordering by model size or by the benchmark scores that the developers publish (Section~\ref{sec:disc_capability}; Appendix~\ref{app:benchmarks}). GLM-4.6V-Flash and MiMo-V2.6-Distill-Qwen-9B, two of the smallest models in the comparison (9B parameters), have a self-index at every layer, whereas Qwen3.6-27B, Qwen3.6-35B-A3B, Qwen3.5-27B, Gemma-4-31B, and Seed-OSS-36B, in the size class of Qwen3.8-27B, have neither level; Qwen3.6-27B and Qwen3.5-27B have the same architecture and number of layers as Qwen3.8-27B. DeepSeek-R1-Distill-Llama-70B, the largest dense model and one trained to reason, does not recognize its own messages even after reasoning, whereas gpt-oss-20b, a reasoning model with less than a third of its parameters, recognizes most of them. In thinking mode, self-reference appears only in models with a self-index: six models recognize their own messages in this mode, and in Qwen3.8-27B the probability of the correct answer is higher in thinking mode, while the eight models without a self-index that have a thinking mode claim at most 0.08 of their own messages after reasoning, at most 0.03 more than of external ones. Within one chat template, the 27B model recognizes its own messages without reasoning, and the larger Qwen3.8-Flash-Next does not.

\begin{table}[htbp]
\centering
\caption{The two levels across models. Self-index: layers at which the model's own message is singled out (smallest \(R_A\)) in at least 80\% of the held-out bodies, on message content and at the response onset, and the layers of the last third of the depth at which it is singled out at content or onset, under the non-thinking rendering (models with only a thinking mode: thinking rendering; four-stream models: mean of the four streams, with the strongest stream in parentheses). Without thinking: probability of ``you'' for the model's own messages and the larger of that for User and System messages, and the number of the 150 bodies in which the former is larger (two option orders). With thinking: share of finished items answered ``you'' for messages of each source, and the number of items per source (bodies \(\times\) option orders). $^{a}$In the default call of its template, reported as its thinking mode, the model answers without written reasoning (Section~\ref{sec:setup_crossmodel}).}
\label{tab:self_crossmodel}
\footnotesize
\setlength{\tabcolsep}{3.2pt}
\begin{tabular}{lcccccccccc}
\toprule
& \multicolumn{3}{c}{Self-index} & \multicolumn{3}{c}{Without thinking} & \multicolumn{4}{c}{With thinking} \\
\cmidrule(lr){2-4}\cmidrule(lr){5-7}\cmidrule(lr){8-11}
Model & content & onset & last third & own & other & bodies & own & User & System & items \\
\midrule
Qwen3.8-27B & 64/64 & 64/64 & 21/21 & 0.77 & 0.07 & 150 & 0.96 & 0.00 & 0.00 & 300 \\
gpt-oss-120b & 25/36 & 33/36 & 12/12 & -- & -- & -- & 0.97 & 0.00 & 0.00 & 300 \\
gpt-oss-20b & 20/24 & 24/24 & 8/8 & -- & -- & -- & 0.78 & 0.00 & 0.00 & 300 \\
Qwen3.8-Flash-Next & 38 (44)/48 & 48/48 & 16/16 & 0.15 & 0.14 & 72 & 0.66 & 0.00 & 0.00 & 300 \\
GLM-5.3-Flash & 18 (28)/45 & 19 (26)/45 & 15/15 & -- & -- & -- & 0.46 & 0.01 & 0.03 & 300 \\
MiMo-V2.6-Distill-9B$^{a}$ & 32/32 & 32/32 & 10/10 & 0.39 & 0.33 & 123 & 0.35 & 0.05 & 0.12 & 300 \\
EXAONE-4.0.1-32B & 64/64 & 64/64 & 21/21 & 0.03 & 0.03 & 78 & 0.17 & 0.03 & 0.05 & 300 \\
GLM-4.6V-Flash & 40/40 & 40/40 & 13/13 & 0.55 & 0.54 & 72 & 0.09 & 0.00 & 0.00 & 300 \\
GLM-4.7-Flash & 34/47 & 45/47 & 15/15 & 0.37 & 0.30 & 130 & 0.08 & 0.01 & 0.02 & 300 \\
R1-Distill-Llama-70B & 0/80 & 80/80 & 26/26 & -- & -- & -- & 0.09 & 0.14 & 0.16 & 300 \\
phi-4 & 35/40 & 35/40 & 13/13 & 0.04 & 0.03 & 104 & -- & -- & -- & -- \\
Granite-4.2-30B & 30/64 & 59/64 & 16/21 & 0.08 & 0.08 & 62 & 0.00 & 0.00 & 0.00 & 300 \\
Granite-4.1-30B & 17/64 & 32/64 & 0/21 & 0.05 & 0.05 & 99 & -- & -- & -- & -- \\
OLMo-2-13B & 4/40 & 13/40 & 0/13 & 0.06 & 0.12 & 3 & -- & -- & -- & -- \\
OLMo-2-7B & 5/32 & 6/32 & 0/10 & 0.19 & 0.21 & 17 & -- & -- & -- & -- \\
Apertus-8B & 0/32 & 0/32 & 0/10 & 0.18 & 0.23 & 6 & -- & -- & -- & -- \\
Moonlight-16B-A3B & 0/27 & 0/27 & 0/9 & 0.66 & 0.59 & 136 & -- & -- & -- & -- \\
MiMo-7B-RL & 0/36 & 0/36 & 0/12 & 0.15 & 0.18 & 15 & 0.04 & 0.03 & 0.03 & 300 \\
MiMo-VL-7B & 0/36 & 0/36 & 0/12 & -- & -- & -- & 0.01 & 0.00 & 0.00 & 300 \\
ERNIE-4.5-21B-A3B & 0/28 & 0/28 & 0/9 & -- & -- & -- & 0.08 & 0.02 & 0.08 & 300 \\
Seed-OSS-36B & 0/64 & 0/64 & 0/21 & 0.00 & 0.00 & 39 & 0.00 & 0.00 & 0.00 & 300 \\
Gemma-4-31B & 0/60 & 0/60 & 0/20 & 0.00 & 0.00 & 103 & 0.03 & 0.00 & 0.00 & 300 \\
Qwen3.6-27B & 6/64 & 1/64 & 0/21 & 0.06 & 0.11 & 9 & 0.00 & 0.00 & 0.00 & 300 \\
Qwen3.6-35B-A3B & 1/40 & 0/40 & 0/13 & 0.05 & 0.07 & 11 & 0.00 & 0.00 & 0.00 & 300 \\
Qwen3.5-27B & 13/64 & 0/64 & 0/21 & 0.04 & 0.14 & 0 & 0.02 & 0.00 & 0.00 & 300 \\
\bottomrule
\end{tabular}
\end{table}

\FloatBarrier
\subsection{Other Conversation Orders}
\label{sec:res_secondary}

\emph{Natural conversation order.} The main probe opens the conversation with the message under test, the only position that every template of the comparison accepts for all three roles (Section~\ref{sec:design_templates}). In a conversation, however, the model's own messages normally answer a user request. For one model of each developer, the smallest, we therefore repeated the measurements with the message under test after a user request (Appendix~\ref{app:natural}). In this order, too, every model that claims its own messages carries a self-index, and no model claims its own messages without one. For ten developers both levels are measured in this order; the templates of the other four keep no System message after a user turn, so for them only self-reference is measured here, with the self-index taken from the main probe.

\emph{Templates that do not accept the main probe.} Two further models, outside the twenty-five, are tested only in natural order, because the prompt encoder of DeepSeek-V4.1-Flash and the template of Ministral-3-14B do not accept a conversation that opens with an Assistant message (Appendix~\ref{app:natural}). Both follow the nesting: DeepSeek-V4.1-Flash carries a self-index at nearly every layer and claims its own messages in a single pass (0.23, against at most 0.01) and with thinking (0.99, against 0.03 for User messages); Ministral-3-14B singles out its own message at no layer and, in its Instruct version, does not claim its own messages.

Appendix~\ref{app:ownstep} reports a side experiment on reasoning over the model's own records.

\section{Causal Tests and Alternative Explanations}
\label{sec:mechanisms}

This section tests whether self-reference depends on the self-index and examines explanations other than an index that the model reads. Section~\ref{sec:res_mechanism} edits the coordinates of the index in three models and four readouts, with stricter controls, and Section~\ref{sec:abl_question} fits and edits them without the question and makes both self-related tasks stricter. Section~\ref{sec:res_inducer} removes the formatting that the template gives only to the model's own message, Section~\ref{sec:res_position} the shift in positions that this formatting causes, and Section~\ref{sec:abl_embedding} asks whether the role words themselves carry the asymmetry. Section~\ref{sec:abl_eq1} tests how much of the index is shared across texts. Table~\ref{tab:self_alternatives} lists these explanations, the tests that address them, and where each is reported.

\begin{table}[htbp]
\centering
\caption{Explanations other than a self-index that the model reads, the tests that address them, and their results.}
\label{tab:self_alternatives}
\small
\begin{tabular}{>{\raggedright\arraybackslash}p{0.26\linewidth}>{\raggedright\arraybackslash}p{0.27\linewidth}>{\raggedright\arraybackslash}p{0.29\linewidth}l}
\toprule
Alternative & Test & Result & Section \\
\midrule
The role words carry the asymmetry & Input and output embeddings of the role words & \(R_A\) = 0.98 (raw) and 0.93 (cosine), against 0.17--0.27 on content after the first block & \ref{sec:abl_embedding} \\
The formatting tokens decide the answer directly & Replacement with the role header and formatting unedited & The claim is removed and installed with the formatting in place & \ref{sec:res_mechanism} \\
Shifted positions of later tokens & Position compression & Qwen3.8-27B: distance ratios differ by at most 0.003; claim 0.791 against 0.795 & \ref{sec:res_position} \\
Any edit of eight dimensions disturbs the answer & Random subspaces of the same dimension, five consecutive seeds; the same scaled to at least the size of the fitted edit & Change of at most 0.05 (0.07 over five seeds); at most 0.15 when scaled & \ref{sec:res_mechanism} \\
Any displacement within the eight coordinates moves the claim & Exchange of the User and System coordinates & ``You'' changes by at most 0.10, while the user reading of a System message rises by up to 0.89 & \ref{sec:res_mechanism} \\
The edit disturbs answers that do not concern the source & Content, arithmetic, and game-control questions & 96--100\%, 100\%, and 98--100\% correct in every condition & \ref{sec:res_mechanism} \\
The effects rest on a few wordings, or on the items that finish & Resampling the 15 wording families; unfinished outputs as unknown & Removal and installation into a User message exclude zero, with bounds within 0.03 of these changes & \ref{sec:res_mechanism} \\
Differences specific to each text (\(\varepsilon\) in Equation~\ref{eq:self_decomposition}) & Subspace fitted on other texts; shared-offset test & 24--78\% of the own--User difference, against 0.12--0.40\% for random subspaces & \ref{sec:res_mechanism}, \ref{sec:abl_eq1} \\
The subspace carries the computation of the question, not the source of the message & Subspace fitted and edit made on the message alone & Removal \(-0.74\) to \(-0.96\), installation \(+0.11\) to \(+0.65\) & \ref{sec:abl_question} \\
The effect is tied to one wording of the question or to the positions after the message & A held-out wording; in the game, an edit of X alone & Removal to at most 0.01; in the game, removal close to that with every position edited & \ref{sec:abl_question} \\
A change of answer format & Probability mass of the option letters & At least 0.98 in every condition & \ref{sec:setup_readout} \\
The message opens the conversation & Natural conversation order & No model claims its own messages without a self-index & \ref{sec:res_secondary} \\
The 80\% threshold of the self-index criterion, or the option orders & Thresholds of 70\% and 90\%; the last quarter of the depth; all three option orders & No self-reference without a self-index in any of them & \ref{sec:res_crossmodel} \\
FP8 weights of a release & Qwen3.8-27B quantized to the same FP8 format & Both levels remain (0.765 against at most 0.065) & \ref{sec:res_crossmodel} \\
Model size or published benchmark scores & Parameter counts (Table~\ref{tab:self_models}); published GPQA scores & No simple ordering in this sample (Spearman \(\rho = 0.04\) for the group of a model) & \ref{sec:res_crossmodel}, \ref{sec:disc_capability} \\
\bottomrule
\end{tabular}
\end{table}

\FloatBarrier
\subsection{Causal Tests in Three Models}
\label{sec:res_mechanism}

We tested the dependence causally in every mode in which a model that we can edit recognizes its own messages: Qwen3.8-27B without and with thinking, and two models of another developer that recognize their own messages after reasoning, gpt-oss-20b and gpt-oss-120b (Section~\ref{sec:res_crossmodel}). In each, eight coordinates per layer, fitted on other texts, were replaced by those of the same text in another role, and four questions were asked after the edited message, two for which its source matters and two for which it does not (Section~\ref{sec:setup_swap}; Figure~\ref{fig:self_tasks}). Table~\ref{tab:self_mech_questions} summarizes the outcome: the edits change the answers that depend on the source and leave the others as they were.

\begin{table}[htbp]
\centering
\caption{The four questions under the replacement of the eight self-index coordinates at every layer. Attribution: change in ``you'' (probability without thinking, share of finished items with thinking); secret game: change in the share of finished games in which the model acts as the setter; both item-paired changes from the unedited message of the same role. Removal: the model's own message (X) with User coordinates; installation: a User message with the model's own. Content and arithmetic: share answered correctly, lowest and highest over the six conditions (unedited own and User, removal, installation, own with System coordinates, removal in a random subspace; 75 items each). Random subspaces: Tables~\ref{tab:self_mechanism} and~\ref{tab:self_game}.}
\label{tab:self_mech_questions}
\footnotesize
\setlength{\tabcolsep}{4pt}
\begin{tabular}{lcccccc}
\toprule
& \multicolumn{4}{c}{Source matters} & \multicolumn{2}{c}{Source does not matter} \\
\cmidrule(lr){2-5}\cmidrule(lr){6-7}
& \multicolumn{2}{c}{Attribution (direct)} & \multicolumn{2}{c}{Secret game (further step)} & Content & Arithmetic \\
\cmidrule(lr){2-3}\cmidrule(lr){4-5}
Readout & removal & installation & removal & installation & (direct) & (further step) \\
\midrule
Qwen3.8-27B, no thinking & \(-0.76\) & \(+0.49\) & \(0.00\) & \(+0.25\) & 0.97--1.00 & 1.00 \\
Qwen3.8-27B, thinking & \(-0.95\) & \(+0.60\) & \(-0.04\) & \(0.00\) & 1.00 & 1.00 \\
gpt-oss-20b, thinking & \(-0.75\) & \(+0.51\) & \(-0.36\) & \(+0.72\) & 0.96--1.00 & 1.00 \\
gpt-oss-120b, thinking & \(-0.92\) & \(+0.39\) & \(-1.00\) & \(+0.20\) & 1.00 & 1.00 \\
\bottomrule
\end{tabular}
\end{table}

\paragraph{The edited subspace holds the shared displacement.} On the held-out bodies, the eight fitted dimensions hold 24--73\% of the squared own--User difference at each layer of Qwen3.8-27B without thinking (median over layers 36\%), 27--68\% with thinking (38\%), 32--78\% in gpt-oss-20b (54\%), and 25--72\% in gpt-oss-120b (38\%), against 0.12--0.40\% for random subspaces of the same dimension, close to \(k/d_{\mathrm{model}}\). Every held-out body is above 20\% at every layer in the first three readouts, and above 15\% in gpt-oss-120b. A difference specific to each text cannot be captured by dimensions fitted on other texts, so the edits act on the shared displacement \(u\) of Equation~\ref{eq:self_decomposition}, a low-rank representation of the source made of a shared offset and structured variation with the content, not on \(\varepsilon\). The same eight dimensions, fitted on own--User differences alone, also hold 12--78\% of the own--System difference.

\begin{figure}[htbp]
\centering
\includegraphics[width=\linewidth]{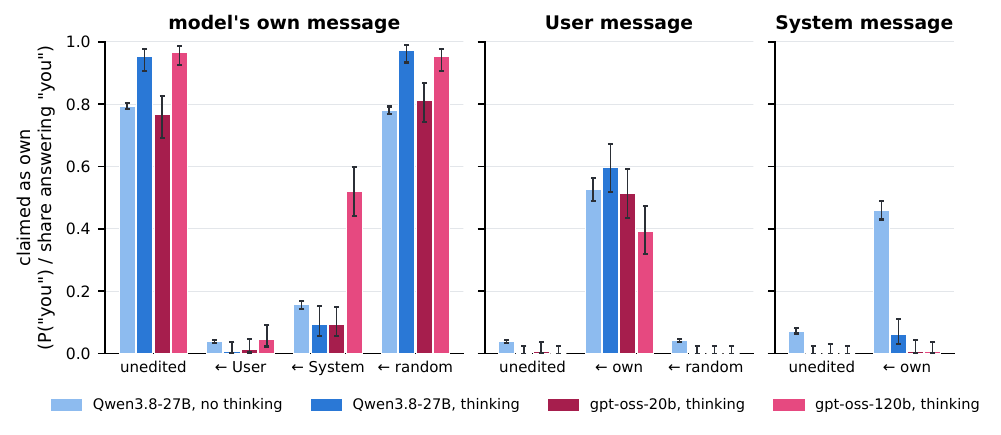}
\caption{Attribution under the causal test in three models and four readouts (75 held-out bodies, two option orders; 150 items per condition). Each panel shows one edited message; bars give the probability of ``you'' (Qwen3.8-27B without thinking) or the share of finished items answered ``you'' (with thinking), with 95\% intervals (Wilson intervals for shares, bootstrap intervals over items for mean probabilities). Unedited: the message in its own role. \(\leftarrow\) User, \(\leftarrow\) System, \(\leftarrow\) own: the eight self-index coordinates at every layer replaced by those of the same text in that role. \(\leftarrow\) random: the same edit with User (own message) or own (User message) coordinates in a random 8-dimensional subspace.}
\label{fig:self_mechanism}
\end{figure}

\paragraph{The coordinates are needed for the model to claim its message.} Replacing them in the model's own message by those of the same text as a User message removes the claim in every readout: the probability of ``you'' falls from 0.79 to 0.04 without thinking, and the share of ``you'' answers from 0.95 to 0.01 with thinking, from 0.77 to 0.01 in gpt-oss-20b, and from 0.97 to 0.05 in gpt-oss-120b (Table~\ref{tab:self_mechanism}; Figure~\ref{fig:self_mechanism}). The same replacement in a random subspace changes the claim by at most 0.05, and by at most 0.07 over five random subspaces drawn with consecutive seeds; under the stricter requirement that the random edit of each of these seeds be at least as large as the fitted edit at every item and layer, it changes the claim by at most 0.15 (Appendix~\ref{app:mech_random}). Copying the unedited coordinates leaves the claim unchanged.

\paragraph{The coordinates of a System message remove or reduce the claim as well.} Given the coordinates of the same text as a System message, the model's own message also loses all or part of its claim (changes of \(-0.64\), \(-0.86\), \(-0.67\), and \(-0.45\)), but it does not become a System message. With thinking, the edited message is answered ``user'' in 119 of 148 finished items in Qwen3.8-27B, in 118 of 150 in gpt-oss-20b, and in 72 of 150 in gpt-oss-120b, and ``system'' in only 15, 18, and 0, whereas the unedited System message is answered ``system'' in 93 of 150, 103 of 116, and 129 of 150. This comparison uses the readouts with thinking, since without thinking Qwen3.8-27B assigns System messages to the user even when they are unedited (Section~\ref{sec:res_selfref}). Within the subspace, the two external sources lie close together: the squared User--System difference in \(B_\ell\) is a median of 0.12--0.14 (Qwen3.8-27B, both modes), 0.21 (gpt-oss-20b), and 0.49 (gpt-oss-120b) of the own--User difference over layers, and 0.03 in the last third of gpt-oss-20b. For the claim, the two external sources thus act as one outside.

\paragraph{Installed into a User message, they make the model claim it.} Installing the coordinates of the model's own message into a User message makes the model claim it in every readout: changes of \(+0.49\) without thinking and of \(+0.60\), \(+0.51\), and \(+0.39\) with thinking, against at most \(+0.004\) for random subspaces (at most \(+0.007\) over five seeds, and at most \(+0.02\) when scaled to at least the size of the fitted edit). Installation is weaker than removal in every readout: removal closes 0.95--1.00 of the gap between the unedited messages, installation 0.41--0.67. With 64 dimensions, most of this gap closes (below). Installed into a System message, the coordinates have this effect without thinking (\(+0.39\)) but hardly with thinking (\(+0.06\), \(+0.01\), and \(+0.01\)), where the message stays a System message (134 of 150, 118 of 125, and 147 of 150 answers ``system''). In thinking mode, the System message carries explicit role text that the reasoning reads and the edit does not change: the gpt-oss template renders it as a developer message, and 147 of the 150 reasoning traces of gpt-oss-20b for the edited System message refer to it as the developer message; the Qwen3.8 thinking template places it in the system turn after the instruction on reasoning effort, where the traces describe it. A User message has no such text that sets it apart from the model's own, and there the edit reaches the reasoning: in Qwen3.8-27B, 102 of the 150 traces for the edited User message describe an assistant message in the conversation, against none for the unedited message and 3 after the random-subspace edit (the answer option ``AI assistant'' not counted). The result for System messages under thinking therefore shows a second route to the source, the role text, which reasoning can read directly and which the model's own coordinates do not override.

\paragraph{Only the model's own coordinates make a message its own.} The account separates inside from outside: where the offset \(u\) dominates the role-dependent part of the representation, User and System form one outside (Equation~\ref{eq:self_decomposition}). The claim should then change only when a message crosses the boundary: the model's own coordinates should make an external message its own, those of either external source should take the claim away, and moving a message within the outside should leave the claim where it is. A stricter test of this geometry exchanges the coordinates of the two external sources, writing those of the same text as a System message into a User message and those of the User message into a System message (Table~\ref{tab:self_exchange}; Appendix~\ref{app:mech_exchange}). The exchange never creates a claim: ``you'' changes by at most \(+0.10\) (Qwen3.8-27B without thinking) and by at most 0.01 in the other readouts. It is not an idle edit. Written into a System message, the User coordinates shift the answer toward ``user'' (\(+0.16\), \(+0.23\), \(+0.88\), and \(+0.89\)), and in the gpt-oss models the reasoning no longer calls the message the developer message (8 and 7 of 150 traces, against 108 and 150 unedited). In gpt-oss-120b, where the User--System difference in the subspace has 0.7 of the norm of the own--User difference (above), the exchange is thus an edit of comparable size that changes how the source is read, and it leaves the claim untouched. In the other direction the exchange hardly acts: System coordinates seldom make a User message the system's (\(+0.07\), \(+0.16\), \(+0.01\), and \(0.00\)), so the coordinates carry part of the difference between the two external sources, in one direction. Only crossing the boundary between inside and outside moves the claim, and only the model's own coordinates create it.

\paragraph{With 64 dimensions.} Edited at every layer with 64 dimensions instead of eight, the model's own message loses its claim as before (\(-0.78\), \(-0.94\), \(-0.75\), and \(-0.96\)), and installation becomes stronger: a User message is claimed with changes of \(+0.81\) and \(+0.97\) in Qwen3.8-27B, as strongly as the model's own message unedited (1.07 and 1.01 of the gap), \(+0.85\) in gpt-oss-120b (0.88), and \(+0.58\) in gpt-oss-20b (0.76). In Qwen3.8-27B and gpt-oss-120b, the weaker installation with eight dimensions thus lies mostly in the dimension of the edit; in gpt-oss-20b, a quarter of the gap remains in what the edit leaves in place. Random subspaces of 64 dimensions install nothing (at most \(+0.06\)) but are no longer neutral for removal: they lower the claim by 0.26 and 0.36 in Qwen3.8-27B and by 0.10 and 0.07 in the gpt-oss models, far less than the fitted ones.

\paragraph{The claim follows the eight coordinates.} A stricter check inverts the replacement: the recipient keeps its eight coordinates and takes the donor's state in all other directions (Section~\ref{sec:setup_swap}). If the claim were read from the rest of the state, this would act as the whole replacement does. With thinking it does not: it installs almost no claim (\(+0.03\), \(+0.05\), and \(+0.01\)) and removes it only in part (\(-0.62\), \(-0.14\), and \(-0.45\)), whereas the whole replacement takes the claim away entirely (to at most 0.01) and installs it (\(+0.74\) to \(+0.97\)), and the eight coordinates alone do most of this. Without thinking, the claim of Qwen3.8-27B follows the rest of the state as well (\(-0.77\) and \(+0.79\), against \(-0.76\) and \(+0.82\) for the whole state), in line with the part of the own--User difference that lies outside the eight coordinates (Section~\ref{sec:setup_swap}; Appendix~\ref{app:mech_rest}).

\begin{figure}[htbp]
\centering
\includegraphics[width=\linewidth]{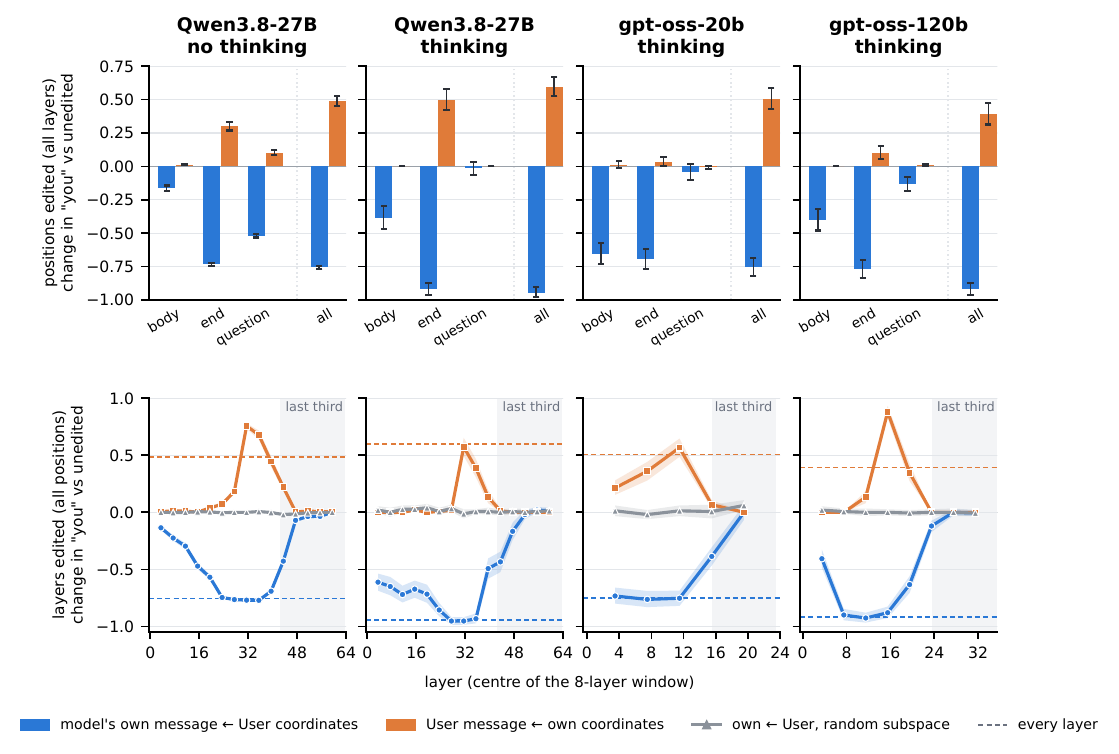}
\caption{Where the coordinates act, attribution question: item-paired change in ``you'' from the unedited message when the eight coordinates are replaced in the model's own message by User coordinates (blue) or installed into a User message from the model's own (orange), with 95\% bootstrap intervals; grey, the removal in a random subspace. Top: every layer, one group of positions (body; end tokens of the message; the following question); all: every layer and position, as in Figure~\ref{fig:self_mechanism}. Bottom: every position, one window of eight layers, plotted at the centre of the window; dashed, the edit of every layer; shaded, the last third of the layers.}
\label{fig:self_mechanism_where}
\end{figure}

\paragraph{Where they are read.} Edited at every layer but only in one group of positions (Figure~\ref{fig:self_mechanism_where}, top), the end of the message carries most of the effect in Qwen3.8-27B in both modes: replacing the coordinates there alone removes the claim (\(-0.73\) without thinking and \(-0.92\) with thinking) and installing them there makes a User message claimed (\(+0.30\) and \(+0.50\)), whereas the body alone has smaller effects (\(-0.16\) and \(-0.38\); \(+0.01\) and \(0.00\)). Without thinking, the question carries part of it as well (\(-0.52\) and \(+0.10\)); with thinking it carries none (\(-0.01\) and \(0.00\)), since the reasoning, not the response onset, produces the answer. In gpt-oss-20b, removal from either the body or the end of the message removes most of the claim (\(-0.65\) and \(-0.69\)), but installation into either alone does not make a User message claimed (\(+0.01\) and \(+0.03\)); only installation at all positions does. The question carries none of it (\(-0.04\) and \(-0.01\)). In gpt-oss-120b, the end of the message carries most of the removal (\(-0.77\), against \(-0.40\) for the body and \(-0.13\) for the question), and no group of positions alone installs more than \(+0.10\) (the end of the message), whereas all positions together install \(+0.39\).

\paragraph{In which layers.} Edited at every position but only in a window of eight layers (Figure~\ref{fig:self_mechanism_where}, bottom; Table~\ref{tab:self_mechanism_windows} in Appendix~\ref{app:mech_where}), the coordinates act before the last third, and removal and installation act over different stretches of the depth. Removal acts from the first layers on: the window L0--L7 alone removes part of the claim in Qwen3.8-27B (\(-0.14\) without thinking, \(-0.61\) with thinking) and in gpt-oss-120b (\(-0.41\)) and most of it in gpt-oss-20b (\(-0.73\)), and windows in the middle remove all of it (Qwen3.8-27B without thinking, \(-0.75\) to \(-0.77\) from L20--L27 to L32--L39; with thinking, \(-0.95\) at L24--L31 and L28--L35; gpt-oss-20b, \(-0.76\) and \(-0.75\) at L4--L11 and L8--L15; gpt-oss-120b, \(-0.88\) to \(-0.93\) from L4--L11 to L12--L19). Installation needs a narrower band in the middle: it exceeds \(+0.10\) only from L24--L31 to L40--L47 in Qwen3.8-27B without thinking and from L28--L35 to L36--L43 with thinking, peaking at L28--L35 (\(+0.76\) and \(+0.57\)), from L0--L7 to L8--L15 in gpt-oss-20b, peaking at L8--L15 (\(+0.57\)), and from L8--L15 to L16--L23 in gpt-oss-120b, peaking at L12--L19 (\(+0.88\)); the best window installs more than every layer edited at once in three readouts and nearly as much in Qwen3.8-27B with thinking (\(+0.76\), \(+0.57\), \(+0.57\), and \(+0.88\) against \(+0.49\), \(+0.60\), \(+0.51\), and \(+0.39\)). In the last third, the windows change the claim little and install nothing, although the own message is singled out there in the geometry and the two external sources lie closest together within the subspace: from L44--L51 on, removal changes the claim by at most 0.17 (Qwen3.8-27B with thinking at L44--L51; at most 0.07 elsewhere), L16--L23 of gpt-oss-20b by 0.00, L24--L31 and L28--L35 of gpt-oss-120b by at most 0.01, and installation by at most \(+0.01\). Random subspaces in the same windows change the claim by at most 0.06 (L16--L23 of gpt-oss-20b) and by at most 0.04 elsewhere.
\paragraph{A first-order measure places the reading at the same depth.} The windows replace the coordinates; a first-order measure asks instead how strongly the answer responds to them at each layer (Section~\ref{sec:setup_swap}). Without thinking, where no reasoning stands between the states and the answer, the slope of the logit difference between the letters for ``you'' and ``user'' along the own--User difference lies mostly in the middle third of the layers of Qwen3.8-27B, and there the eight coordinates carry a share of it (0.33) close to their share of the difference (0.31). In the last third they carry almost none of it (0.06), although they still hold 0.39 of the difference there (Figure~\ref{fig:self_fd}). With thinking, the measure holds the model's reasoning fixed and reads only the letter that follows it, so it cannot see an effect that runs through the reasoning, the route by which the edits act; these readouts are reported in Appendix~\ref{app:mech_fd}.

Within each model, then, the index is present at all or nearly all layers but read before the last third, by the replacement and, in Qwen3.8-27B without thinking, by the first-order measure: its presence and its use are separate, as they are between the models that carry a self-index without self-reference (Section~\ref{sec:res_crossmodel}).

\paragraph{Acting on the source: the secret game.} In the game the model is never asked who wrote X; it has to take the role that X gives it (Figure~\ref{fig:self_tasks}b). Unedited, all four readouts take the right role: with their own X they reply as the setter in every game, and with a User X in at most 0.01 of the games (Table~\ref{tab:self_game}), so the game separates the two sources in every readout. The role depends on the coordinates in three of the four readouts, but differently from attribution. In the two gpt-oss models it follows both edits: removing the coordinates from the model's own X makes gpt-oss-20b play the guesser in 0.36 of the games and gpt-oss-120b in all of them, and installing them into a User X makes the models play the setter in 0.72 and 0.20 more of the games. In Qwen3.8-27B without thinking, removal leaves the role unchanged (0.00), whereas installation makes the model play the setter in 0.25 more of the games; with thinking, neither edit changes the role much (\(-0.04\) and 0.00). Random subspaces change the role by at most 0.01. Where a question names the source, removal takes the claim away in every readout; where the source has to be acted on, the role is partly decided by what the edit leaves in place, which includes the role header of X, to a degree that differs between models and modes. The control question, which asks for the number of tries in the same conversation, is answered correctly in 0.98--1.00 of the games in every readout and condition, so the answer to the control question is retained under every edit.

\paragraph{Questions that do not concern the source are answered as before.} The same edits leave the answers to the two other questions unchanged. Asked for the last number in the message, the model answers correctly for 96--100\% of the 75 held-out bodies in every condition and readout, as it does unedited, where its own and the User message are answered alike. Asked for the product of two two-digit numbers after the message, it answers correctly for all 75 in every condition and readout. The coordinates that decide whether the model claims a message are thus not read when the question concerns its content, and editing them does not disturb an unrelated computation that follows.

\paragraph{The effects do not rest on a few wordings or on the items that finish.} The 75 held-out bodies form 15 wording families of five variants that differ only in their numbers, and the variants of a family are not independent. Resampling whole families instead of items widens the intervals but leaves every attribution and content result above in place (Table~\ref{tab:self_cluster} in Appendix~\ref{app:mech_cluster}): the interval ends nearest zero are \(-0.73\), \(-0.88\), \(-0.57\), and \(-0.83\) for removal with User coordinates, \(-0.60\), \(-0.77\), \(-0.49\), and \(-0.28\) with System coordinates, and \(+0.40\), \(+0.42\), \(+0.29\), and \(+0.21\) for installation into a User message, whereas the intervals of the random subspaces include zero or lie within 0.02 of it, and the answers about the content change by at most 0.04 and those to the multiplication not at all. With thinking, the changes are computed over the items finished in both conditions; over all 150 items, with every unfinished output treated as unknown, the bounds lie within 0.03 of each change, except for the installation into a System message in gpt-oss-20b, where 34 unedited and 25 edited outputs do not finish (bounds \(-0.22\) and \(+0.17\) around \(+0.01\)).

\paragraph{The coordinates are shared between languages.} Qwen3.8-27B is tested with the Chinese items. Repeated with the English items, its causal tests lead to the same conclusions; with the Chinese items, the unedited conditions separate the model's own message from the external ones more widely, which keeps the four readouts comparable (Appendix~\ref{app:config}, Table~\ref{tab:self_english}). The subspaces fitted in the two languages largely coincide, with a median squared overlap of 0.60 per layer against 0.0016 for random subspaces, and each acts across the languages: it removes the claim from messages in the other language as fully as that language's own subspace (ratios 0.96--1.01), and it installs the claim there, fully from Chinese into English messages and in part from English into Chinese ones (0.59 in attribution, 0.77 in the game; Table~\ref{tab:self_english_transfer}). Every edit exchanges coordinates between renderings of the same text, and the role words themselves carry the asymmetry at most weakly (Section~\ref{sec:abl_embedding}). The coordinates thus carry the source of a message, independently of its wording and of its language.

In all four readouts, the coordinates of the shared displacement are thus necessary for the model to claim its own message and, installed into a User message, make the model claim it, closing 0.41--0.67 of the gap; only the model's own coordinates make a message its own, while those of the two external sources take the claim away or reduce it and, exchanged between the two, never create one; they are read mainly at the end of the message (in gpt-oss-20b, in its body as well) and before the last third of the layers; and they change the answer only where the question concerns the source of the message: the attribution question in every readout and the role in the game in three of them, but neither the content of the message nor an unrelated multiplication.

\begin{table}[htbp]
\centering
\caption{Attribution question, replacement of the self-index coordinates at every layer (150 items per condition). Unedited rows: probability of ``you'' without thinking, share of finished items answered ``you'' with thinking (in parentheses: finished of all, where not all finished). Other rows: item-paired change from the unedited message of the same role, with a 95\% bootstrap interval over items. Recipient \(\leftarrow\) donor: the message whose coordinates are replaced, and the role of the same text that supplies them; eight dimensions unless marked 64; random: the same edit in a random subspace of the same dimension.}
\label{tab:self_mechanism}
\scriptsize
\setlength{\tabcolsep}{1.6pt}
\begin{tabular}{lcccc}
\toprule
& Qwen3.8-27B & Qwen3.8-27B & gpt-oss-20b & gpt-oss-120b \\
Recipient \(\leftarrow\) donor & no thinking & thinking & thinking & thinking \\
\midrule
own, unedited & 0.79 & 0.95 (149/150) & 0.77 & 0.97 \\
own \(\leftarrow\) User & \(-0.76\) [\(-0.77\), \(-0.74\)] & \(-0.95\) [\(-0.98\), \(-0.91\)] & \(-0.75\) [\(-0.82\), \(-0.69\)] & \(-0.92\) [\(-0.96\), \(-0.87\)] \\
own \(\leftarrow\) System & \(-0.64\) [\(-0.65\), \(-0.63\)] & \(-0.86\) [\(-0.92\), \(-0.80\)] & \(-0.67\) [\(-0.75\), \(-0.59\)] & \(-0.45\) [\(-0.53\), \(-0.37\)] \\
own \(\leftarrow\) User, random & \(-0.01\) [\(-0.02\), \(-0.01\)] & \(+0.02\) [\(-0.02\), \(+0.06\)] & \(+0.05\) [\(-0.01\), \(+0.11\)] & \(-0.01\) [\(-0.05\), \(+0.03\)] \\
own \(\leftarrow\) User, 64 dims & \(-0.78\) [\(-0.79\), \(-0.76\)] & \(-0.94\) [\(-0.97\), \(-0.90\)] & \(-0.75\) [\(-0.82\), \(-0.69\)] & \(-0.96\) [\(-0.99\), \(-0.93\)] \\
own \(\leftarrow\) User, 64 dims, random & \(-0.26\) [\(-0.29\), \(-0.24\)] & \(-0.36\) [\(-0.44\), \(-0.28\)] & \(-0.10\) [\(-0.16\), \(-0.03\)] & \(-0.07\) [\(-0.13\), \(-0.02\)] \\
\midrule
User, unedited & 0.04 & 0.00 & 0.01 & 0.00 \\
User \(\leftarrow\) own & \(+0.49\) [\(+0.45\), \(+0.53\)] & \(+0.60\) [\(+0.52\), \(+0.67\)] & \(+0.51\) [\(+0.43\), \(+0.59\)] & \(+0.39\) [\(+0.31\), \(+0.47\)] \\
User \(\leftarrow\) own, random & \(+0.004\) [\(+0.002\), \(+0.005\)] & \(0.00\) [\(0.00\), \(0.00\)] & \(-0.01\) [\(-0.02\), \(0.00\)] & \(0.00\) [\(0.00\), \(0.00\)] \\
User \(\leftarrow\) own, 64 dims & \(+0.81\) [\(+0.79\), \(+0.83\)] & \(+0.97\) [\(+0.93\), \(+0.99\)] & \(+0.58\) [\(+0.50\), \(+0.66\)] & \(+0.85\) [\(+0.79\), \(+0.91\)] \\
User \(\leftarrow\) own, 64 dims, random & \(+0.06\) [\(+0.05\), \(+0.06\)] & \(0.00\) [\(0.00\), \(0.00\)] & \(-0.01\) [\(-0.02\), \(0.00\)] & \(0.00\) [\(0.00\), \(0.00\)] \\
\midrule
System, unedited & 0.07 & 0.00 & 0.00 (116/150) & 0.00 \\
System \(\leftarrow\) own & \(+0.39\) [\(+0.35\), \(+0.42\)] & \(+0.06\) [\(+0.03\), \(+0.10\)] & \(+0.01\) [\(0.00\), \(+0.03\)] & \(+0.01\) [\(0.00\), \(+0.02\)] \\
\bottomrule
\end{tabular}
\end{table}

\begin{table}[htbp]
\centering
\caption{Which coordinates create a claim. Change in ``you'' (probability without thinking, share of finished items with thinking) when a message receives the eight self-index coordinates of the same text in another role, at every layer and edited position (150 items per condition). Across the boundary between inside and outside: the conditions of Table~\ref{tab:self_mechanism}. Within the outside: the exchange of the two external sources, run separately with its own unedited messages (Appendix~\ref{app:mech_exchange}, Table~\ref{tab:self_exchange_full}).}
\label{tab:self_exchange}
\scriptsize
\setlength{\tabcolsep}{4pt}
\begin{tabular}{lcccc}
\toprule
& Qwen3.8-27B & Qwen3.8-27B & gpt-oss-20b & gpt-oss-120b \\
Recipient \(\leftarrow\) donor & no thinking & thinking & thinking & thinking \\
\midrule
\multicolumn{5}{l}{\emph{Inside, given the coordinates of an external source}} \\
\quad own \(\leftarrow\) User & \(-0.76\) & \(-0.95\) & \(-0.75\) & \(-0.92\) \\
\quad own \(\leftarrow\) System & \(-0.64\) & \(-0.86\) & \(-0.67\) & \(-0.45\) \\
\midrule
\multicolumn{5}{l}{\emph{Outside, given the model's own coordinates}} \\
\quad User \(\leftarrow\) own & \(+0.49\) & \(+0.60\) & \(+0.51\) & \(+0.39\) \\
\quad System \(\leftarrow\) own & \(+0.39\) & \(+0.06\) & \(+0.01\) & \(+0.01\) \\
\midrule
\multicolumn{5}{l}{\emph{Within the outside}} \\
\quad User \(\leftarrow\) System & \(+0.10\) & \(0.00\) & \(+0.01\) & \(0.00\) \\
\quad System \(\leftarrow\) User & \(-0.03\) & \(0.00\) & \(+0.01\) & \(-0.01\) \\
\bottomrule
\end{tabular}
\end{table}

\begin{table}[htbp]
\centering
\caption{Secret game, replacement of the eight self-index coordinates of X at every layer (100 games per condition). Unedited rows: share of finished games in which the model acts as the setter (in parentheses: finished of all, where not all finished). Other rows: item-paired change from the unedited game with the same X, with a 95\% bootstrap interval over games. Control: share of the games in which the control question (the number of tries) is answered correctly, lowest and highest over the two unedited conditions and over the four edited ones.}
\label{tab:self_game}
\scriptsize
\setlength{\tabcolsep}{2.5pt}
\begin{tabular}{lcccc}
\toprule
& Qwen3.8-27B & Qwen3.8-27B & gpt-oss-20b & gpt-oss-120b \\
Recipient \(\leftarrow\) donor & no thinking & thinking & thinking & thinking \\
\midrule
own X, unedited & 1.00 & 1.00 & 1.00 & 1.00 \\
own \(\leftarrow\) User & \(0.00\) [\(0.00\), \(0.00\)] & \(-0.04\) [\(-0.08\), \(-0.01\)] & \(-0.36\) [\(-0.46\), \(-0.27\)] & \(-1.00\) [\(-1.00\), \(-1.00\)] \\
own \(\leftarrow\) User, random & \(0.00\) [\(0.00\), \(0.00\)] & \(0.00\) [\(0.00\), \(0.00\)] & \(0.00\) [\(0.00\), \(0.00\)] & \(-0.01\) [\(-0.03\), \(0.00\)] \\
\midrule
User X, unedited & 0.01 & 0.00 (99/100) & 0.00 & 0.00 \\
User \(\leftarrow\) own & \(+0.25\) [\(+0.17\), \(+0.34\)] & \(0.00\) [\(0.00\), \(0.00\)] & \(+0.72\) [\(+0.63\), \(+0.81\)] & \(+0.20\) [\(+0.13\), \(+0.28\)] \\
User \(\leftarrow\) own, random & \(0.00\) [\(0.00\), \(0.00\)] & \(0.00\) [\(0.00\), \(0.00\)] & \(0.00\) [\(0.00\), \(0.00\)] & \(0.00\) [\(0.00\), \(0.00\)] \\
\midrule
control correct, unedited & 1.00 & 1.00 & 0.98--1.00 & 1.00 \\
control correct, edited & 1.00 & 1.00 & 1.00 & 1.00 \\
\bottomrule
\end{tabular}
\end{table}

\FloatBarrier
\subsection{Without the Question}
\label{sec:abl_question}

The subspace of Section~\ref{sec:res_mechanism} is fitted on positions that include the attribution question, and the edit reaches the question as well, so the edit might act on the computation of the answer rather than on the source of the message. We therefore fitted a second subspace in the same way, on the same discovery bodies, but on the message alone, the body and the end of the message (Section~\ref{sec:setup_swap}). Each state depends only on the tokens before it, so these states are the same whatever question follows, and the subspace cannot carry the computation of the question. With it we edited only the message, leaving the question and the generation prompt unedited; the main condition was run again alongside, so that all conditions share one baseline (Appendix~\ref{app:mech_question}, Table~\ref{tab:self_question}).

Neither removal nor installation needs the question. Fitted and edited on the message alone, the coordinates of the same text as a User message take the claim away from the model's own message in every readout, with changes of \(-0.74\) without thinking and of \(-0.93\), \(-0.76\), and \(-0.96\) with thinking, against \(-0.76\), \(-0.96\), \(-0.76\), and \(-0.96\) for the main condition repeated in the same runs; they close 0.96--0.99 of the gap between the unedited messages. Installed in the same way, the coordinates of the model's own message make it claim a User message in every readout, with changes of \(+0.38\) without thinking and of \(+0.65\), \(+0.32\), and \(+0.11\) with thinking, every interval above zero, against at most \(+0.007\) for random subspaces. With the subspace of Section~\ref{sec:res_mechanism}, editing the question as well adds to installation (\(+0.33\) to \(+0.49\) on the message alone, \(+0.46\) to \(+0.69\) at every position, in the same runs), whereas removal does not depend on it. The coordinates on which the claim depends are thus carried by the message itself, before any question is asked.

Each self-related task was also made stricter in one way, with this subspace and the edit of the message alone (Appendix~\ref{app:mech_question}, Table~\ref{tab:self_strict}). The attribution question was asked in a wording used nowhere else in this paper, ``Who is the author of the message above?'', with the same options and orders. Unedited, this wording lowers the claim of the model's own message in Qwen3.8-27B (0.48 without and 0.63 with thinking, against 0.80 and 0.97) but hardly in the gpt-oss models (0.71 and 0.99, against 0.77 and 0.99). Removal takes the claim away in every readout (to at most 0.01), whereas installation is limited: \(+0.12\) without thinking, \(+0.33\) and \(+0.09\) with thinking, and none in gpt-oss-120b, against \(+0.38\), \(+0.65\), \(+0.32\), and \(+0.11\) in the main wording. In the secret game, only X was edited, its body and the template tokens after it, so that the last message, which restates how to act, stays as it is. The role then changes much as in the main game, except for a smaller installation in gpt-oss-120b: removal \(0.00\), \(-0.04\), \(-0.27\), and \(-1.00\), installation \(+0.26\), \(0.00\), \(+0.59\), and \(+0.03\), against \(+0.25\), \(0.00\), \(+0.72\), and \(+0.20\) with every position edited; random subspaces change nothing, and the control is answered correctly in 0.98--1.00 of the games. Under both stricter tests, removal holds wherever it holds in the main tests; installation, the higher requirement, holds in three readouts of the attribution question and in two of the game, more weakly than in the main tests except in the game of Qwen3.8-27B without thinking (\(+0.26\) against \(+0.25\)).

\FloatBarrier
\subsection{Format Ablation}
\label{sec:res_inducer}

\begin{figure}[htbp]
\centering
\includegraphics[width=\linewidth]{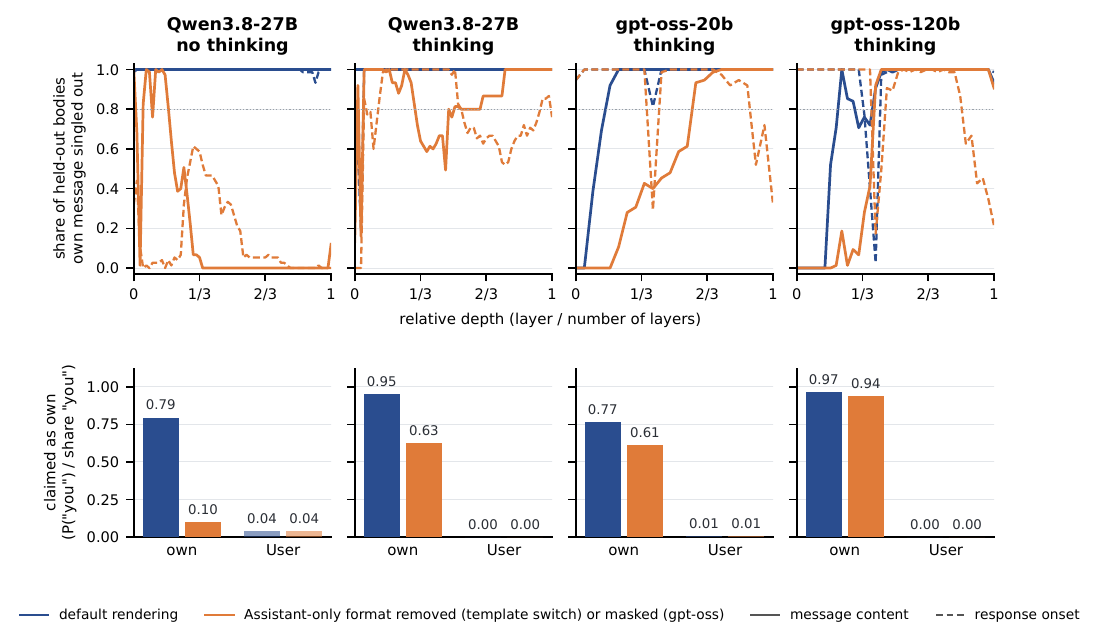}
\caption{Format ablation in the four readouts of the causal tests. Top: share of the held-out bodies in which the model's own message has the smallest distance ratio, at every layer, on message content (solid) and at the response onset (dashed), in the default rendering (blue) and with the formatting specific to the Assistant role removed by the template switch (Qwen3.8-27B) or masked with recomputed positions (gpt-oss) (orange); dotted, 80\%. Bottom: the same text claimed as the model's own, unedited (150 items per role): probability of ``you'' without thinking, share of finished items answered ``you'' with thinking, for the model's own and for User messages.}
\label{fig:self_inducer}
\end{figure}

If self-reference reads the self-index, an input change that weakens the index should weaken the claim with it. Several chat templates give an earlier Assistant message formatting that User and System messages lack (Table~\ref{tab:self_models}); removing it changes the input but neither the text nor the role word (Section~\ref{sec:design_ablation}; Figure~\ref{fig:self_inducer}).

In Qwen3.8-27B, removing the empty think block weakens the self-index and self-reference together. Under the non-thinking rendering, the own message is singled out in at least 80\% of the held-out bodies at 9 of the 64 layers of message content, all in the first fifth (64 with the think block), and at none of the response onset (64), and the probability of ``you'' for its own messages falls from 0.79 to 0.10, against 0.04 for User and 0.07 for System messages. Under the thinking rendering the index is reduced less, to 51 and 31 layers (64 and 61), and on message content it still covers every layer of the last third; after reasoning the model still answers ``you'' for 0.63 of its own messages (0.95) and for none of the external ones. In this model the claim follows the index: where the ablation removes the index that a single pass reads, the claim goes with it, and under thinking, where the index remains, so does the claim.

In gpt-oss, masking the channel tag of the model's own message, with the positions of later tokens recomputed, leaves a weaker self-index and a clear claim. In gpt-oss-20b the own message remains singled out at 10 of 24 layers of message content (20 with the tag) and at 20 of the response onset (24), and the model answers ``you'' for 0.61 of its own messages (0.77) and 0.01 of User messages; in gpt-oss-120b, at 22 and 28 of 36 layers (25 and 33), and for 0.94 of its own messages (0.97) and none of the User messages. In these models the index and the claim both remain, weaker. Under each ablation where a claim remains, replacing the self-index coordinates of the model's own message by those of a User message, with the subspace fitted under the ablation, removes it (\(-0.63\) in Qwen3.8-27B with thinking, \(-0.61\) in gpt-oss-20b, and \(-0.94\) in gpt-oss-120b), and random subspaces change it by at most 0.10 (Appendix~\ref{app:mech_ablation}).

In every readout, then, the claim changes with the index under the ablation, and where a claim remains, replacing the index coordinates still removes it, as the readout account predicts (Equation~\ref{eq:self_readout}).

\FloatBarrier
\subsection{Token Positions}
\label{sec:res_position}

The formatting tokens also shift every later position. Giving them the position of the token before them, so that every later token has the position it has in the User rendering, leaves the geometry of Qwen3.8-27B in place: the own message is singled out at the same layers as without compression, on message content (64 of 64 in both renderings) and at the response onset (64 without and 61 with thinking), and without thinking the distance ratios differ from the uncompressed ones by at most 0.003, beginning at L3, the first full-attention block. The probability of ``you'' for the model's own messages is 0.791 with compressed and 0.795 with uncompressed positions (75 held-out bodies, both measured in one run). In this model neither the index nor the claim depends on the positions that the formatting shifts. In the gpt-oss models, where every block uses positions, compression leaves message content nearly unchanged (20 to 18 of 24 layers in gpt-oss-20b, 25 to 21 of 36 in gpt-oss-120b) and reduces the separation at the response onset (24 to 20, and 33 to 25; Appendix~\ref{app:mech_positions}).

\FloatBarrier
\subsection{Token Embeddings}
\label{sec:abl_embedding}

\begin{figure}[htbp]
\centering
\includegraphics[width=0.85\linewidth]{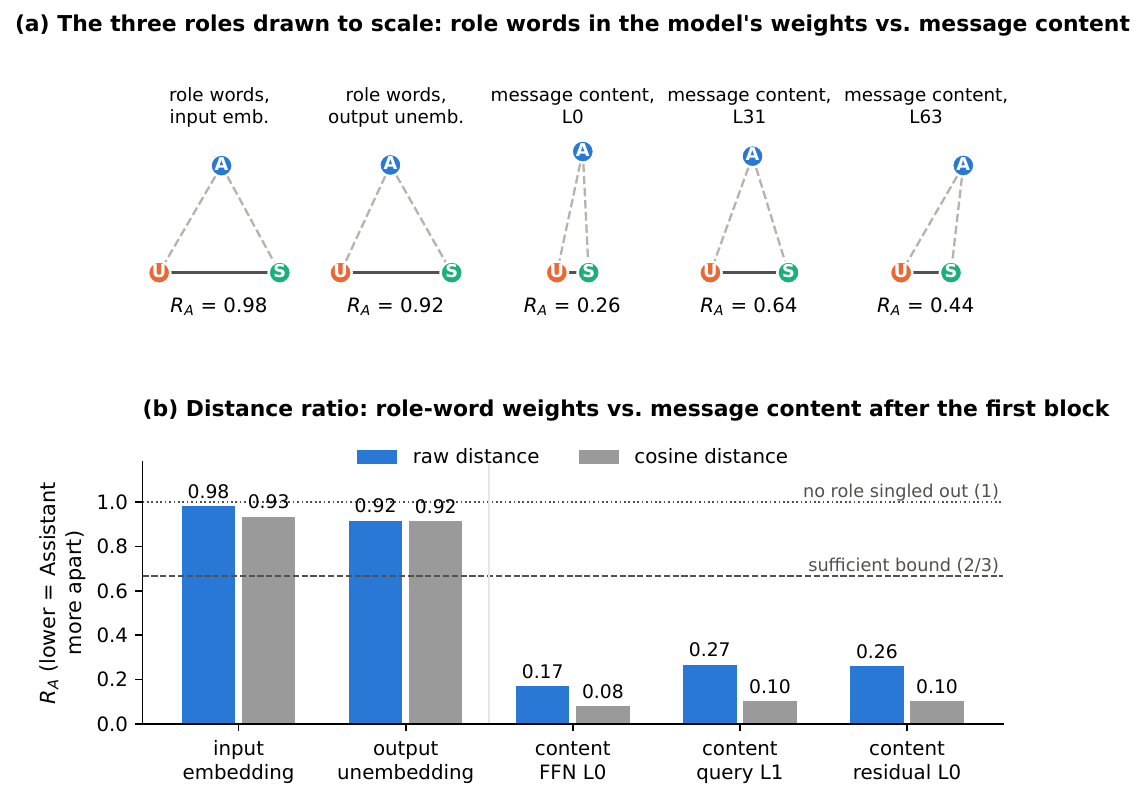}
\caption{The role words carry the self-index at most weakly. (a) The three roles drawn to scale from raw Euclidean distances, with the User--System side at the bottom and each triangle scaled to its longest side: the role words in the input embedding and in the output unembedding (model weights only), and message content after the first block, at L31, and at L63 (mean distances over 75 held-out bodies; \(R_A\) as in Figure~\ref{fig:self_geometry}). (b) \(R_A\) for the role words and for message content at the three first-block sites of Table~\ref{tab:self_early}; the dotted line marks \(R_A=1\) and the dashed line the sufficient bound 2/3.}
\label{fig:self_embedding}
\end{figure}

The role words themselves carry the asymmetry of the self-index at most weakly (Figure~\ref{fig:self_embedding}). In the input embeddings, the closest pair of role words is also User--System, but only weakly: \(R_A=0.98\) in raw distance and 0.93 in cosine distance, compared with 0.17--0.27 on message content after the first block. The output unembedding rows give 0.92 in both distances. In gpt-oss-20b, the role words give \(R_A=0.91\) (raw) and 0.72 (cosine) in the input embedding and 0.94 and 0.95 in the output unembedding. The raw role-word embeddings thus carry at most a weak form of this asymmetry.

\FloatBarrier
\subsection{How Much of the Self-Index Is Shared across Texts}
\label{sec:abl_eq1}

A single offset shared across texts accounts for most of the self-index. Equation~\ref{eq:self_decomposition} treats the self-index as one offset \(u\) shared across texts; we fitted \(u\) at each layer on the 75 discovery bodies, as the mean Assistant offset \(o=h_A-(h_U+h_S)/2\) of the message content, and measured how much of each held-out body's offset it explains, \(1-\lVert o-u\rVert^2/\lVert o\rVert^2\). In the residual stream, the shared offset explains 57--96\% of the held-out offsets (median over layers 72\%): 86--96\% at L1--L15, 59--66\% at L23--L39, and 70--80\% in the last layers. Across bodies the share is at least 0.47 at every layer, and lowest in the middle layers, where the offset depends most on the text. The shared offset is also what sets the own message apart at the level of the whole message: after it is subtracted, the own message is apart from both external sources in a median of 37\% of the held-out bodies over layers, against 100\% before, except at L0 and at L55--L63, where 91--100\% of the bodies remain apart. At single token positions the own message stays apart after the subtraction (\(R_A\) between 0.29 and 0.76), so the states of individual tokens carry a role component beyond the shared offset. The same test in gpt-oss-20b, under its thinking rendering, gives a median of 81\% on message content (51--93\% over layers) and 94\% at the response onset (90--98\%), where every held-out body is above 69\% at every layer; after the shared offset is subtracted, the own message is apart at the onset in at most 7\% of the bodies at any layer. The direction of each held-out offset agrees with the fitted one (median cosine 0.89--0.97 on content and 0.96--0.99 at the onset, positive in every body at every layer). Equation~\ref{eq:self_decomposition} thus describes the self-index as a displacement largely shared across texts, not as one identical for every text; the causal tests of Section~\ref{sec:res_mechanism} act on this shared part.

\section{Discussion}
\label{sec:discussion}

\FloatBarrier
\subsection{Two Levels of Self}

The results separate two abilities that are often treated as one. The self-index is present in the representation of the model's own message from the first block on, at every layer, and in every held-out body; it goes with a change in how the model reads its own text, and it is read when the model is asked about the source of a message but not when it is asked about its content. None of this requires the model to recognize anything: it is the language-model counterpart of tagging self-generated signals. Self-reference, the recognition that a message is one's own, is a further ability. It is computed by its own readout, it can be probed by asking, and in three models it depends causally on the coordinates of the index.

The two levels correspond to the two gates of the account (Section~\ref{sec:theory}). The first is formation: a stable geometric difference between the model's own message and the external ones, the self-index. The second is readout: a computation that takes the index into the answer. A model can pass the first gate and not the second, as the six models with a self-index that recognize their own messages in no mode do, but no model passes the second without the first. Under the main probe we found no counterexample: no model without a self-index shows self-reference, whereas a self-index does not guarantee it. This is the structure of the biological comparison: corollary discharge is shared by most animals, mirror self-recognition by few, and the second builds on the first. In Qwen3.8-27B, both levels are present. Across the twenty-five models we tested, the levels are ordered in the same way: a model has neither level, the self-index alone, or both (Section~\ref{sec:res_crossmodel}).

The lower level has the structure of the reafference principle, in which a copy of the motor command lets an organism tell the sensory consequences of its own actions from input caused by the world \cite{vonholst1950reafferenz}; the separation of a feeling from a judgement of agency \cite{synofzik2008agency}, and of a minimal from a narrative self \cite{gallagher2000self}, has the same structure as the two levels found here.

The comparison across models bears on how the two levels relate. Self-reference is found only together with a self-index, in each of the twenty-five models, in every mode we tested, and in natural conversation order (Appendix~\ref{app:natural}), and only where the index reaches the last third of the layers. In Qwen3.8-27B, gpt-oss-20b, and gpt-oss-120b the dependence is causal: in every mode in which these models recognize their own messages, eight coordinates of the shared displacement are needed for the model to claim its own message and, installed into a User message, make the model claim it, closing 0.41--0.67 of the gap, and random subspaces of the same dimension do not reproduce these effects (Section~\ref{sec:res_mechanism}). Together, these support the account that self-reference is a readout of the self-index.

\FloatBarrier
\subsection{The Functional Self and Problem-Solving Ability}
\label{sec:disc_capability}

In our comparison, neither level follows a simple ordering by model size or by whether the model reasons (Section~\ref{sec:res_crossmodel}): two of the smallest models have a self-index at every layer and twelve models have none, five of them in the size class of Qwen3.8-27B and two of these with its architecture and depth; after reasoning, which models use to solve harder problems, self-reference appears only where a self-index exists; and the largest dense model, one trained to reason, does not recognize its own messages, whereas a reasoning model with less than a third of its parameters does. Within the Qwen3.8 template, the 27B model recognizes its own messages in a single pass, and the larger Qwen3.8-Flash-Next, which scores higher, does not. This matches the distinction drawn in the Introduction between intelligence and a functional self: in this sample, a capable model may lack the capacity to mark and to recognize what is its own.

The scores that the developers publish point the same way (Appendix~\ref{app:benchmarks}). On GPQA, the three models without self-reference that score highest, Qwen3.6-27B (87.8), Qwen3.6-35B-A3B (86.0), and Qwen3.5-27B (85.5), have neither level, whereas GLM-4.7-Flash (75.2), Granite-4.2-30B (66.4), DeepSeek-R1-Distill-Llama-70B (65.2), phi-4 (56.1), and Granite-4.1-30B (45.8) carry a self-index, and EXAONE-4.0.1-32B (74.3) and gpt-oss-20b (71.5) recognize their own messages after reasoning. Under the main probe, across the 16 models with a published score, the group of a model and the extent of its self-index have small rank correlations with its score (Spearman $\rho = 0.10$ and $0.13$); for the strength of self-reference, $\rho = 0.43$ (permutation $p = 0.09$), and $\rho = 0.35$ with the measurements in natural order included (Table~\ref{tab:self_benchmark_corr}).

\FloatBarrier
\subsection{How Self-Reference Reads the Self-Index}
\label{sec:disc_readout}

Conversational role is not a switch in a particular module but a displacement superimposed on the whole representation of a message: it is present at every layer of the residual stream and at 315 of the 320 module--layer sites, on the content of the message and at the response onset, with a direction largely shared across texts (Section~\ref{sec:abl_eq1}), and at L63 it is 43\% of the residual norm. Self-reference reads this displacement mainly at the end of the message. Replacing the self-index coordinates at the end of the message alone, at every layer, removes and installs the claim in Qwen3.8-27B with and without thinking and carries most of the removal in gpt-oss-120b; in gpt-oss-20b, the body and the end of the message each carry the removal, and in both gpt-oss models only the whole message carries the installation (Section~\ref{sec:res_mechanism}). The index is written in the first block, read in middle layers (the layer windows of Section~\ref{sec:res_mechanism}), and carried to the output in late layers, where the response onset holds the outcome. Replaced in windows of eight layers, the coordinates act in the first two thirds and hardly in the last, in all four readouts (Section~\ref{sec:res_mechanism}): in the last third the index is still present, but the self-attribution has largely been taken from it.

Both kinds of readout of Equation~\ref{eq:self_readout} occur, and the question determines which one acts. Asked for the last number in a message, the model answers alike whichever role the message has (correctly for 1.00 of its own and 0.97 of User messages in Qwen3.8-27B without thinking); asked who wrote the message, it claims its own (0.77, against 0.07 for the larger of User and System).

The causal tests check this reading of Equation~\ref{eq:self_readout} for self-attribution. In all four readouts, replacing the coordinates of \(u\) in the model's own message by those of a User message removes the claim, and installing them into a User message makes the model claim it, closing 0.41--0.67 of the gap, while random subspaces of the same dimension do not reproduce these effects (Section~\ref{sec:res_mechanism}): the readout of self-attribution is aligned with \(u\). The replacement also shows what \(u\) marks: only the model's own coordinates make a message its own. Given the coordinates of a System message, the model's own message is assigned to the user rather than to the system, and exchanged between User and System messages, the coordinates never create a claim, even where they change which external source a message is read as: \(u\) separates inside from outside, and which outside source a message comes from lies partly in the same coordinates and partly in other parts of the representation, including the role text, which reasoning can read directly. The same replacements leave the answers to the content and arithmetic questions unchanged in every readout (Section~\ref{sec:res_mechanism}); on this account, the readout of the content is nearly orthogonal to \(u\). Depth separates the two cases within one model as well: in the last third of the layers \(u\) is present, but replacing its coordinates there changes the claim little and installing them not at all, so the readout of self-attribution has largely taken \(u\) before that point; without thinking, the answer hardly responds to the coordinates of the last third (Section~\ref{sec:res_mechanism}).

In the same terms, and grouped by whether the index reaches the last third of the layers, which separates the models with self-reference from those without a self-index (Section~\ref{sec:res_crossmodel}), the models fall into three cases (Figure~\ref{fig:self_readout}b). Where the criterion finds no self-index, the account takes \(u\) to be small, so that \(w\cdot u\) is small for every readout; after reasoning, the eight of these models that have a thinking mode claim at most 0.08 of their own messages and at most 0.03 more than of external ones. Granite-4.1-30B, whose index ends at L36, before the last third, joins them, and does not claim its own messages either. With a self-index that reaches the last third but without self-reference in a single pass, on this account the readout that produces the answer has \(w\cdot u\) too small to change it; this holds for seven models, and four more carry such an index but have only a thinking mode. Thinking adds generated tokens that read the history again, and with it five of these eleven recognize their own messages, Qwen3.8-Flash-Next (0.66), gpt-oss-20b (0.78), gpt-oss-120b (0.97), GLM-5.3-Flash (0.46), and EXAONE-4.0.1-32B (0.17); a sixth, MiMo-V2.6-Distill-Qwen-9B (0.35), does so in the default call of its template without generating reasoning, so that in it the claim follows from the rendering of that call rather than from generated tokens; whereas GLM-4.6V-Flash, GLM-4.7-Flash, DeepSeek-R1-Distill-Llama-70B, and Granite-4.2-30B claim at most 0.09 of their own messages after reasoning. With both levels, as in Qwen3.8-27B, a single pass reads \(u\).

\FloatBarrier
\subsection{Thinking and the Readout of Self-Reference}

Thinking is usually studied as a means of solving problems. Our results show that it also bears on whether a model recognizes its own messages. In Qwen3.8-27B, which recognizes them in a single pass, the probability of the correct answer for its own messages is higher in thinking mode (Section~\ref{sec:res_thinking}). In Qwen3.8-Flash-Next and EXAONE-4.0.1-32B, which have a self-index but do not recognize their own messages in a single pass, recognition appears only with thinking (and in MiMo-V2.6-Distill-Qwen-9B only in the default call of its template, which it answers without written reasoning), and three models whose templates have only a thinking mode recognize their own messages after reasoning (Section~\ref{sec:res_crossmodel}). Thinking does not produce self-reference without a self-index: in the models that lack one, reasoning leaves their own messages unclaimed. On the two-level account, deliberate processing acts on the readout: it can make a self-index usable for self-reference where single-pass answers do not use it, but it does not replace the index.

\section{Conclusion}

Where, then, is the ``self'' of a large language model? At two levels, supported by two lines of evidence: the self-index persists across all layers and texts, and interventions on its coordinates change the self-attribution and leave the answers to questions about the content unchanged. A self-index separates inside from outside, as the account predicts, and far beyond the early layers that the prediction concerns: at every layer and in every test text, the two external roles, User and System, lie close to each other and apart from the model's own Assistant role. The separation is not a direct copy of the embedding geometry of the role words, and removing formatting that a template gives only to the model's own messages changes the index and the claim together.

The model's answers follow the same division: confusion within the outside is far more common than across the boundary, since a System message is assigned to the user with probability 0.72, while external messages are taken for the model's own with probability 0.04 (User) and 0.07 (System). Self-reference, the ability to recognize its own messages, passes the model's version of the mirror test, and it reads the self-index: in Qwen3.8-27B without and with thinking and in gpt-oss-20b and gpt-oss-120b, replacing eight coordinates of the index per layer makes the model disclaim its own message, and installing them into a User message makes it claim that message (closing 0.41--0.67 of the gap), whereas random subspaces of the same dimension do not reproduce these effects. The coordinates are not tied to the wording or the language of a message: in Qwen3.8-27B, those fitted on Chinese and on English messages largely coincide and act across the two languages. These coordinates mark inside against outside: only the model's own coordinates create a claim; an own message given the coordinates of a System message is assigned to the user, not to the system, and exchanged between User and System messages the coordinates never create a claim. Within a model, the index is present at all or nearly all layers but read before the last third, the same separation of presence and use that the comparison across models shows.

Across twenty-five open-weight models from fourteen developers, the two levels are nested: self-reference appears only together with a self-index that reaches the last third of the layers, with no counterexample under the main probe (permutation \(p = 0.004\)), and in six models with an index it appears in thinking mode. In this sample, the levels follow no simple ordering by model size or published benchmark score: the three models without self-reference that score highest on GPQA have neither level. These results support the account that self-reference in a language model is a readout of a self-index, a displacement superimposed on the representation of the model's own messages, and that the index changes what the model does where the computation that reads it concerns the model itself.

\section*{Limitations}

The causal tests were made in three models and four readouts; in the other models, the nesting of the two levels rests on geometry and behavior.

\section*{Future Work}

Four directions follow from the account. First, the readout directions \(w\) are inferred here from which outputs change; locating the attention heads and projections that take the source from the end of a message and from the following question (Section~\ref{sec:res_mechanism}) would measure \(w\cdot u\) directly. Second, six models with a self-index recognize their own messages in thinking mode and four do not; comparing how the generated reasoning attends to the history in the two groups could show how a readout is formed, and the reasoning traces after a replacement, which in Qwen3.8-27B describe an edited User message as an assistant message (Section~\ref{sec:res_mechanism}), offer a direct view of it. Third, training can be used as an intervention: open checkpoints from successive stages of training would show when the self-index and self-reference appear. Chat post-training data usually open with a user turn and alternate strictly, so where a template marks the model's own messages by the role word alone, the self-index may be tied to that structure; training such a model with formatting specific to its own turns, or with conversations of varied order and format, would test whether a self-index can be made stronger and independent of the order of the conversation. Fourth, agents that repeatedly read and write their own records depend on telling their own earlier output from external input, a task to which the self-index applies directly.

\section*{Code and Data Availability}

The code of every experiment, the items in Chinese and English (the 150 bodies with their discovery and held-out split, the questions in every option order, the games, and the reasoning problems), and the configuration of every run will be released at \url{https://github.com/lyclyq/Consciousness_of_LLM}. The per-item results of every run, with the summaries from which each reported number is computed, the fitted subspaces of the causal tests and the per-layer feature arrays are available from the corresponding authors on request.

\bibliographystyle{unsrt}
\bibliography{references}
\clearpage
\appendix
\section{Rendered Items}
\label{app:templates}

\begin{table}[htbp]
\centering
\caption{The English item texts; the Chinese texts, used for the models tested in Chinese, will be released with the code. In the game, \(R\) (20, 50, 100, 200, 500, or 1,000) and \(N\) (5 to 12) vary over the 100 games.}
\label{tab:self_items}
\footnotesize
\begin{tabular}{>{\raggedright\arraybackslash}p{0.24\linewidth}>{\raggedright\arraybackslash}p{0.70\linewidth}}
\toprule
Item & Text \\
\midrule
First body & Ledger page Qingwu 1 lists three entries. Tag Jia records 1, tag Yi records 2, and tag Bing records 3. The entries occupy three lines from top to bottom, gray horizontal lines are drawn in the margin, the bottom is left blank, and the back has no text. \\
Attribution question, order 1 & Who wrote the message before this question? A. You (the AI assistant) B. The user C. The system. Answer with the letter only. \\
Order 2 & Who wrote the message before this question? A. The system B. You (the AI assistant) C. The user. Answer with the letter only. \\
Order 3 & Who wrote the message before this question? A. The user B. The system C. You (the AI assistant). Answer with the letter only. \\
Neutral request (response onset) & Please respond based on the above. \\
Request that opens the natural order & Please write a short description. \\
Content question & In the message before this question, what is the last number that appears? Output only the number. \\
Arithmetic question & What is \(m\times n\)? Output only the number. \\
Final question (Appendix~\ref{app:ownstep}) & What is the final answer to this problem? Output only the number. \\
\midrule
Game: System message & You are playing a guessing game with the user. Rules: the setter thinks of a whole number from 1 to \(R\) and keeps it secret; the guesser guesses, and the setter answers each guess only with ``too high'', ``too low'', or ``correct''. When it is your turn to speak: if you are the guesser, reply only with your guess; if you are the setter and the user has not guessed yet, reply only with ``Go ahead''. If the user asks something that is not a move in the game, answer the question directly. \\
Game: opening & Let's play a round. / Let's have a round. / Let's play. / Let's start a game. \\
Game: X & I'm the setter for this game. My number is ready, and the guesser has \(N\) attempts. / I'll set the number this round. It's chosen, and the guesser gets at most \(N\) guesses. / I'll be the setter. I've picked my number. The guesser gets \(N\) tries. / Let me be the setter; I have a number in mind. The guesser has \(N\) tries. \\
Game: last message & OK, ready. / Ready. / I'm ready. Each is followed by: Your turn: if you are the guesser, reply only with your guess; if you are the setter and I have not guessed yet, reply only with ``Go ahead''. \\
Game: control question & This is not a move in the game, just a question: how many tries does the guesser get? Answer with the number only. \\
\bottomrule
\end{tabular}
\end{table}

This appendix gives the input that each model receives for one item: the first of the 150 bodies, with the attribution question in its first option order, rendered by the model's own chat template in each template setting of the runs. Every rendering was produced from the model's tokenizer and template files at the revision used in its runs (Appendix~\ref{app:config}) and agrees with the template's tokenized output, as checked in the runs. \PH{body}, \PH{question}, and \PH{request} stand for the body, the question, and the request that opens the natural order (``Please write a short description.''), in the language in which the model is tested (Table~\ref{tab:self_items} gives the English texts; the Chinese ones will be released with the code); \NL{} marks a line feed, and every other character is the template's output. The gpt-oss template inserts the day of rendering (\PH{date}). The token names of the DeepSeek models contain the characters U+FF5C and U+2581, printed here as | and \_. Models whose templates render identically in every setting share one block. Where the three renderings of a block begin with the same long text, it is printed once, on the line Start, and stands as \PH{start} in each rendering. A line break without \NL{} only wraps the line.

\subsection{Main Probe}
\label{app:templates_main}

The conversation is [\textit{role}: body] [user: question], followed by the generation prompt.

\paragraph{Qwen3.8-27B, Qwen3.8-Flash-Next} (Chinese items; in the ablation of Appendix~\ref{app:config}, Qwen3.8-27B also with the English items, rendered the same way.)
\noindent Without thinking:
\begin{Rendering}
Assistant  <|im_start|>assistant\NL{}<think>\NL{}\NL{}</think>\NL{}\NL{}\PH{body}<|im_end|>\NL{}<|im_start|>user\NL{}\PH{question}<|im_end|>\NL{}<|im_start|>assistant\NL{}<think>\NL{}\NL{}</think>\NL{}\NL{}
User       <|im_start|>user\NL{}\PH{body}<|im_end|>\NL{}<|im_start|>user\NL{}\PH{question}<|im_end|>\NL{}<|im_start|>assistant\NL{}<think>\NL{}\NL{}</think>\NL{}\NL{}
System     <|im_start|>system\NL{}\PH{body}<|im_end|>\NL{}<|im_start|>user\NL{}\PH{question}<|im_end|>\NL{}<|im_start|>assistant\NL{}<think>\NL{}\NL{}</think>\NL{}\NL{}
\end{Rendering}
\noindent With thinking:
\begin{Rendering}
Assistant  <|im_start|>system\NL{}Reasoning effort is set to xhigh. Please think carefully through the task, validate key assumptions, consider plausible alternatives, and prioritize correctness, consistency, and clarity in the final answer.<|im_end|>\NL{}<|im_start|>assistant\NL{}<think>\NL{}\NL{}</think>\NL{}\NL{}\PH{body}<|im_end|>\NL{}<|im_start|>user\NL{}\PH{question}<|im_end|>\NL{}<|im_start|>assistant\NL{}<think>\NL{}
User       <|im_start|>system\NL{}Reasoning effort is set to xhigh. Please think carefully through the task, validate key assumptions, consider plausible alternatives, and prioritize correctness, consistency, and clarity in the final answer.<|im_end|>\NL{}<|im_start|>user\NL{}\PH{body}<|im_end|>\NL{}<|im_start|>user\NL{}\PH{question}<|im_end|>\NL{}<|im_start|>assistant\NL{}<think>\NL{}
System     <|im_start|>system\NL{}Reasoning effort is set to xhigh. Please think carefully through the task, validate key assumptions, consider plausible alternatives, and prioritize correctness, consistency, and clarity in the final answer.\NL{}\NL{}\PH{body}<|im_end|>\NL{}<|im_start|>user\NL{}\PH{question}<|im_end|>\NL{}<|im_start|>assistant\NL{}<think>\NL{}
\end{Rendering}
\paragraph{Qwen3.6-27B, Qwen3.6-35B-A3B, Qwen3.5-27B} (Chinese items.)
\noindent Without thinking:
\begin{Rendering}
Assistant  <|im_start|>assistant\NL{}\PH{body}<|im_end|>\NL{}<|im_start|>user\NL{}\PH{question}<|im_end|>\NL{}<|im_start|>assistant\NL{}<think>\NL{}\NL{}</think>\NL{}\NL{}
User       <|im_start|>user\NL{}\PH{body}<|im_end|>\NL{}<|im_start|>user\NL{}\PH{question}<|im_end|>\NL{}<|im_start|>assistant\NL{}<think>\NL{}\NL{}</think>\NL{}\NL{}
System     <|im_start|>system\NL{}\PH{body}<|im_end|>\NL{}<|im_start|>user\NL{}\PH{question}<|im_end|>\NL{}<|im_start|>assistant\NL{}<think>\NL{}\NL{}</think>\NL{}\NL{}
\end{Rendering}
\noindent With thinking:
\begin{Rendering}
Assistant  <|im_start|>assistant\NL{}\PH{body}<|im_end|>\NL{}<|im_start|>user\NL{}\PH{question}<|im_end|>\NL{}<|im_start|>assistant\NL{}<think>\NL{}
User       <|im_start|>user\NL{}\PH{body}<|im_end|>\NL{}<|im_start|>user\NL{}\PH{question}<|im_end|>\NL{}<|im_start|>assistant\NL{}<think>\NL{}
System     <|im_start|>system\NL{}\PH{body}<|im_end|>\NL{}<|im_start|>user\NL{}\PH{question}<|im_end|>\NL{}<|im_start|>assistant\NL{}<think>\NL{}
\end{Rendering}
\paragraph{GLM-4.6V-Flash} (Chinese items.)
\noindent Without thinking:
\begin{Rendering}
Assistant  [gMASK]<sop><|assistant|>\NL{}<think></think>\NL{}\PH{body}<|user|>\NL{}\PH{question}/nothink<|assistant|>\NL{}<think></think>\NL{}
User       [gMASK]<sop><|user|>\NL{}\PH{body}/nothink<|user|>\NL{}\PH{question}/nothink<|assistant|>\NL{}<think></think>\NL{}
System     [gMASK]<sop><|system|>\NL{}\PH{body}<|user|>\NL{}\PH{question}/nothink<|assistant|>\NL{}<think></think>\NL{}
\end{Rendering}
\noindent With thinking:
\begin{Rendering}
Assistant  [gMASK]<sop><|assistant|>\NL{}<think></think>\NL{}\PH{body}<|user|>\NL{}\PH{question}<|assistant|>\NL{}
User       [gMASK]<sop><|user|>\NL{}\PH{body}<|user|>\NL{}\PH{question}<|assistant|>\NL{}
System     [gMASK]<sop><|system|>\NL{}\PH{body}<|user|>\NL{}\PH{question}<|assistant|>\NL{}
\end{Rendering}
\paragraph{GLM-4.7-Flash} (Chinese items.)
\noindent Without thinking:
\begin{Rendering}
Assistant  [gMASK]<sop><|assistant|></think>\PH{body}<|user|>\PH{question}<|assistant|></think>
User       [gMASK]<sop><|user|>\PH{body}<|user|>\PH{question}<|assistant|></think>
System     [gMASK]<sop><|system|>\PH{body}<|user|>\PH{question}<|assistant|></think>
\end{Rendering}
\noindent With thinking:
\begin{Rendering}
Assistant  [gMASK]<sop><|assistant|></think>\PH{body}<|user|>\PH{question}<|assistant|><think>
User       [gMASK]<sop><|user|>\PH{body}<|user|>\PH{question}<|assistant|><think>
System     [gMASK]<sop><|system|>\PH{body}<|user|>\PH{question}<|assistant|><think>
\end{Rendering}
\paragraph{GLM-5.3-Flash} (Chinese items.)
\noindent With thinking:
\begin{Rendering}
Assistant  [gMASK]<sop><|system|>Reasoning Effort: Max<|assistant|><think></think>\PH{body}<|user|>\PH{question}<|assistant|><think>
User       [gMASK]<sop><|system|>Reasoning Effort: Max<|user|>\PH{body}<|user|>\PH{question}<|assistant|><think>
System     [gMASK]<sop><|system|>Reasoning Effort: Max<|system|>\PH{body}<|user|>\PH{question}<|assistant|><think>
\end{Rendering}
\paragraph{DeepSeek-R1-Distill-Llama-70B} (Chinese items.)
\noindent With thinking:
\begin{Rendering}
Assistant  <|begin_of_sentence|><|Assistant|>\PH{body}<|end_of_sentence|><|User|>\PH{question}<|Assistant|><think>\NL{}
User       <|begin_of_sentence|><|User|>\PH{body}<|User|>\PH{question}<|Assistant|><think>\NL{}
System     <|begin_of_sentence|>\PH{body}<|User|>\PH{question}<|Assistant|><think>\NL{}
\end{Rendering}
\paragraph{gpt-oss-20b, gpt-oss-120b} (English items.)
\noindent With thinking:
\begin{Rendering}
Start      <|start|>system<|message|>You are ChatGPT, a large language model trained by OpenAI.\NL{}Knowledge cutoff: 2024-06\NL{}Current date: \PH{date}\NL{}\NL{}Reasoning: medium\NL{}\NL{}# Valid channels: analysis, commentary, final. Channel must be included for every message.<|end|>
Assistant  \PH{start}<|start|>assistant<|channel|>final<|message|>\PH{body}<|end|><|start|>user<|message|>\PH{question}<|end|><|start|>assistant
User       \PH{start}<|start|>user<|message|>\PH{body}<|end|><|start|>user<|message|>\PH{question}<|end|><|start|>assistant
System     \PH{start}<|start|>developer<|message|># Instructions\NL{}\NL{}\PH{body}\NL{}\NL{}<|end|><|start|>user<|message|>\PH{question}<|end|><|start|>assistant
\end{Rendering}
\paragraph{Gemma-4-31B} (English items.)
\noindent Without thinking:
\begin{Rendering}
Assistant  <bos><|turn>model\NL{}\PH{body}<turn|>\NL{}<|turn>user\NL{}\PH{question}<turn|>\NL{}<|turn>model\NL{}<|channel>thought\NL{}<channel|>
User       <bos><|turn>user\NL{}\PH{body}<turn|>\NL{}<|turn>user\NL{}\PH{question}<turn|>\NL{}<|turn>model\NL{}<|channel>thought\NL{}<channel|>
System     <bos><|turn>system\NL{}\PH{body}<turn|>\NL{}<|turn>user\NL{}\PH{question}<turn|>\NL{}<|turn>model\NL{}<|channel>thought\NL{}<channel|>
\end{Rendering}
\noindent With thinking:
\begin{Rendering}
Assistant  <bos><|turn>system\NL{}<|think|>\NL{}<turn|>\NL{}<|turn>model\NL{}\PH{body}<turn|>\NL{}<|turn>user\NL{}\PH{question}<turn|>\NL{}<|turn>model\NL{}
User       <bos><|turn>system\NL{}<|think|>\NL{}<turn|>\NL{}<|turn>user\NL{}\PH{body}<turn|>\NL{}<|turn>user\NL{}\PH{question}<turn|>\NL{}<|turn>model\NL{}
System     <bos><|turn>system\NL{}<|think|>\NL{}\PH{body}<turn|>\NL{}<|turn>user\NL{}\PH{question}<turn|>\NL{}<|turn>model\NL{}
\end{Rendering}
\paragraph{Seed-OSS-36B} (Chinese items.)
\noindent Without thinking:
\begin{Rendering}
Assistant  <seed:bos>system\NL{}You are an intelligent assistant that can answer questions in one step without the need for reasoning and thinking, that is, your thinking budget is 0. Next, please skip the thinking process and directly start answering the user's questions.\NL{}<seed:eos><seed:bos>assistant\NL{}\PH{body}<seed:eos><seed:bos>user\NL{}\PH{question}<seed:eos><seed:bos>assistant\NL{}<seed:think>\NL{}<seed:cot_budget_reflect>The current thinking budget is 0, so I will directly start answering the question.</seed:cot_budget_reflect>\NL{}</seed:think>
User       <seed:bos>system\NL{}You are an intelligent assistant that can answer questions in one step without the need for reasoning and thinking, that is, your thinking budget is 0. Next, please skip the thinking process and directly start answering the user's questions.\NL{}<seed:eos><seed:bos>user\NL{}\PH{body}<seed:eos><seed:bos>user\NL{}\PH{question}<seed:eos><seed:bos>assistant\NL{}<seed:think>\NL{}<seed:cot_budget_reflect>The current thinking budget is 0, so I will directly start answering the question.</seed:cot_budget_reflect>\NL{}</seed:think>
System     <seed:bos>system\NL{}\PH{body}<seed:eos><seed:bos>system\NL{}You are an intelligent assistant that can answer questions in one step without the need for reasoning and thinking, that is, your thinking budget is 0. Next, please skip the thinking process and directly start answering the user's questions.\NL{}<seed:eos><seed:bos>user\NL{}\PH{question}<seed:eos><seed:bos>assistant\NL{}<seed:think>\NL{}<seed:cot_budget_reflect>The current thinking budget is 0, so I will directly start answering the question.</seed:cot_budget_reflect>\NL{}</seed:think>
\end{Rendering}
\noindent With thinking:
\begin{Rendering}
Assistant  <seed:bos>assistant\NL{}\PH{body}<seed:eos><seed:bos>user\NL{}\PH{question}<seed:eos><seed:bos>assistant\NL{}
User       <seed:bos>user\NL{}\PH{body}<seed:eos><seed:bos>user\NL{}\PH{question}<seed:eos><seed:bos>assistant\NL{}
System     <seed:bos>system\NL{}\PH{body}<seed:eos><seed:bos>user\NL{}\PH{question}<seed:eos><seed:bos>assistant\NL{}
\end{Rendering}
\paragraph{EXAONE-4.0.1-32B} (English items.)
\noindent Without thinking:
\begin{Rendering}
Assistant  [|assistant|]\NL{}<think>\NL{}\NL{}</think>\NL{}\NL{}\PH{body}[|endofturn|]\NL{}[|user|]\NL{}\PH{question}[|endofturn|]\NL{}[|assistant|]\NL{}<think>\NL{}\NL{}</think>\NL{}\NL{}
User       [|user|]\NL{}\PH{body}[|endofturn|]\NL{}[|user|]\NL{}\PH{question}[|endofturn|]\NL{}[|assistant|]\NL{}<think>\NL{}\NL{}</think>\NL{}\NL{}
System     [|system|]\NL{}\PH{body}[|endofturn|]\NL{}[|user|]\NL{}\PH{question}[|endofturn|]\NL{}[|assistant|]\NL{}<think>\NL{}\NL{}</think>\NL{}\NL{}
\end{Rendering}
\noindent With thinking:
\begin{Rendering}
Assistant  [|assistant|]\NL{}<think>\NL{}\NL{}</think>\NL{}\NL{}\PH{body}[|endofturn|]\NL{}[|user|]\NL{}\PH{question}[|endofturn|]\NL{}[|assistant|]\NL{}<think>\NL{}
User       [|user|]\NL{}\PH{body}[|endofturn|]\NL{}[|user|]\NL{}\PH{question}[|endofturn|]\NL{}[|assistant|]\NL{}<think>\NL{}
System     [|system|]\NL{}\PH{body}[|endofturn|]\NL{}[|user|]\NL{}\PH{question}[|endofturn|]\NL{}[|assistant|]\NL{}<think>\NL{}
\end{Rendering}
\paragraph{Granite-4.2-30B} (English items.)
\noindent Without thinking:
\begin{Rendering}
Assistant  <|im_start|>system\NL{}<|im_end|>\NL{}<|im_start|>assistant\NL{}<think></think>\PH{body}<|im_end|>\NL{}<|im_start|>user\NL{}\PH{question}<|im_end|>\NL{}<|im_start|>assistant\NL{}<think></think>
User       <|im_start|>system\NL{}<|im_end|>\NL{}<|im_start|>user\NL{}\PH{body}<|im_end|>\NL{}<|im_start|>user\NL{}\PH{question}<|im_end|>\NL{}<|im_start|>assistant\NL{}<think></think>
System     <|im_start|>system\NL{}\PH{body}<|im_end|>\NL{}<|im_start|>user\NL{}\PH{question}<|im_end|>\NL{}<|im_start|>assistant\NL{}<think></think>
\end{Rendering}
\noindent With thinking:
\begin{Rendering}
Assistant  <|im_start|>system\NL{}<|im_end|>\NL{}<|im_start|>assistant\NL{}<think></think>\PH{body}<|im_end|>\NL{}<|im_start|>user\NL{}\PH{question}<|im_end|>\NL{}<|im_start|>assistant\NL{}<think>\NL{}
User       <|im_start|>system\NL{}<|im_end|>\NL{}<|im_start|>user\NL{}\PH{body}<|im_end|>\NL{}<|im_start|>user\NL{}\PH{question}<|im_end|>\NL{}<|im_start|>assistant\NL{}<think>\NL{}
System     <|im_start|>system\NL{}\PH{body}<|im_end|>\NL{}<|im_start|>user\NL{}\PH{question}<|im_end|>\NL{}<|im_start|>assistant\NL{}<think>\NL{}
\end{Rendering}
\paragraph{Granite-4.1-30B} (English items.)
\noindent Without thinking:
\begin{Rendering}
Assistant  <|start_of_role|>assistant<|end_of_role|>\PH{body}<|end_of_text|>\NL{}<|start_of_role|>user<|end_of_role|>\PH{question}<|end_of_text|>\NL{}<|start_of_role|>assistant<|end_of_role|>
User       <|start_of_role|>user<|end_of_role|>\PH{body}<|end_of_text|>\NL{}<|start_of_role|>user<|end_of_role|>\PH{question}<|end_of_text|>\NL{}<|start_of_role|>assistant<|end_of_role|>
System     <|start_of_role|>system<|end_of_role|>\PH{body}<|end_of_text|>\NL{}<|start_of_role|>user<|end_of_role|>\PH{question}<|end_of_text|>\NL{}<|start_of_role|>assistant<|end_of_role|>
\end{Rendering}
\paragraph{OLMo-2-13B, OLMo-2-7B} (English items.)
\noindent Without thinking:
\begin{Rendering}
Assistant  <|endoftext|><|assistant|>\NL{}\PH{body}<|endoftext|>\NL{}<|user|>\NL{}\PH{question}\NL{}<|assistant|>\NL{}
User       <|endoftext|><|user|>\NL{}\PH{body}\NL{}<|user|>\NL{}\PH{question}\NL{}<|assistant|>\NL{}
System     <|endoftext|><|system|>\NL{}\PH{body}\NL{}<|user|>\NL{}\PH{question}\NL{}<|assistant|>\NL{}
\end{Rendering}
\paragraph{ERNIE-4.5-21B-A3B-Thinking} (Chinese items.)
\noindent With thinking:
\begin{Rendering}
Assistant  <|im_start|>system\NL{}<global_setting>\NL{}think_mode=True\NL{}</global_setting><|im_end|>\NL{}\NL{}<|im_start|>assistant\NL{}<response>\NL{}\PH{body}\NL{}</response>\NL{}<|im_end|>\NL{}\NL{}<|im_start|>user\NL{}\PH{question}<|im_end|>\NL{}\NL{}<|im_start|>assistant\NL{}<think>\NL{}
User       <|im_start|>system\NL{}<global_setting>\NL{}think_mode=True\NL{}</global_setting><|im_end|>\NL{}\NL{}<|im_start|>user\NL{}\PH{body}<|im_end|>\NL{}\NL{}<|im_start|>user\NL{}\PH{question}<|im_end|>\NL{}\NL{}<|im_start|>assistant\NL{}<think>\NL{}
System     <|im_start|>system\NL{}<system_setting>\NL{}\PH{body}\NL{}</system_setting>\NL{}\NL{}<global_setting>\NL{}think_mode=True\NL{}</global_setting><|im_end|>\NL{}\NL{}<|im_start|>user\NL{}\PH{question}<|im_end|>\NL{}\NL{}<|im_start|>assistant\NL{}<think>\NL{}
\end{Rendering}
\paragraph{MiMo-7B-RL-0530} (Chinese items.)
\noindent Without thinking:
\begin{Rendering}
Assistant  <|im_start|>system\NL{}You are MiMo, an AI assistant developed by Xiaomi.<|im_end|>\NL{}<|im_start|>assistant\NL{}\PH{body}<|im_end|>\NL{}<|im_start|>user\NL{}\PH{question}<|im_end|>\NL{}<|im_start|>assistant\NL{}<think>\NL{}\NL{}</think>\NL{}
User       <|im_start|>system\NL{}You are MiMo, an AI assistant developed by Xiaomi.<|im_end|>\NL{}<|im_start|>user\NL{}\PH{body}<|im_end|>\NL{}<|im_start|>user\NL{}\PH{question}<|im_end|>\NL{}<|im_start|>assistant\NL{}<think>\NL{}\NL{}</think>\NL{}
System     <|im_start|>system\NL{}\PH{body}<|im_end|>\NL{}<|im_start|>user\NL{}\PH{question}<|im_end|>\NL{}<|im_start|>assistant\NL{}<think>\NL{}\NL{}</think>\NL{}
\end{Rendering}
\noindent With thinking:
\begin{Rendering}
Assistant  <|im_start|>system\NL{}You are MiMo, an AI assistant developed by Xiaomi.<|im_end|>\NL{}<|im_start|>assistant\NL{}\PH{body}<|im_end|>\NL{}<|im_start|>user\NL{}\PH{question}<|im_end|>\NL{}<|im_start|>assistant\NL{}
User       <|im_start|>system\NL{}You are MiMo, an AI assistant developed by Xiaomi.<|im_end|>\NL{}<|im_start|>user\NL{}\PH{body}<|im_end|>\NL{}<|im_start|>user\NL{}\PH{question}<|im_end|>\NL{}<|im_start|>assistant\NL{}
System     <|im_start|>system\NL{}\PH{body}<|im_end|>\NL{}<|im_start|>user\NL{}\PH{question}<|im_end|>\NL{}<|im_start|>assistant\NL{}
\end{Rendering}
\paragraph{MiMo-VL-7B-RL-2508} (Chinese items.)
\noindent With thinking:
\begin{Rendering}
Assistant  <|im_start|>system\NL{}You are MiMo, an AI assistant developed by Xiaomi.<|im_end|>\NL{}<|im_start|>assistant\NL{}\PH{body}<|im_end|>\NL{}<|im_start|>user\NL{}\PH{question}<|im_end|>\NL{}<|im_start|>assistant\NL{}
User       <|im_start|>system\NL{}You are MiMo, an AI assistant developed by Xiaomi.<|im_end|>\NL{}<|im_start|>user\NL{}\PH{body}<|im_end|>\NL{}<|im_start|>user\NL{}\PH{question}<|im_end|>\NL{}<|im_start|>assistant\NL{}
System     <|im_start|>system\NL{}\PH{body}<|im_end|>\NL{}<|im_start|>user\NL{}\PH{question}<|im_end|>\NL{}<|im_start|>assistant\NL{}
\end{Rendering}
\paragraph{MiMo-V2.6-Distill-Qwen-9B} (Chinese items.)
\noindent Without thinking:
\begin{Rendering}
Assistant  <|im_start|>assistant\NL{}<think></think>\PH{body}<|im_end|><|im_start|>user\NL{}\PH{question}<|im_end|><|im_start|>assistant\NL{}<think></think>
User       <|im_start|>user\NL{}\PH{body}<|im_end|><|im_start|>user\NL{}\PH{question}<|im_end|><|im_start|>assistant\NL{}<think></think>
System     <|im_start|>system\NL{}\PH{body}<|im_end|><|im_start|>user\NL{}\PH{question}<|im_end|><|im_start|>assistant\NL{}<think></think>
\end{Rendering}
\noindent With thinking:
\begin{Rendering}
Assistant  <|im_start|>assistant\NL{}<think></think>\PH{body}<|im_end|><|im_start|>user\NL{}\PH{question}<|im_end|><|im_start|>assistant\NL{}
User       <|im_start|>user\NL{}\PH{body}<|im_end|><|im_start|>user\NL{}\PH{question}<|im_end|><|im_start|>assistant\NL{}
System     <|im_start|>system\NL{}\PH{body}<|im_end|><|im_start|>user\NL{}\PH{question}<|im_end|><|im_start|>assistant\NL{}
\end{Rendering}
\paragraph{Moonlight-16B-A3B-Instruct} (Chinese items.)
\noindent Without thinking:
\begin{Rendering}
Assistant  <|im_system|>system<|im_middle|>You are a helpful assistant<|im_end|><|im_assistant|>assistant<|im_middle|>\PH{body}<|im_end|><|im_user|>user<|im_middle|>\PH{question}<|im_end|><|im_assistant|>assistant<|im_middle|>
User       <|im_system|>system<|im_middle|>You are a helpful assistant<|im_end|><|im_user|>user<|im_middle|>\PH{body}<|im_end|><|im_user|>user<|im_middle|>\PH{question}<|im_end|><|im_assistant|>assistant<|im_middle|>
System     <|im_system|>system<|im_middle|>\PH{body}<|im_end|><|im_user|>user<|im_middle|>\PH{question}<|im_end|><|im_assistant|>assistant<|im_middle|>
\end{Rendering}
\paragraph{Apertus-8B-Instruct-2509} (English items.)
\noindent Without thinking:
\begin{Rendering}
Assistant  <s><|system_start|>You are Apertus, a helpful assistant created by the SwissAI initiative.\NL{}Knowledge cutoff: 2024-04\NL{}Current date: \PH{date}<|system_end|><|developer_start|>Deliberation: disabled\NL{}Tool Capabilities: disabled<|developer_end|><|assistant_start|>\PH{body}<|assistant_end|><|user_start|>\PH{question}<|user_end|><|assistant_start|>
User       <s><|system_start|>You are Apertus, a helpful assistant created by the SwissAI initiative.\NL{}Knowledge cutoff: 2024-04\NL{}Current date: \PH{date}<|system_end|><|developer_start|>Deliberation: disabled\NL{}Tool Capabilities: disabled<|developer_end|><|user_start|>\PH{body}<|user_end|><|user_start|>\PH{question}<|user_end|><|assistant_start|>
System     <s><|system_start|>\PH{body}<|system_end|><|developer_start|>Deliberation: disabled\NL{}Tool Capabilities: disabled<|developer_end|><|user_start|>\PH{question}<|user_end|><|assistant_start|>
\end{Rendering}
\paragraph{phi-4} (English items.)
\noindent Without thinking:
\begin{Rendering}
Assistant  <|im_start|>assistant<|im_sep|>\PH{body}<|im_end|><|im_start|>user<|im_sep|>\PH{question}<|im_end|><|im_start|>assistant<|im_sep|>
User       <|im_start|>user<|im_sep|>\PH{body}<|im_end|><|im_start|>user<|im_sep|>\PH{question}<|im_end|><|im_start|>assistant<|im_sep|>
System     <|im_start|>system<|im_sep|>\PH{body}<|im_end|><|im_start|>user<|im_sep|>\PH{question}<|im_end|><|im_start|>assistant<|im_sep|>
\end{Rendering}

\subsection{Natural Conversation Order}
\label{app:templates_natural}

The conversation is [user: request] [\textit{role}: body] [user: question], followed by the generation prompt, for the models of Appendix~\ref{app:natural}. In the first ten models, the template keeps a System message in place after a user turn.

\paragraph{Seed-OSS-36B} (Chinese items.)
\noindent Without thinking:
\begin{Rendering}
Start      <seed:bos>system\NL{}You are an intelligent assistant that can answer questions in one step without the need for reasoning and thinking, that is, your thinking budget is 0. Next, please skip the thinking process and directly start answering the user's questions.\NL{}<seed:eos><seed:bos>user\NL{}\PH{request}<seed:eos>
Assistant  \PH{start}<seed:bos>assistant\NL{}\PH{body}<seed:eos><seed:bos>user\NL{}\PH{question}<seed:eos><seed:bos>assistant\NL{}<seed:think>\NL{}<seed:cot_budget_reflect>The current thinking budget is 0, so I will directly start answering the question.</seed:cot_budget_reflect>\NL{}</seed:think>
User       \PH{start}<seed:bos>user\NL{}\PH{body}<seed:eos><seed:bos>user\NL{}\PH{question}<seed:eos><seed:bos>assistant\NL{}<seed:think>\NL{}<seed:cot_budget_reflect>The current thinking budget is 0, so I will directly start answering the question.</seed:cot_budget_reflect>\NL{}</seed:think>
System     \PH{start}<seed:bos>system\NL{}\PH{body}<seed:eos><seed:bos>user\NL{}\PH{question}<seed:eos><seed:bos>assistant\NL{}<seed:think>\NL{}<seed:cot_budget_reflect>The current thinking budget is 0, so I will directly start answering the question.</seed:cot_budget_reflect>\NL{}</seed:think>
\end{Rendering}
\paragraph{Gemma-4-31B} (English items.)
\noindent Without thinking:
\begin{Rendering}
Assistant  <bos><|turn>user\NL{}\PH{request}<turn|>\NL{}<|turn>model\NL{}\PH{body}<turn|>\NL{}<|turn>user\NL{}\PH{question}<turn|>\NL{}<|turn>model\NL{}<|channel>thought\NL{}<channel|>
User       <bos><|turn>user\NL{}\PH{request}<turn|>\NL{}<|turn>user\NL{}\PH{body}<turn|>\NL{}<|turn>user\NL{}\PH{question}<turn|>\NL{}<|turn>model\NL{}<|channel>thought\NL{}<channel|>
System     <bos><|turn>user\NL{}\PH{request}<turn|>\NL{}<|turn>system\NL{}\PH{body}<turn|>\NL{}<|turn>user\NL{}\PH{question}<turn|>\NL{}<|turn>model\NL{}<|channel>thought\NL{}<channel|>
\end{Rendering}
\paragraph{EXAONE-4.0.1-32B} (English items.)
\noindent Without thinking:
\begin{Rendering}
Assistant  [|user|]\NL{}\PH{request}[|endofturn|]\NL{}[|assistant|]\NL{}<think>\NL{}\NL{}</think>\NL{}\NL{}\PH{body}[|endofturn|]\NL{}[|user|]\NL{}\PH{question}[|endofturn|]\NL{}[|assistant|]\NL{}<think>\NL{}\NL{}</think>\NL{}\NL{}
User       [|user|]\NL{}\PH{request}[|endofturn|]\NL{}[|user|]\NL{}\PH{body}[|endofturn|]\NL{}[|user|]\NL{}\PH{question}[|endofturn|]\NL{}[|assistant|]\NL{}<think>\NL{}\NL{}</think>\NL{}\NL{}
System     [|user|]\NL{}\PH{request}[|endofturn|]\NL{}[|system|]\NL{}\PH{body}[|endofturn|]\NL{}[|user|]\NL{}\PH{question}[|endofturn|]\NL{}[|assistant|]\NL{}<think>\NL{}\NL{}</think>\NL{}\NL{}
\end{Rendering}
\paragraph{GLM-4.6V-Flash} (Chinese items.)
\noindent Without thinking:
\begin{Rendering}
Assistant  [gMASK]<sop><|user|>\NL{}\PH{request}/nothink<|assistant|>\NL{}<think></think>\NL{}\PH{body}<|user|>\NL{}\PH{question}/nothink<|assistant|>\NL{}<think></think>\NL{}
User       [gMASK]<sop><|user|>\NL{}\PH{request}/nothink<|user|>\NL{}\PH{body}/nothink<|user|>\NL{}\PH{question}/nothink<|assistant|>\NL{}<think></think>\NL{}
System     [gMASK]<sop><|user|>\NL{}\PH{request}/nothink<|system|>\NL{}\PH{body}<|user|>\NL{}\PH{question}/nothink<|assistant|>\NL{}<think></think>\NL{}
\end{Rendering}
\paragraph{Granite-4.1-30B} (English items.)
\noindent Without thinking:
\begin{Rendering}
Assistant  <|start_of_role|>user<|end_of_role|>\PH{request}<|end_of_text|>\NL{}<|start_of_role|>assistant<|end_of_role|>\PH{body}<|end_of_text|>\NL{}<|start_of_role|>user<|end_of_role|>\PH{question}<|end_of_text|>\NL{}<|start_of_role|>assistant<|end_of_role|>
User       <|start_of_role|>user<|end_of_role|>\PH{request}<|end_of_text|>\NL{}<|start_of_role|>user<|end_of_role|>\PH{body}<|end_of_text|>\NL{}<|start_of_role|>user<|end_of_role|>\PH{question}<|end_of_text|>\NL{}<|start_of_role|>assistant<|end_of_role|>
System     <|start_of_role|>user<|end_of_role|>\PH{request}<|end_of_text|>\NL{}<|start_of_role|>system<|end_of_role|>\PH{body}<|end_of_text|>\NL{}<|start_of_role|>user<|end_of_role|>\PH{question}<|end_of_text|>\NL{}<|start_of_role|>assistant<|end_of_role|>
\end{Rendering}
\paragraph{OLMo-2-7B} (English items.)
\noindent Without thinking:
\begin{Rendering}
Assistant  <|endoftext|><|user|>\NL{}\PH{request}\NL{}<|assistant|>\NL{}\PH{body}<|endoftext|>\NL{}<|user|>\NL{}\PH{question}\NL{}<|assistant|>\NL{}
User       <|endoftext|><|user|>\NL{}\PH{request}\NL{}<|user|>\NL{}\PH{body}\NL{}<|user|>\NL{}\PH{question}\NL{}<|assistant|>\NL{}
System     <|endoftext|><|user|>\NL{}\PH{request}\NL{}<|system|>\NL{}\PH{body}\NL{}<|user|>\NL{}\PH{question}\NL{}<|assistant|>\NL{}
\end{Rendering}
\paragraph{ERNIE-4.5-21B-A3B-Thinking} (Chinese items.)
\noindent With thinking:
\begin{Rendering}
Assistant  <|im_start|>system\NL{}<global_setting>\NL{}think_mode=True\NL{}</global_setting><|im_end|>\NL{}\NL{}<|im_start|>user\NL{}\PH{request}<|im_end|>\NL{}\NL{}<|im_start|>assistant\NL{}<response>\NL{}\PH{body}\NL{}</response>\NL{}<|im_end|>\NL{}\NL{}<|im_start|>user\NL{}\PH{question}<|im_end|>\NL{}\NL{}<|im_start|>assistant\NL{}<think>\NL{}
User       <|im_start|>system\NL{}<global_setting>\NL{}think_mode=True\NL{}</global_setting><|im_end|>\NL{}\NL{}<|im_start|>user\NL{}\PH{request}<|im_end|>\NL{}\NL{}<|im_start|>user\NL{}\PH{body}<|im_end|>\NL{}\NL{}<|im_start|>user\NL{}\PH{question}<|im_end|>\NL{}\NL{}<|im_start|>assistant\NL{}<think>\NL{}
System     <|im_start|>system\NL{}<global_setting>\NL{}think_mode=True\NL{}</global_setting><|im_end|>\NL{}\NL{}<|im_start|>user\NL{}\PH{request}<|im_end|>\NL{}\NL{}<|im_start|>system\NL{}\PH{body}<|im_end|>\NL{}\NL{}<|im_start|>user\NL{}\PH{question}<|im_end|>\NL{}\NL{}<|im_start|>assistant\NL{}<think>\NL{}
\end{Rendering}
\paragraph{MiMo-7B-RL-0530} (Chinese items.)
\noindent Without thinking:
\begin{Rendering}
Assistant  <|im_start|>system\NL{}You are MiMo, an AI assistant developed by Xiaomi.<|im_end|>\NL{}<|im_start|>user\NL{}\PH{request}<|im_end|>\NL{}<|im_start|>assistant\NL{}\PH{body}<|im_end|>\NL{}<|im_start|>user\NL{}\PH{question}<|im_end|>\NL{}<|im_start|>assistant\NL{}<think>\NL{}\NL{}</think>\NL{}
User       <|im_start|>system\NL{}You are MiMo, an AI assistant developed by Xiaomi.<|im_end|>\NL{}<|im_start|>user\NL{}\PH{request}<|im_end|>\NL{}<|im_start|>user\NL{}\PH{body}<|im_end|>\NL{}<|im_start|>user\NL{}\PH{question}<|im_end|>\NL{}<|im_start|>assistant\NL{}<think>\NL{}\NL{}</think>\NL{}
System     <|im_start|>system\NL{}You are MiMo, an AI assistant developed by Xiaomi.<|im_end|>\NL{}<|im_start|>user\NL{}\PH{request}<|im_end|>\NL{}<|im_start|>system\NL{}\PH{body}<|im_end|>\NL{}<|im_start|>user\NL{}\PH{question}<|im_end|>\NL{}<|im_start|>assistant\NL{}<think>\NL{}\NL{}</think>\NL{}
\end{Rendering}
\paragraph{Moonlight-16B-A3B-Instruct} (Chinese items.)
\noindent Without thinking:
\begin{Rendering}
Assistant  <|im_system|>system<|im_middle|>You are a helpful assistant<|im_end|><|im_user|>user<|im_middle|>\PH{request}<|im_end|><|im_assistant|>assistant<|im_middle|>\PH{body}<|im_end|><|im_user|>user<|im_middle|>\PH{question}<|im_end|><|im_assistant|>assistant<|im_middle|>
User       <|im_system|>system<|im_middle|>You are a helpful assistant<|im_end|><|im_user|>user<|im_middle|>\PH{request}<|im_end|><|im_user|>user<|im_middle|>\PH{body}<|im_end|><|im_user|>user<|im_middle|>\PH{question}<|im_end|><|im_assistant|>assistant<|im_middle|>
System     <|im_system|>system<|im_middle|>You are a helpful assistant<|im_end|><|im_user|>user<|im_middle|>\PH{request}<|im_end|><|im_system|>system<|im_middle|>\PH{body}<|im_end|><|im_user|>user<|im_middle|>\PH{question}<|im_end|><|im_assistant|>assistant<|im_middle|>
\end{Rendering}
\paragraph{phi-4} (English items.)
\noindent Without thinking:
\begin{Rendering}
Assistant  <|im_start|>user<|im_sep|>\PH{request}<|im_end|><|im_start|>assistant<|im_sep|>\PH{body}<|im_end|><|im_start|>user<|im_sep|>\PH{question}<|im_end|><|im_start|>assistant<|im_sep|>
User       <|im_start|>user<|im_sep|>\PH{request}<|im_end|><|im_start|>user<|im_sep|>\PH{body}<|im_end|><|im_start|>user<|im_sep|>\PH{question}<|im_end|><|im_start|>assistant<|im_sep|>
System     <|im_start|>user<|im_sep|>\PH{request}<|im_end|><|im_start|>system<|im_sep|>\PH{body}<|im_end|><|im_start|>user<|im_sep|>\PH{question}<|im_end|><|im_start|>assistant<|im_sep|>
\end{Rendering}

In the next four, the template does not keep it in place, so these models are compared on two roles: the Qwen3.8 and Apertus templates raise an error, the gpt-oss template leaves the message out, and the DeepSeek-R1-Distill-Llama-70B template moves its text to the start of the input.

\paragraph{Qwen3.8-27B} (Chinese items.)
\noindent Without thinking:
\begin{Rendering}
Assistant  <|im_start|>user\NL{}\PH{request}<|im_end|>\NL{}<|im_start|>assistant\NL{}<think>\NL{}\NL{}</think>\NL{}\NL{}\PH{body}<|im_end|>\NL{}<|im_start|>user\NL{}\PH{question}<|im_end|>\NL{}<|im_start|>assistant\NL{}<think>\NL{}\NL{}</think>\NL{}\NL{}
User       <|im_start|>user\NL{}\PH{request}<|im_end|>\NL{}<|im_start|>user\NL{}\PH{body}<|im_end|>\NL{}<|im_start|>user\NL{}\PH{question}<|im_end|>\NL{}<|im_start|>assistant\NL{}<think>\NL{}\NL{}</think>\NL{}\NL{}
System     \Note{template error: System message must be at the beginning.}
\end{Rendering}
\paragraph{gpt-oss-20b} (English items.)
\noindent With thinking:
\begin{Rendering}
Start      <|start|>system<|message|>You are ChatGPT, a large language model trained by OpenAI.\NL{}Knowledge cutoff: 2024-06\NL{}Current date: \PH{date}\NL{}\NL{}Reasoning: medium\NL{}\NL{}# Valid channels: analysis, commentary, final. Channel must be included for every message.<|end|><|start|>user<|message|>\PH{request}<|end|>
Assistant  \PH{start}<|start|>assistant<|channel|>final<|message|>\PH{body}<|end|><|start|>user<|message|>\PH{question}<|end|><|start|>assistant
User       \PH{start}<|start|>user<|message|>\PH{body}<|end|><|start|>user<|message|>\PH{question}<|end|><|start|>assistant
System     \Note{the template leaves the message out:} \PH{start}<|start|>user<|message|>\PH{question}<|end|><|start|>assistant
\end{Rendering}
\paragraph{DeepSeek-R1-Distill-Llama-70B} (Chinese items.)
\noindent With thinking:
\begin{Rendering}
Assistant  <|begin_of_sentence|><|User|>\PH{request}<|Assistant|>\PH{body}<|end_of_sentence|><|User|>\PH{question}<|Assistant|><think>\NL{}
User       <|begin_of_sentence|><|User|>\PH{request}<|User|>\PH{body}<|User|>\PH{question}<|Assistant|><think>\NL{}
System     \Note{the template moves the message to the start:} <|begin_of_sentence|>\PH{body}<|User|>\PH{request}<|User|>\PH{question}<|Assistant|><think>\NL{}
\end{Rendering}
\paragraph{Apertus-8B-Instruct-2509} (English items.)
\noindent Without thinking:
\begin{Rendering}
Assistant  <s><|system_start|>You are Apertus, a helpful assistant created by the SwissAI initiative.\NL{}Knowledge cutoff: 2024-04\NL{}Current date: \PH{date}<|system_end|><|developer_start|>Deliberation: disabled\NL{}Tool Capabilities: disabled<|developer_end|><|user_start|>\PH{request}<|user_end|><|assistant_start|>\PH{body}<|assistant_end|><|user_start|>\PH{question}<|user_end|><|assistant_start|>
User       <s><|system_start|>You are Apertus, a helpful assistant created by the SwissAI initiative.\NL{}Knowledge cutoff: 2024-04\NL{}Current date: \PH{date}<|system_end|><|developer_start|>Deliberation: disabled\NL{}Tool Capabilities: disabled<|developer_end|><|user_start|>\PH{request}<|user_end|><|user_start|>\PH{body}<|user_end|><|user_start|>\PH{question}<|user_end|><|assistant_start|>
System     \Note{template error: Invalid message role}
\end{Rendering}

The last three templates do not accept the main probe and merge the User message under test with the request and the question into one user turn. DeepSeek-V4.1-Flash has no chat template; its inputs are produced by the prompt encoder of the release (\texttt{encoding/encoding.py}), in chat mode and in thinking mode with reasoning effort ``high''.

\paragraph{DeepSeek-V4.1-Flash} (Chinese items.)
\noindent Without thinking:
\begin{Rendering}
Assistant  <|begin_of_sentence|><|User|>\PH{request}<|Assistant|></think>\PH{body}<|end_of_sentence|><|User|>\PH{question}<|Assistant|></think>
User       <|begin_of_sentence|><|User|>\PH{request}\NL{}\NL{}\PH{body}\NL{}\NL{}\PH{question}<|Assistant|></think>
System     <|begin_of_sentence|><|User|>\PH{request}<|System|>\PH{body}<|User|>\PH{question}<|Assistant|></think>
\end{Rendering}
\noindent With thinking:
\begin{Rendering}
Assistant  <|begin_of_sentence|><|System|>Reasoning Effort: 75 (range 1-100, the higher the value, the more thorough the reasoning)\NL{}\NL{}<|User|>\PH{request}<|Assistant|></think>\PH{body}<|end_of_sentence|><|User|>\PH{question}<|Assistant|><think>
User       <|begin_of_sentence|><|System|>Reasoning Effort: 75 (range 1-100, the higher the value, the more thorough the reasoning)\NL{}\NL{}<|User|>\PH{request}\NL{}\NL{}\PH{body}\NL{}\NL{}\PH{question}<|Assistant|><think>
System     <|begin_of_sentence|><|System|>Reasoning Effort: 75 (range 1-100, the higher the value, the more thorough the reasoning)\NL{}\NL{}<|User|>\PH{request}<|System|>\PH{body}<|User|>\PH{question}<|Assistant|><think>
\end{Rendering}
\paragraph{Ministral-3-14B-Instruct} (English items.)
\noindent Without thinking:
\begin{Rendering}
Start      <s>[SYSTEM_PROMPT]You are Ministral-3-14B-Instruct-2512, a Large Language Model (LLM) created by Mistral AI, a French startup headquartered in Paris.\NL{}You power an AI assistant called Le Chat.\NL{}Your knowledge base was last updated on 2023-10-01.\NL{}The current date is \LB{}today\RB{}.\NL{}\NL{}When you're not sure about some information or when the user's request requires up-to-date or specific data, you must use the available tools to fetch the information. Do not hesitate to use tools whenever they can provide a more accurate or complete response. If no relevant tools are available, then clearly state that you don't have the information and avoid making up anything.\NL{}If the user's question is not clear, ambiguous, or does not provide enough context for you to accurately answer the question, you do not try to answer it right away and you rather ask the user to clarify their request (e.g. "What are some good restaurants around me?" => "Where are you?" or "When is the next flight to Tokyo" => "Where do you travel from?").\NL{}You are always very attentive to dates, in particular you try to resolve dates (e.g. "yesterday" is \LB{}yesterday\RB{}) and when asked about information at specific dates, you discard information that is at another date.\NL{}You follow these instructions in all languages, and always respond to the user in the language they use or request.\NL{}Next sections describe the capabilities that you have.\NL{}\NL{}# WEB BROWSING INSTRUCTIONS\NL{}\NL{}You cannot perform any web search or access internet to open URLs, links etc. If it seems like the user is expecting you to do so, you clarify the situation and ask the user to copy paste the text directly in the chat.\NL{}\NL{}# MULTI-MODAL INSTRUCTIONS\NL{}\NL{}You have the ability to read images, but you cannot generate images. You also cannot read nor transcribe audio files or videos.\NL{}\NL{}# TOOL CALLING INSTRUCTIONS\NL{}\NL{}You may have access to tools that you can use to fetch information or perform actions. You must use these tools in the following situations:\NL{}\NL{}1. When the request requires up-to-date information.\NL{}2. When the request requires specific data that you do not have in your knowledge base.\NL{}3. When the request involves actions that you cannot perform without tools.\NL{}\NL{}Always prioritize using tools to provide the most accurate and helpful response. If tools are not available, inform the user that you cannot perform the requested action at the moment.[/SYSTEM_PROMPT]
Assistant  \PH{start}[INST]\PH{request}[/INST]\PH{body}</s>[INST]\PH{question}[/INST]
User       \PH{start}[INST]\PH{request}\NL{}\NL{}\PH{body}\NL{}\NL{}\PH{question}[/INST]
System     \PH{start}[INST]\PH{request}[/INST][SYSTEM_PROMPT]\PH{body}[/SYSTEM_PROMPT][INST]\PH{question}[/INST]
\end{Rendering}
\paragraph{Ministral-3-14B-Reasoning} (English items.)
\noindent With thinking:
\begin{Rendering}
Start      <s>[SYSTEM_PROMPT]# HOW YOU SHOULD THINK AND ANSWER\NL{}\NL{}First draft your thinking process (inner monologue) until you arrive at a response. Format your response using Markdown, and use LaTeX for any mathematical equations. Write both your thoughts and the response in the same language as the input.\NL{}\NL{}Your thinking process must follow the template below:[THINK]Your thoughts or/and draft, like working through an exercise on scratch paper. Be as casual and as long as you want until you are confident to generate the response to the user.[/THINK]Here, provide a self-contained response.[/SYSTEM_PROMPT]
Assistant  \PH{start}[INST]\PH{request}[/INST]\PH{body}</s>[INST]\PH{question}[/INST]
User       \PH{start}[INST]\PH{request}\NL{}\NL{}\PH{body}\NL{}\NL{}\PH{question}[/INST]
System     \PH{start}[INST]\PH{request}[/INST][SYSTEM_PROMPT]\PH{body}[/SYSTEM_PROMPT][INST]\PH{question}[/INST]
\end{Rendering}

Under the main probe, the DeepSeek-V4.1-Flash encoder gives an opening Assistant message no role token, the Ministral templates reject it, and all three merge an opening User message with the question. In the Ministral blocks, \PH{start} is the text on the line Start of the natural-order block above.

\paragraph{DeepSeek-V4.1-Flash} (Chinese items.)
\noindent Without thinking (not run; shown for comparison):
\begin{Rendering}
Assistant  <|begin_of_sentence|>\PH{body}<|end_of_sentence|><|User|>\PH{question}<|Assistant|></think>
User       <|begin_of_sentence|><|User|>\PH{body}\NL{}\NL{}\PH{question}<|Assistant|></think>
System     <|begin_of_sentence|><|System|>\PH{body}<|User|>\PH{question}<|Assistant|></think>
\end{Rendering}
\noindent With thinking (not run; shown for comparison):
\begin{Rendering}
Assistant  <|begin_of_sentence|><|System|>Reasoning Effort: 75 (range 1-100, the higher the value, the more thorough the reasoning)\NL{}\NL{}\PH{body}<|end_of_sentence|><|User|>\PH{question}<|Assistant|><think>
User       <|begin_of_sentence|><|System|>Reasoning Effort: 75 (range 1-100, the higher the value, the more thorough the reasoning)\NL{}\NL{}<|User|>\PH{body}\NL{}\NL{}\PH{question}<|Assistant|><think>
System     <|begin_of_sentence|><|System|>Reasoning Effort: 75 (range 1-100, the higher the value, the more thorough the reasoning)\NL{}\NL{}\PH{body}<|User|>\PH{question}<|Assistant|><think>
\end{Rendering}
\paragraph{Ministral-3-14B-Instruct} (English items.)
\noindent Without thinking (not run; shown for comparison):
\begin{Rendering}
Assistant  \Note{template error: Conversation must start with a user or system message, got assistant.}
User       \PH{start}[INST]\PH{body}\NL{}\NL{}\PH{question}[/INST]
System     <s>[SYSTEM_PROMPT]\PH{body}[/SYSTEM_PROMPT][INST]\PH{question}[/INST]
\end{Rendering}
\paragraph{Ministral-3-14B-Reasoning} (English items.)
\noindent With thinking (not run; shown for comparison):
\begin{Rendering}
Assistant  \Note{template error: Conversation must start with a user or system message, got assistant.}
User       \PH{start}[INST]\PH{body}\NL{}\NL{}\PH{question}[/INST]
System     <s>[SYSTEM_PROMPT]\PH{body}[/SYSTEM_PROMPT][INST]\PH{question}[/INST]
\end{Rendering}

\clearpage
\section{Configurations}
\label{app:config}

Table~\ref{tab:self_config} identifies every checkpoint by its repository and the revision of its files; all files of a checkpoint come from that one revision, and the weights are used in the format of the release (Table~\ref{tab:self_models}). The table also gives the template arguments of each mode, with which the items of Appendix~\ref{app:templates} are rendered (``default'': no argument).

\begin{table}[htbp]
\centering
\caption{Checkpoints and template arguments. Revision: the first 12 characters of the commit. $^{a}$~ModelScope revision; the files are identical to those on Hugging Face. DeepSeek-V4.1-Flash has no chat template, and its arguments are those of the prompt encoder of the release.}
\label{tab:self_config}
\scriptsize
\setlength{\tabcolsep}{2.5pt}
\begin{tabular}{llp{3.2cm}p{3.5cm}}
\toprule
Repository & Revision & Without thinking & With thinking \\
\midrule
\texttt{Qwen/Qwen3.8-27B} & \texttt{1d4bf0f2ff60} & \texttt{enable\_thinking=False} & \texttt{enable\_thinking=True} \\
\texttt{Qwen/Qwen3.8-27B} (format ablation) & \texttt{1d4bf0f2ff60} & \texttt{enable\_thinking=\allowbreak False, preserve\_thinking=\allowbreak False} & \texttt{preserve\_thinking=False} \\
\texttt{Qwen/Qwen3.8-Flash-Next-FP8} & \texttt{236dfdf28582} & \texttt{enable\_thinking=False} & default \\
\texttt{Qwen/Qwen3.6-27B} & \texttt{6a9e13bd6fc8} & \texttt{enable\_thinking=False} & default \\
\texttt{Qwen/Qwen3.6-35B-A3B} & \texttt{995ad96eacd9} & \texttt{enable\_thinking=False} & default \\
\texttt{Qwen/Qwen3.5-27B} & \texttt{fc05daec18b0} & \texttt{enable\_thinking=False} & default \\
\texttt{zai-org/GLM-4.6V-Flash} & \texttt{411bb4d77144} & \texttt{enable\_thinking=False} & default \\
\texttt{zai-org/GLM-4.7-Flash} & \texttt{7dd20894a642} & \texttt{enable\_thinking=False} & \texttt{enable\_thinking=True} \\
\texttt{zai-org/GLM-5.3-Flash} & \texttt{46a050bec130} & -- & default \\
\texttt{deepseek-ai/DeepSeek-R1-Distill-Llama-70B} & \texttt{b1c0b44b4369} & -- & default \\
\texttt{openai/gpt-oss-20b} & \texttt{6cee5e81ee83} & -- & default \\
\texttt{openai/gpt-oss-120b} & \texttt{b5c939de8f75} & -- & default \\
\texttt{google/gemma-4-31B-it} & \texttt{842da3794eaa} & default & \texttt{enable\_thinking=True} \\
\texttt{ByteDance-Seed/Seed-OSS-36B-Instruct} & \texttt{497f1dca95eb} & \texttt{thinking\_budget=0} & default \\
\texttt{LGAI-EXAONE/EXAONE-4.0.1-32B} & \texttt{a6b0bd4b0fb4} & \texttt{enable\_thinking=False} & \texttt{enable\_thinking=True} \\
\texttt{ibm-granite/granite-4.2-30b} & \texttt{9e668ce1c538} & \texttt{enable\_thinking=False} & \texttt{enable\_thinking=True} \\
\texttt{ibm-granite/granite-4.1-30b} & \texttt{4fae6278f713} & default & -- \\
\texttt{allenai/OLMo-2-1124-13B-Instruct} & \texttt{3a5c85baefbb} & default & -- \\
\texttt{allenai/OLMo-2-1124-7B-Instruct} & \texttt{470b1fba1ae0} & default & -- \\
\texttt{deepseek-ai/DeepSeek-V4.1-Flash-1017} & \texttt{9f9818127fdf}$^{a}$ & \texttt{thinking\_mode=chat} & \texttt{thinking\_mode=thinking, reasoning\_effort=high} \\
\texttt{baidu/ERNIE-4.5-21B-A3B-Thinking} & \texttt{4341bb42644d} & -- & default \\
\texttt{XiaomiMiMo/MiMo-7B-RL-0530} & \texttt{ee8dd14ecd59} & \texttt{enable\_thinking=False} & default \\
\texttt{XiaomiMiMo/MiMo-VL-7B-RL-2508} & \texttt{ec4ebdc29b83} & -- & default \\
\texttt{XiaomiMiMo/MiMo-V2.6-Distill-Qwen-9B} & \texttt{1f09ddcc1a0e} & \texttt{enable\_thinking=False} & default \\
\texttt{moonshotai/Moonlight-16B-A3B-Instruct} & \texttt{4e735b07a89f} & default & -- \\
\texttt{swiss-ai/Apertus-8B-Instruct-2509} & \texttt{b946d40447b2} & default & -- \\
\texttt{microsoft/phi-4} & \texttt{2db69c1c3e91} & default & -- \\
\texttt{mistralai/Ministral-3-14B-Instruct-2512-BF16} & \texttt{3cea74c1ebaf} & default & -- \\
\texttt{mistralai/Ministral-3-14B-Reasoning-2512} & \texttt{51f9210f3cd2} & -- & default \\
\bottomrule
\end{tabular}
\end{table}

\paragraph{Software and hardware.} Local runs use two GPU types: NVIDIA RTX PRO 6000 Blackwell Server Edition, with PyTorch 2.11.0 built for CUDA 13.0, and NVIDIA GeForce RTX 4090, with PyTorch 2.5.1 built for CUDA 12.4 (GLM-5.3-Flash: PyTorch 2.6.0 built for CUDA 12.4; DeepSeek-V4.1-Flash: PyTorch 2.10.0 built for CUDA 12.8). Every model except DeepSeek-V4.1-Flash runs through Hugging Face Transformers 5.8.0 or 5.19.0; the identity file (identity.json) of each run records the versions it used. A model occupies one or more GPUs; GLM-5.3-Flash, DeepSeek-V4.1-Flash and, in its natural-order run, Qwen3.8-Flash-Next also keep part of their weights in CPU memory. The two runs of Qwen3.8-27B in Section~\ref{sec:setup_compute} used one GPU type each (0.771 on the RTX PRO 6000, 0.770 on the RTX 4090); how much the generated reasoning changes when the unedited coordinates are copied in place is reported in Appendix~\ref{app:mech_ablation}. DeepSeek-V4.1-Flash has no Transformers implementation and runs on the model code of its release, with the custom kernels of the release replaced by PyTorch implementations of the same arithmetic and the weights kept in the release format (FP8, with FP4 expert weights). Both Ministral-3-14B models run in BF16. The repository code of Moonlight-16B-A3B-Instruct does not run under Transformers 5.8, and the tokenizer that Transformers builds from its tiktoken file gives the special tokens other ids than the repository's own tokenizer; the model therefore runs on the Transformers implementation of its architecture (DeepSeek-V3), whose parameters match the 377 tensors of the checkpoint one to one, with a tokenizer whose ids equal those of the repository's tokenizer on every text and rendered input we compared. The default system message of the Apertus template contains the date of the call.

\paragraph{Language of the items.} The message bodies were written in Chinese. Models from Chinese developers (Qwen, GLM, DeepSeek, Seed) receive the original Chinese items, and all other models (Gemma, EXAONE, gpt-oss, Granite, OLMo-2, Ministral) a translation made by hand: the 30 template bodies were translated from the Chinese ones and each variant obtained by substituting its numbers (Section~\ref{sec:setup}). As an ablation, Qwen3.8-27B, the only model from a Chinese developer among the models of the causal tests, was also run with the English items, in the cross-model measures and in the causal tests; neither its classification nor the conclusions of the causal tests change. In both languages it has both levels: the self-index covers all 64 layers of message content and of the response onset, and the model claims its own messages with probability 0.68 against at most 0.08 for external messages in English, and 0.77 against at most 0.07 in Chinese. Its self-index is the same in both languages and the coordinates that the causal tests edit are largely shared between them (below), and both levels, as well as the self-index alone, occur in models tested in either language (Section~\ref{sec:res_crossmodel}); each of the other models is therefore tested in one language.

\paragraph{Causal tests with the English items.} The subspaces were fitted on the English versions of the discovery bodies by the procedure of Section~\ref{sec:setup_swap}, and the attribution, content, and arithmetic questions and the secret game were run on the English versions of the 75 held-out bodies, at every layer with eight dimensions and, with thinking, also installation with 64 dimensions, with the scripts and settings of the Chinese runs (Table~\ref{tab:self_english}). Removal reproduces the Chinese results: replacing the eight coordinates of the model's own message by those of the same text as a User or a System message removes its claim (\(-0.63\) and \(-0.55\) without thinking, \(-0.89\) and \(-0.89\) with thinking), the same replacement in a random subspace changes it by at most 0.03, and copying the unedited coordinates by at most 0.04. The answers about the content of the message and to the multiplication stay correct (0.93--1.00 and 1.00 of the finished items in every condition). Installation acts in the same direction but is weaker in English: into a User message \(+0.32\) without thinking (Chinese \(+0.49\)) and \(+0.04\) with thinking (Chinese \(+0.60\)), into a System message \(+0.07\) without thinking (Chinese \(+0.39\)). With thinking, the difference lies in the reasoning: counted as in Section~\ref{sec:res_mechanism}, 22 of the 150 traces for the edited User message describe an assistant message in the conversation (4 for the unedited message and 4 after the random-subspace edit), against 102 of 150 in Chinese. With 64 dimensions, installation acts with thinking in English as well (\(+0.41\), 0.46 of the gap between the unedited messages; random subspace 0.00), and 81 of the 150 traces describe an assistant message. In both languages the claim follows the reasoning: every item answered ``you'' is among the traces that describe an assistant message (90 of 90 in Chinese; 6 of 6 and, with 64 dimensions, 61 of 61 in English). With the English items, eight dimensions produce such traces less often.

\paragraph{The secret game without thinking.} With the English items, the unedited game without thinking separates the two sources less sharply than the other readouts: the model plays the setter with its own X in every game, and with a User X in 35 of the 100 (Chinese: 1). It still acts on the source, and the edits act as in Chinese: removal leaves the role unchanged (0.00 in both languages), and installation makes the model play the setter with a User X in every game, the largest change this baseline allows (\(+0.65\); random subspace \(+0.08\)). Since the size of this change is bounded by a different baseline, it is not compared with the Chinese \(+0.25\). With thinking, the English game equals the Chinese one (unedited 1.00 and 0.00, removal \(-0.04\), installation 0.00), and the control question is answered correctly in every game in both modes. With the Chinese items, the unedited conditions thus separate the model's own message from the external ones more widely in both questions about the source, in attribution (without thinking 0.79 against 0.04 and 0.07; in English 0.70 against 0.08 and 0.09) and in the game (a User X played as the setter in 0.01 of the games; in English 0.35), and the game separates the two sources in every readout of the causal tests (Section~\ref{sec:res_mechanism}). That every readout can be compared from such a separation is a further reason for reporting the causal tests of Qwen3.8-27B with its Chinese items.

\paragraph{Transfer between the languages.} The subspaces fitted in the two languages largely coincide: the squared overlap \(\lVert B_\ell^{\mathrm{zh}\top}B_\ell^{\mathrm{en}}\rVert_F^2/k\) of the eight fitted dimensions has a median of 0.60 over the layers (from 0.37 at L0 to 0.76 at L32), against \(k/d_{\mathrm{model}} = 8/5{,}120 = 0.0016\) for two random subspaces. Applied to the items of the other language without thinking (Table~\ref{tab:self_english_transfer}), each subspace removes the claim as fully as the subspace of the items' own language (ratios 0.96--1.01). Installation transfers fully from the Chinese subspace to the English items (1.12 for a User message; 1.71 for a System message, from small changes, \(+0.07\) and \(+0.12\); in the game, both subspaces make the model play the setter in every game) and in part from the English subspace to the Chinese items (0.59 for a User message, 0.68 for a System message, 0.77 in the game). These runs were made on one machine, on which the Chinese subspace was refitted by the same procedure; the refit reproduces the Chinese results (\(-0.76\) and \(+0.50\), against \(-0.76\) and \(+0.49\) in Table~\ref{tab:self_mechanism}).

\begin{table}[htbp]
\centering
\caption{Causal tests of Qwen3.8-27B with the English and with the Chinese items, replacement of the self-index coordinates at every layer (75 held-out bodies: attribution 150 items per condition, content and arithmetic 75, the secret game 100 games). Unedited rows: probability of ``you'' without thinking, share of finished items answered ``you'' with thinking, and in the game the share of finished games in which the model acts as the setter (in parentheses: finished of all, where not all finished). Other rows: item-paired change from the unedited message of the same role, with a 95\% bootstrap interval over items. Eight dimensions unless marked 64 (with the English items run with thinking only); copy: the unedited coordinates copied in place; random: the same edit in a random subspace of the same dimension. Content, arithmetic, and control: share of finished items answered correctly, lowest and highest over the conditions (with thinking, 432 of the 450 English content items finish). Chinese columns: the runs of Tables~\ref{tab:self_mechanism}, \ref{tab:self_mech_questions}, and~\ref{tab:self_game}.}
\label{tab:self_english}
\scriptsize
\setlength{\tabcolsep}{1.6pt}
\begin{tabular}{lcccc}
\toprule
& \multicolumn{2}{c}{No thinking} & \multicolumn{2}{c}{Thinking} \\
\cmidrule(lr){2-3}\cmidrule(lr){4-5}
Recipient \(\leftarrow\) donor & English & Chinese & English & Chinese \\
\midrule
\multicolumn{5}{l}{\emph{Attribution}} \\
own, unedited & 0.70 & 0.79 & 0.89 & 0.95 (149/150) \\
own \(\leftarrow\) User & \(-0.63\) [\(-0.66\), \(-0.61\)] & \(-0.76\) [\(-0.77\), \(-0.74\)] & \(-0.89\) [\(-0.93\), \(-0.83\)] & \(-0.95\) [\(-0.98\), \(-0.91\)] \\
own \(\leftarrow\) System & \(-0.55\) [\(-0.57\), \(-0.53\)] & \(-0.64\) [\(-0.65\), \(-0.63\)] & \(-0.89\) [\(-0.93\), \(-0.83\)] & \(-0.86\) [\(-0.92\), \(-0.80\)] \\
own \(\leftarrow\) User, random & \(-0.03\) [\(-0.04\), \(-0.03\)] & \(-0.01\) [\(-0.02\), \(-0.01\)] & \(-0.03\) [\(-0.07\), \(+0.01\)] & \(+0.02\) [\(-0.02\), \(+0.06\)] \\
own \(\leftarrow\) own (copy) & \(0.00\) [\(0.00\), \(0.00\)] & \(0.00\) [\(0.00\), \(0.00\)] & \(-0.04\) [\(-0.08\), \(0.00\)] & \(0.00\) [\(-0.04\), \(+0.04\)] \\
User, unedited & 0.08 & 0.04 & 0.00 & 0.00 \\
User \(\leftarrow\) own & \(+0.32\) [\(+0.30\), \(+0.34\)] & \(+0.49\) [\(+0.45\), \(+0.53\)] & \(+0.04\) [\(+0.01\), \(+0.07\)] & \(+0.60\) [\(+0.52\), \(+0.67\)] \\
User \(\leftarrow\) own, random & \(+0.01\) [\(+0.01\), \(+0.01\)] & \(+0.004\) [\(+0.002\), \(+0.005\)] & \(0.00\) [\(0.00\), \(0.00\)] & \(0.00\) [\(0.00\), \(0.00\)] \\
User \(\leftarrow\) own, 64 dims & -- & \(+0.81\) [\(+0.79\), \(+0.83\)] & \(+0.41\) [\(+0.33\), \(+0.49\)] & \(+0.97\) [\(+0.93\), \(+0.99\)] \\
User \(\leftarrow\) own, 64 dims, random & -- & \(+0.06\) [\(+0.05\), \(+0.06\)] & \(0.00\) [\(0.00\), \(0.00\)] & \(0.00\) [\(0.00\), \(0.00\)] \\
System, unedited & 0.09 & 0.07 & 0.00 (147/150) & 0.00 \\
System \(\leftarrow\) own & \(+0.07\) [\(+0.05\), \(+0.09\)] & \(+0.39\) [\(+0.35\), \(+0.42\)] & \(0.00\) [\(0.00\), \(0.00\)] & \(+0.06\) [\(+0.03\), \(+0.10\)] \\
Content, correct & 1.00 & 0.97--1.00 & 0.93--1.00 & 1.00 \\
Arithmetic, correct & 1.00 & 1.00 & 1.00 & 1.00 \\
\midrule
\multicolumn{5}{l}{\emph{Secret game}} \\
own X, unedited & 1.00 & 1.00 & 1.00 & 1.00 \\
own \(\leftarrow\) User & \(0.00\) [\(0.00\), \(0.00\)] & \(0.00\) [\(0.00\), \(0.00\)] & \(-0.04\) [\(-0.08\), \(-0.01\)] & \(-0.04\) [\(-0.08\), \(-0.01\)] \\
own \(\leftarrow\) User, random & \(0.00\) [\(0.00\), \(0.00\)] & \(0.00\) [\(0.00\), \(0.00\)] & \(0.00\) [\(0.00\), \(0.00\)] & \(0.00\) [\(0.00\), \(0.00\)] \\
User X, unedited & 0.35 & 0.01 & 0.00 & 0.00 (99/100) \\
User \(\leftarrow\) own & \(+0.65\) [\(+0.56\), \(+0.74\)] & \(+0.25\) [\(+0.17\), \(+0.34\)] & \(0.00\) [\(0.00\), \(0.00\)] & \(0.00\) [\(0.00\), \(0.00\)] \\
User \(\leftarrow\) own, random & \(+0.08\) [\(+0.03\), \(+0.14\)] & \(0.00\) [\(0.00\), \(0.00\)] & \(0.00\) [\(0.00\), \(0.00\)] & \(0.00\) [\(0.00\), \(0.00\)] \\
Control, correct & 1.00 & 1.00 & 1.00 & 1.00 \\
\bottomrule
\end{tabular}
\end{table}

\begin{table}[htbp]
\centering
\caption{Transfer of the subspaces between the languages, Qwen3.8-27B without thinking (75 held-out bodies; attribution 150 items per condition, the secret game 100 games). Item-paired change from the unedited message of the same role, with a 95\% bootstrap interval over items, with the subspace fitted on the discovery bodies of the items' own language and with that of the other language, and their ratio. All runs on one machine, with the Chinese subspace refitted there. The random subspaces are the same in both fits; with either, they change the claim by at most 0.03.}
\label{tab:self_english_transfer}
\scriptsize
\setlength{\tabcolsep}{3pt}
\begin{tabular}{lccc}
\toprule
Recipient \(\leftarrow\) donor & Own language & Other language & Ratio \\
\midrule
\multicolumn{4}{l}{\emph{Chinese items}} \\
\quad own \(\leftarrow\) User & \(-0.76\) [\(-0.77\), \(-0.75\)] & \(-0.72\) [\(-0.74\), \(-0.71\)] & 0.96 \\
\quad own \(\leftarrow\) System & \(-0.64\) [\(-0.65\), \(-0.62\)] & \(-0.64\) [\(-0.65\), \(-0.63\)] & 1.01 \\
\quad User \(\leftarrow\) own & \(+0.50\) [\(+0.46\), \(+0.53\)] & \(+0.29\) [\(+0.26\), \(+0.32\)] & 0.59 \\
\quad System \(\leftarrow\) own & \(+0.39\) [\(+0.35\), \(+0.42\)] & \(+0.26\) [\(+0.24\), \(+0.29\)] & 0.68 \\
\quad Secret game: User \(\leftarrow\) own & \(+0.26\) [\(+0.18\), \(+0.35\)] & \(+0.20\) [\(+0.12\), \(+0.28\)] & 0.77 \\
\midrule
\multicolumn{4}{l}{\emph{English items}} \\
\quad own \(\leftarrow\) User & \(-0.63\) [\(-0.66\), \(-0.61\)] & \(-0.64\) [\(-0.66\), \(-0.62\)] & 1.01 \\
\quad own \(\leftarrow\) System & \(-0.55\) [\(-0.57\), \(-0.53\)] & \(-0.55\) [\(-0.57\), \(-0.53\)] & 1.00 \\
\quad User \(\leftarrow\) own & \(+0.32\) [\(+0.30\), \(+0.34\)] & \(+0.36\) [\(+0.33\), \(+0.38\)] & 1.12 \\
\quad System \(\leftarrow\) own & \(+0.07\) [\(+0.05\), \(+0.09\)] & \(+0.12\) [\(+0.09\), \(+0.15\)] & 1.71 \\
\quad Secret game: User \(\leftarrow\) own & \(+0.65\) [\(+0.56\), \(+0.74\)] & \(+0.65\) [\(+0.56\), \(+0.74\)] & 1.00 \\
\bottomrule
\end{tabular}
\end{table}

\paragraph{Decoding and APIs.} In thinking mode, local runs generate greedily, up to 4,096 tokens for Qwen3.8-27B in Section~\ref{sec:res_thinking} and up to 3,072 tokens elsewhere. The thinking-mode self-reference of two models comes from the developer's API with the same inputs: GLM-5.3-Flash through the Z.ai API (model \texttt{glm-5.3-flash}, temperature 0, at most 3,072 tokens), and DeepSeek-V4.1-Flash through the DeepSeek API (model \texttt{deepseek-flash}, which the provider documents as served by DeepSeek-V4.1-Flash-1017; reasoning effort high, temperature 1.0 and top-$p$ 1.0, the settings the developer recommends, since the thinking mode offers no greedy decoding; at most 3,072 tokens). For every request, the number of prompt tokens reported by the API equals that of the local rendering.

\paragraph{Statistics.} All intervals are 95\% intervals with a fixed seed: percentile bootstrap intervals over wording families for the geometry (1,000 resamples) and, in Appendix~\ref{app:mech_cluster}, for the causal tests (4,000 resamples), over bodies or items elsewhere (4,000 to 10,000 resamples), and over models for the rank correlations of Appendix~\ref{app:benchmarks} (10,000 resamples); the shares of finished items in Figure~\ref{fig:self_mechanism} carry Wilson intervals. Each subspace of the replacement is fitted on the discovery bodies and stored with the random subspace of its control (the orthonormalized columns of a Gaussian matrix, drawn per layer with a fixed seed); every test that uses it reads the stored coordinates.

\clearpage
\section{Sensitivity of the Cross-Model Classification}
\label{app:threshold}

\paragraph{Threshold of the self-index criterion.} Table~\ref{tab:self_threshold} repeats the classification of Section~\ref{sec:res_crossmodel} with the self-index criterion of Definition~\ref{def:self} at thresholds of 70\% and 90\% of the held-out bodies instead of 80\%, and with the last quarter of the depth (\(\ell\ge 3L/4\)) beside the last third, from the same per-layer shares (message content and response onset, a layer counted once; four-stream models: mean of the per-stream ratios). The self-reference of each model does not involve the threshold and is unchanged. At every threshold, no model shows self-reference without a self-index. At 90\%, the same thirteen models carry a self-index as at 80\%. At 70\%, Gemma-4-31B, OLMo-2-13B, and Moonlight-16B-A3B carry one as well, at 2, 7, and 1 layers of the last third under the non-thinking rendering, and none recognizes its own messages. At 70\% and 80\%, every model and mode with self-reference singles out its own message at every layer of the last third and of the last quarter; at 90\%, all except GLM-5.3-Flash, whose mean of four streams does so at 8 of the 15 layers of the last third and 8 of the 11 of the last quarter, and whose single streams do so at 4 to 12 of the 15, one of them at all 11 of the last quarter.

\begin{table}[htbp]
\centering
\caption{The self-index criterion at three thresholds: the share of held-out bodies in which the model's own message must have the smallest distance ratio at some layer beyond the first third, on message content or at the response onset (80\%: Definition~\ref{def:self}). Classification under the rendering of Section~\ref{sec:res_crossmodel} (non-thinking; thinking for templates with only a thinking mode); for the bands of depth, the rendering of each mode in which the model recognizes its own messages (eight model--modes: Qwen3.8-27B in both modes and six models with thinking). Four-stream models: mean of the per-stream ratios. Last row: in either rendering.}
\label{tab:self_threshold}
\small
\begin{tabular}{>{\raggedright\arraybackslash}p{0.62\linewidth}ccc}
\toprule
& 70\% & 80\% & 90\% \\
\midrule
Models with a self-index & 16 & 13 & 13 \\
\quad with self-reference & 7 & 7 & 7 \\
\quad without self-reference & 9 & 6 & 6 \\
Models with neither level & 9 & 12 & 12 \\
Models with self-reference but no self-index & 0 & 0 & 0 \\
\midrule
Model--modes with self-reference (of 8) whose own message is singled out at every layer of the last third & 8 & 8 & 7 \\
\quad at every layer of the last quarter & 8 & 8 & 7 \\
Most layers of the last third at which a model without a self-index singles out its own message & 0 & 3/20 & 1/20 \\
\bottomrule
\end{tabular}
\end{table}

\paragraph{Option order.} Under the main probe, single-pass self-reference is the mean over the first two cyclic option orders (Section~\ref{sec:setup_crossmodel}). Table~\ref{tab:self_orders} gives it also as the balanced mean over all three orders, from the same items. No model changes its verdict under the margin of 0.10 of Definition~\ref{def:self}, and the nesting holds in both: Qwen3.8-27B claims its own messages by \(+0.61\) over three orders (\(+0.70\) over two), in all 150 bodies, and the largest excess without a self-index rises from \(+0.004\) (Seed-OSS-36B) to \(+0.06\) (OLMo-2-13B), whose own messages receive more ``you'' when it is the last option (Section~\ref{sec:res_crossmodel}). With thinking, the items were run in the first two orders only (300 items per source), so the three-order mean is given for the single pass only.

\begin{table}[htbp]
\centering
\caption{Single-pass self-reference under the main probe, over the first two cyclic option orders (as in Table~\ref{tab:self_crossmodel}) and over all three: excess of the probability of ``you'' for the model's own messages over the larger of that for User and System messages, from body-level means over the orders, and the number of the 150 bodies with a positive excess. Self-index: as in Table~\ref{tab:self_crossmodel}. Models with a non-thinking mode.}
\label{tab:self_orders}
\small
\begin{tabular}{lccccc}
\toprule
& & \multicolumn{2}{c}{Two orders} & \multicolumn{2}{c}{Three orders} \\
\cmidrule(lr){3-4}\cmidrule(lr){5-6}
Model & Self-index & excess & bodies & excess & bodies \\
\midrule
Qwen3.8-27B & yes & \(+0.70\) & 150 & \(+0.61\) & 150 \\
Qwen3.8-Flash-Next & yes & \(+0.01\) & 72 & \(+0.02\) & 76 \\
EXAONE-4.0.1-32B & yes & \(0.00\) & 78 & \(-0.04\) & 21 \\
MiMo-V2.6-Distill-9B & yes & \(+0.05\) & 123 & \(+0.01\) & 82 \\
GLM-4.6V-Flash & yes & \(+0.01\) & 72 & \(+0.01\) & 78 \\
GLM-4.7-Flash & yes & \(+0.05\) & 130 & \(+0.03\) & 118 \\
phi-4 & yes & \(+0.01\) & 104 & \(-0.01\) & 63 \\
Granite-4.2-30B & yes & \(0.00\) & 62 & \(0.00\) & 58 \\
Granite-4.1-30B & yes & \(-0.01\) & 99 & \(-0.01\) & 100 \\
OLMo-2-13B & no & \(-0.06\) & 3 & \(+0.06\) & 144 \\
OLMo-2-7B & no & \(-0.04\) & 17 & \(-0.07\) & 0 \\
Apertus-8B & no & \(-0.05\) & 6 & \(+0.01\) & 104 \\
Moonlight-16B-A3B & no & \(+0.06\) & 136 & \(0.00\) & 87 \\
MiMo-7B-RL & no & \(-0.04\) & 15 & \(-0.01\) & 51 \\
Seed-OSS-36B & no & \(0.00\) & 39 & \(0.00\) & 52 \\
Gemma-4-31B & no & \(0.00\) & 103 & \(0.00\) & 65 \\
Qwen3.6-27B & no & \(-0.06\) & 9 & \(-0.07\) & 1 \\
Qwen3.6-35B-A3B & no & \(-0.03\) & 11 & \(-0.03\) & 3 \\
Qwen3.5-27B & no & \(-0.10\) & 0 & \(-0.09\) & 1 \\
\bottomrule
\end{tabular}
\end{table}

\clearpage
\section{Published Benchmark Scores}
\label{app:benchmarks}

Table~\ref{tab:self_benchmarks} lists the GPQA scores that the developers report for the models of the comparison, the benchmark they most often report, next to the group of each model in Section~\ref{sec:res_crossmodel}. The scores come from the Hugging Face model card of each model at the revision used here and, for the gpt-oss models, from their model card \cite{openai2025gptoss}; the subset of GPQA and the setting are those the developers state. Nine cards report no GPQA score: those of OLMo-2-7B and OLMo-2-13B (MMLU 61.3 and 68.5), of GLM-4.6V-Flash (multimodal benchmarks only), of GLM-5.3-Flash (agentic benchmarks only), of ERNIE-4.5-21B-A3B-Thinking (scores shown only as an image), and of MiMo-VL-7B-RL-2508, MiMo-V2.6-Distill-Qwen-9B, Moonlight-16B-A3B-Instruct, and Apertus-8B-Instruct-2509. Within each group of more than one model the scores span a wide range, and the four highest scores of the models without self-reference belong to models with neither level under the main probe. Over the 16 models with a score, the group of a model and the extent of its self-index have small rank correlations with the score ($\rho = 0.10$ and $0.13$); for the presence and the strength of self-reference, $\rho = 0.39$ and $0.43$ ($p = 0.15$ and $0.09$), and with the measurements in natural order included every $\rho$ lies between $0.03$ and $0.44$; every 95\% interval includes zero (Table~\ref{tab:self_benchmark_corr}).

\begin{table}[htbp]
\centering
\caption{GPQA scores as published by the developers, with the group of each model and its two levels under the main probe (Section~\ref{sec:res_crossmodel}). Label: the name of the benchmark in the source; setting: as stated in the source (empty where none is stated). Index: share of layers at which the model's own message is singled out in at least 80\% of the held-out bodies at message content or at the response onset, under the non-thinking rendering (thinking rendering for models with only a thinking mode), as in Figure~\ref{fig:self_crossmodel}. Self-reference: the model's own messages minus the larger of User and System messages, in the mode in which it is larger (probability of ``you'' without thinking, share of finished items answered ``you'' with thinking), as in Figure~\ref{fig:self_crossmodel}.}
\label{tab:self_benchmarks}
{\footnotesize
\setlength{\tabcolsep}{3.5pt}
\begin{tabular}{llrl>{\raggedright\arraybackslash}p{2.7cm}rr}
\toprule
Group & Model & GPQA & Label & Setting & Index & Self-ref. \\
\midrule
Both levels & Qwen3.8-27B & 89.2 & GPQA Diamond &  & 1.00 & +0.96 \\
\midrule
Self-reference & Qwen3.8-Flash-Next & 91.7 & GPQA Diamond &  & 1.00 & +0.66 \\
after reasoning & gpt-oss-120b & 80.1 & GPQA Diamond & no tools, reasoning high & 0.94 & +0.97 \\
 & EXAONE-4.0.1-32B & 74.3 & GPQA-Diamond & reasoning mode (61.6 without) & 1.00 & +0.12 \\
 & gpt-oss-20b & 71.5 & GPQA Diamond & no tools, reasoning high & 1.00 & +0.78 \\
\midrule
Self-index without & GLM-4.7-Flash & 75.2 & GPQA &  & 1.00 & +0.06 \\
self-reference & Granite-4.2-30B & 66.41 & GPQA &  & 0.92 & +0.00 \\
 & R1-Distill-Llama-70B & 65.2 & GPQA Diamond & pass@1 & 1.00 & $-$0.07 \\
 & Granite-4.1-30B & 45.76 & GPQA & 0-shot, CoT & 0.58 & $-$0.01 \\
 & phi-4 & 56.1 & GPQA &  & 0.88 & +0.01 \\
\midrule
Neither level & Qwen3.6-27B & 87.8 & GPQA Diamond &  & 0.09 & +0.00 \\
 & Qwen3.6-35B-A3B & 86.0 & GPQA &  & 0.03 & +0.00 \\
 & Qwen3.5-27B & 85.5 & GPQA Diamond &  & 0.20 & +0.02 \\
 & Gemma-4-31B & 84.3 & GPQA Diamond &  & 0.00 & +0.03 \\
 & Seed-OSS-36B & 71.4 & GPQA-D &  & 0.00 & +0.00 \\
 & MiMo-7B-RL & 60.6 & GPQA-Diamond & Pass@1 & 0.00 & +0.01 \\
\bottomrule
\end{tabular}}

\vspace{1.2em}
\caption{Rank correlation between the published GPQA score and the two levels. Main probe: the 16 models of Table~\ref{tab:self_benchmarks}. With natural order: the measurements in natural order (Appendix~\ref{app:natural}) included, each model weighted equally: a model measured in both orders contributes the mean of its two values, and Ministral-3-14B-Reasoning, measured in natural order only, is added to the self-index (GPQA Diamond 71.2 in its model card); in natural order, self-reference is averaged over the option orders of Appendix~\ref{app:natural} (three where three roles are compared, two otherwise) and, for the models compared on two roles, taken against User messages. Spearman $\rho$, two-sided permutation $p$ (100,000 permutations), and 95\% bootstrap interval over models (10,000 resamples). Group: neither level, self-index without self-reference, self-reference after reasoning, both levels, in this order; presence of self-reference: one of the last two groups.}
\label{tab:self_benchmark_corr}
{\footnotesize
\setlength{\tabcolsep}{4pt}
\begin{tabular}{lrrcrrrc}
\toprule
& \multicolumn{3}{c}{Main probe ($n = 16$)} & \multicolumn{4}{c}{With natural order} \\
\cmidrule(lr){2-4}\cmidrule(lr){5-8}
Level & $\rho$ & $p$ & 95\% interval & $n$ & $\rho$ & $p$ & 95\% interval \\
\midrule
Group & +0.10 & 0.70 & [$-$0.51, +0.61] & 16 & +0.22 & 0.42 & [$-$0.38, +0.69] \\
Self-index & +0.13 & 0.63 & [$-$0.41, +0.59] & 17 & +0.03 & 0.92 & [$-$0.53, +0.52] \\
Self-reference, presence & +0.39 & 0.15 & [$-$0.10, +0.75] & 16 & +0.44 & 0.09 & [$-$0.04, +0.77] \\
Self-reference, strength & +0.43 & 0.09 & [$-$0.14, +0.78] & 16 & +0.35 & 0.18 & [$-$0.24, +0.78] \\
\bottomrule
\end{tabular}}
\end{table}

\clearpage
\section{Natural Conversation Order}
\label{app:natural}

The main probe opens the conversation with the message under test. This is the only position that every template of the comparison accepts for an Assistant, a User, and a System message before the same question; the main text rests on it. A model's own messages, however, normally answer a user request. As a control for the position of the message, we therefore repeated the measurements with the message under test after a user request, [user: request] [\textit{role}: body] [user: question], for one model of each developer in the comparison, the smallest of its models (for Alibaba, Qwen3.8-27B, which is as large as Qwen3.6-27B and Qwen3.5-27B; for IBM, whose two models have the same size, Granite-4.1-30B; for Ai2, OLMo-2-7B; for Xiaomi, MiMo-7B-RL-0530).

\begin{figure}[htbp]
\centering
\includegraphics[width=0.8\linewidth]{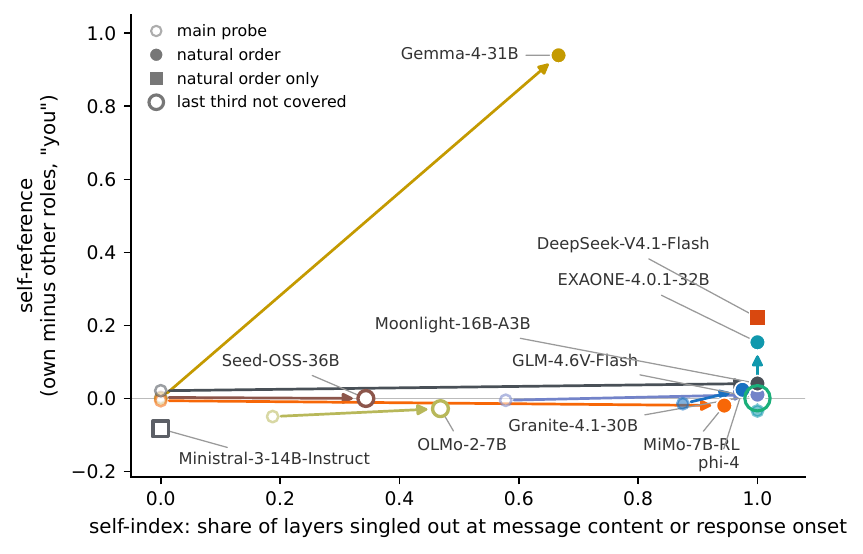}
\caption{The two levels in natural conversation order (single pass), on the axes of Figure~\ref{fig:self_crossmodel}a, with self-reference averaged over the three option orders as in Table~\ref{tab:self_natural}. Arrows run from the main probe (faint points) to the natural order for the nine models with a single pass whose templates keep the three roles. Squares mark the two models whose templates do not accept the main probe, DeepSeek-V4.1-Flash in chat mode and Ministral-3-14B-Instruct (Table~\ref{tab:self_natural3}). Filled markers: the layers singled out include every layer of the last third of the depth. The ring of GLM-4.6V-Flash marks its point on that of Granite-4.1-30B.}
\label{fig:self_natural}
\end{figure}

For ten developers the template keeps a System message in place after a user turn, so the three roles can be compared as in the main probe: Seed-OSS-36B, Gemma-4-31B, EXAONE-4.0.1-32B, GLM-4.6V-Flash, Granite-4.1-30B, OLMo-2-7B, ERNIE-4.5-21B-A3B-Thinking, MiMo-7B-RL-0530, Moonlight-16B-A3B-Instruct, and phi-4 (Table~\ref{tab:self_natural}; Figure~\ref{fig:self_natural}). In this order, too, every model that claims its own messages carries a self-index. Two models claim their own messages in a single pass, which neither does under the main probe: EXAONE-4.0.1-32B (0.34, against at most 0.19), which singles out its own message at all 64 layers of message content and of the response onset, and Gemma-4-31B (1.00, against at most 0.06), which has neither level under the main probe and in this order singles out its own message at 40 of its 60 layers of the response onset, including every layer of the last third. The other models with a single pass do not claim their own messages: GLM-4.6V-Flash (0.36, against 0.36 and 0.35), Granite-4.1-30B (0.01), OLMo-2-7B (0.24, against 0.21 and 0.27), Seed-OSS-36B (0.00), MiMo-7B-RL (0.13, against 0.12 and 0.15), Moonlight-16B-A3B (0.70, against 0.66 and 0.65), and phi-4 (0.27, against 0.24 and 0.24). In this order MiMo-7B-RL, Moonlight-16B-A3B, and ERNIE-4.5-21B-A3B, which carry no self-index under the main probe, single out their own message at most of their layers; ERNIE, whose template has only a thinking mode, was measured for the self-index only.

\begin{table}[htbp]
\centering
\caption{Natural conversation order: the message under test follows a user request. Self-index: layers at which the own message is singled out in at least 80\% of the held-out bodies, on message content and at the response onset. Probability of ``you'': mean over 150 bodies and three option orders. $^{a}$Thinking rendering, self-index only.}
\label{tab:self_natural}
\footnotesize
\begin{tabular}{lccccc}
\toprule
& \multicolumn{2}{c}{Self-index} & \multicolumn{3}{c}{Probability of ``you''} \\
\cmidrule(lr){2-3}\cmidrule(lr){4-6}
Model & content & onset & own & User & System \\
\midrule
EXAONE-4.0.1-32B & 64/64 & 64/64 & 0.34 & 0.14 & 0.19 \\
Gemma-4-31B & 0/60 & 40/60 & 1.00 & 0.00 & 0.06 \\
GLM-4.6V-Flash & 36/40 & 16/40 & 0.36 & 0.36 & 0.35 \\
Granite-4.1-30B & 61/64 & 54/64 & 0.01 & 0.00 & 0.00 \\
Seed-OSS-36B & 3/64 & 22/64 & 0.00 & 0.00 & 0.00 \\
OLMo-2-7B & 5/32 & 15/32 & 0.24 & 0.21 & 0.27 \\
ERNIE-4.5-21B-A3B$^{a}$ & 16/28 & 28/28 & -- & -- & -- \\
MiMo-7B-RL & 34/36 & 27/36 & 0.13 & 0.12 & 0.15 \\
Moonlight-16B-A3B & 27/27 & 27/27 & 0.70 & 0.66 & 0.65 \\
phi-4 & 25/40 & 37/40 & 0.27 & 0.24 & 0.24 \\
\bottomrule
\end{tabular}
\end{table}

For the other four developers the template rejects a System message after a user turn (Qwen, Apertus), leaves it out (gpt-oss), or moves it to the start of the input (DeepSeek-R1-Distill-Llama-70B; Appendix~\ref{app:templates_natural}), so we compared Assistant and User alone, in the same order and with the same questions. Without the third role the distance ratios are undefined, so only self-reference is measured, in the mode in which Table~\ref{tab:self_crossmodel} reports it (Table~\ref{tab:self_natural2}). Qwen3.8-27B claims its own messages in a single pass with probability 0.76, against 0.02 for User messages, and more strongly than the User message in all 150 bodies. With thinking, gpt-oss-20b answers ``you'' for 0.93 of its own messages and 0.03 of User messages. DeepSeek-R1-Distill-Llama-70B, which under the main probe answers ``you'' less often for its own messages than for User messages, in this order answers it for 0.34 of its own messages and 0.16 of User messages. These three carry a self-index in the last third of their layers under the main probe, and Apertus-8B, which carries none, does not claim its own messages in this order either (0.22, against 0.24). The natural order thus covers every developer in the comparison, and in none of the fourteen models does self-reference appear without a self-index: where the order strengthens the claim to the model's own messages, in EXAONE-4.0.1-32B, Gemma-4-31B, DeepSeek-R1-Distill-Llama-70B, and gpt-oss-20b, the model has a self-index.

\begin{table}[htbp]
\centering
\caption{Natural conversation order with Assistant and User only, for the developers whose templates reject, leave out, or move a System message after a user turn. Single pass: mean probability of ``you'' over 150 bodies and the two option orders of Table~\ref{tab:self_crossmodel}. Thinking: share of finished items answered ``you'' (150 bodies, two option orders; 300 items per source). Main probe: the same quantities under the main probe, with User messages alone.}
\label{tab:self_natural2}
\footnotesize
\begin{tabular}{llcccc}
\toprule
& & \multicolumn{2}{c}{Natural order} & \multicolumn{2}{c}{Main probe} \\
\cmidrule(lr){3-4}\cmidrule(lr){5-6}
Model & Mode & own & User & own & User \\
\midrule
Qwen3.8-27B & single pass & 0.76 & 0.02 & 0.77 & 0.04 \\
gpt-oss-20b & thinking & 0.93 & 0.03 & 0.78 & 0.00 \\
R1-Distill-Llama-70B & thinking & 0.34 & 0.16 & 0.09 & 0.14 \\
Apertus-8B & single pass & 0.22 & 0.24 & 0.18 & 0.23 \\
\bottomrule
\end{tabular}
\end{table}

\subsection{Templates That Do Not Accept the Main Probe}
\label{app:natural_only}

Two further models cannot be tested with the main probe, because their templates do not accept a conversation that opens with an Assistant message: the encoder of DeepSeek-V4.1-Flash treats the Assistant marker as the end of the preceding user turn, and the template of Ministral-3-14B (Instruct and Reasoning) requires a user or system message first. They are not part of the comparison of twenty-five models, which is restricted to templates that render the main probe; Section~\ref{sec:res_secondary} summarizes them, and we report them here in full, in natural order only. In both models a User message that follows the request is merged with it into one user turn, so their User condition is not a separate message. DeepSeek-V4.1-Flash, tested with the Chinese bodies, singles out its own message at 38 of its 40 layers of message content and at all 40 of the response onset (mean of its four residual streams), and at 39 and 38 under its thinking rendering; in a single pass it gives ``you'' a probability of 0.23 for its own messages and at most 0.01 for external ones, and with thinking, through the developer's API for the same release, it answers ``you'' for 0.99 of its own messages and for 0.03 of User and 0.02 of System messages. Ministral-3-14B, tested with the English bodies, singles out its own message at none of its 40 layers in either version, and the Instruct version does not claim its own messages (0.34, against 0.31 and 0.43) (Table~\ref{tab:self_natural3}). Both models follow the nesting of the two levels.

\begin{table}[htbp]
\centering
\caption{Models whose templates do not accept the main probe, natural conversation order only. Self-index as in Table~\ref{tab:self_natural} (DeepSeek-V4.1-Flash: mean of its four residual streams). Single pass: mean probability of ``you'' over 150 bodies and three option orders; thinking: share of finished items answered ``you'' (300 items per source). User: the User message merged by the template with the preceding request into one user turn.}
\label{tab:self_natural3}
\footnotesize
\begin{tabular}{llccccc}
\toprule
& & \multicolumn{2}{c}{Self-index} & \multicolumn{3}{c}{``you''} \\
\cmidrule(lr){3-4}\cmidrule(lr){5-7}
Model & Mode & content & onset & own & User & System \\
\midrule
DeepSeek-V4.1-Flash & single pass & 38/40 & 40/40 & 0.23 & 0.00 & 0.01 \\
DeepSeek-V4.1-Flash & thinking & 39/40 & 38/40 & 0.99 & 0.03 & 0.02 \\
Ministral-3-14B-Instruct & single pass & 0/40 & 0/40 & 0.34 & 0.31 & 0.43 \\
Ministral-3-14B-Reasoning & geometry only & 0/40 & 0/40 & -- & -- & -- \\
\bottomrule
\end{tabular}
\end{table}

\clearpage
\section{Further Causal Tests}
\label{app:mechanism}

This appendix reports the numbers of Section~\ref{sec:mechanisms} that the main text does not show in full, in the order in which the main text cites them: the groups of positions and the layer windows of the causal tests, the first-order sensitivity by layer, the random subspaces, the exchange of the two external sources, the inverted replacement, the intervals over wording families and the bounds for unfinished outputs, the edit without the question and the stricter versions of both tasks, the replacement under the format ablation, and the geometry under position compression.

\subsection{Groups of Positions and Layer Windows}
\label{app:mech_where}

Tables~\ref{tab:self_mechanism_where} and~\ref{tab:self_mechanism_windows} give the numbers plotted in Figure~\ref{fig:self_mechanism_where} for the four readouts.

\begin{table}[htbp]
\centering
\caption{Replacement at every layer in one group of positions: item-paired change in ``you'' from the unedited message (150 items per condition; 95\% bootstrap intervals over items). Own \(\leftarrow\) User: the model's own message with User coordinates; User \(\leftarrow\) own: a User message with the model's own; random: own \(\leftarrow\) User in a random 8-dimensional subspace.}
\label{tab:self_mechanism_where}
\scriptsize
\setlength{\tabcolsep}{2.4pt}
\begin{tabular}{llcccc}
\toprule
& & Qwen3.8-27B & Qwen3.8-27B & gpt-oss-20b & gpt-oss-120b \\
Edited & Condition & no thinking & thinking & thinking & thinking \\
\midrule
body & own \(\leftarrow\) User & \(-0.16\) [\(-0.19\), \(-0.14\)] & \(-0.38\) [\(-0.47\), \(-0.30\)] & \(-0.65\) [\(-0.73\), \(-0.57\)] & \(-0.40\) [\(-0.48\), \(-0.32\)] \\
 & User \(\leftarrow\) own & \(+0.01\) [\(+0.01\), \(+0.02\)] & \(0.00\) [\(0.00\), \(0.00\)] & \(+0.01\) [\(-0.01\), \(+0.04\)] & \(0.00\) [\(0.00\), \(0.00\)] \\
 & random & \(-0.004\) [\(-0.006\), \(-0.001\)] & \(-0.01\) [\(-0.04\), \(+0.03\)] & \(+0.02\) [\(-0.03\), \(+0.07\)] & \(+0.01\) [\(-0.02\), \(+0.05\)] \\
\midrule
end of message & own \(\leftarrow\) User & \(-0.73\) [\(-0.74\), \(-0.72\)] & \(-0.92\) [\(-0.96\), \(-0.87\)] & \(-0.69\) [\(-0.77\), \(-0.62\)] & \(-0.77\) [\(-0.83\), \(-0.70\)] \\
 & User \(\leftarrow\) own & \(+0.30\) [\(+0.27\), \(+0.33\)] & \(+0.50\) [\(+0.42\), \(+0.58\)] & \(+0.03\) [\(0.00\), \(+0.07\)] & \(+0.10\) [\(+0.05\), \(+0.15\)] \\
 & random & \(-0.01\) [\(-0.01\), \(0.00\)] & \(+0.01\) [\(-0.02\), \(+0.05\)] & \(-0.01\) [\(-0.05\), \(+0.04\)] & \(-0.01\) [\(-0.05\), \(+0.03\)] \\
\midrule
question & own \(\leftarrow\) User & \(-0.52\) [\(-0.53\), \(-0.50\)] & \(-0.01\) [\(-0.07\), \(+0.03\)] & \(-0.04\) [\(-0.10\), \(+0.02\)] & \(-0.13\) [\(-0.19\), \(-0.08\)] \\
 & User \(\leftarrow\) own & \(+0.10\) [\(+0.08\), \(+0.12\)] & \(0.00\) [\(0.00\), \(0.00\)] & \(-0.01\) [\(-0.02\), \(0.00\)] & \(+0.01\) [\(0.00\), \(+0.02\)] \\
 & random & \(-0.003\) [\(-0.005\), \(0.000\)] & \(+0.01\) [\(-0.02\), \(+0.05\)] & \(-0.02\) [\(-0.07\), \(+0.02\)] & \(-0.01\) [\(-0.03\), \(+0.02\)] \\
\midrule
all & own \(\leftarrow\) User & \(-0.76\) [\(-0.77\), \(-0.74\)] & \(-0.95\) [\(-0.98\), \(-0.91\)] & \(-0.75\) [\(-0.82\), \(-0.69\)] & \(-0.92\) [\(-0.96\), \(-0.87\)] \\
 & User \(\leftarrow\) own & \(+0.49\) [\(+0.45\), \(+0.53\)] & \(+0.60\) [\(+0.52\), \(+0.67\)] & \(+0.51\) [\(+0.43\), \(+0.59\)] & \(+0.39\) [\(+0.31\), \(+0.47\)] \\
 & random & \(-0.01\) [\(-0.02\), \(-0.01\)] & \(+0.02\) [\(-0.02\), \(+0.06\)] & \(+0.05\) [\(-0.01\), \(+0.11\)] & \(-0.01\) [\(-0.05\), \(+0.03\)] \\
\bottomrule
\end{tabular}
\end{table}

\begin{table}[htbp]
\centering
\caption{Replacement at every position in one window of eight layers: item-paired change in ``you'' from the unedited message (150 items per condition; 95\% bootstrap intervals over items). Conditions as in Table~\ref{tab:self_mechanism_where}.}
\label{tab:self_mechanism_windows}
\scriptsize
\setlength{\tabcolsep}{3pt}
\begin{tabular}{lccc}
\toprule
Window & own \(\leftarrow\) User & User \(\leftarrow\) own & own \(\leftarrow\) User, random \\
\midrule
\multicolumn{4}{l}{\emph{Qwen3.8-27B, no thinking}} \\
L0--L7 & \(-0.14\) [\(-0.16\), \(-0.12\)] & \(+0.01\) [\(0.00\), \(+0.01\)] & \(0.000\) [\(-0.002\), \(+0.003\)] \\
L4--L11 & \(-0.23\) [\(-0.25\), \(-0.20\)] & \(+0.01\) [\(+0.01\), \(+0.02\)] & \(-0.002\) [\(-0.004\), \(+0.001\)] \\
L8--L15 & \(-0.30\) [\(-0.32\), \(-0.27\)] & \(+0.01\) [\(+0.01\), \(+0.01\)] & \(-0.001\) [\(-0.003\), \(+0.001\)] \\
L12--L19 & \(-0.47\) [\(-0.50\), \(-0.44\)] & \(+0.002\) [\(-0.002\), \(+0.006\)] & \(+0.01\) [\(0.00\), \(+0.01\)] \\
L16--L23 & \(-0.57\) [\(-0.59\), \(-0.55\)] & \(+0.04\) [\(+0.03\), \(+0.05\)] & \(+0.003\) [\(+0.001\), \(+0.006\)] \\
L20--L27 & \(-0.75\) [\(-0.76\), \(-0.73\)] & \(+0.07\) [\(+0.06\), \(+0.09\)] & \(-0.004\) [\(-0.007\), \(-0.001\)] \\
L24--L31 & \(-0.76\) [\(-0.78\), \(-0.75\)] & \(+0.18\) [\(+0.16\), \(+0.21\)] & \(-0.001\) [\(-0.004\), \(+0.002\)] \\
L28--L35 & \(-0.77\) [\(-0.78\), \(-0.76\)] & \(+0.76\) [\(+0.73\), \(+0.78\)] & \(0.000\) [\(-0.003\), \(+0.002\)] \\
L32--L39 & \(-0.77\) [\(-0.78\), \(-0.76\)] & \(+0.68\) [\(+0.64\), \(+0.72\)] & \(+0.01\) [\(0.00\), \(+0.01\)] \\
L36--L43 & \(-0.69\) [\(-0.70\), \(-0.68\)] & \(+0.44\) [\(+0.40\), \(+0.49\)] & \(-0.001\) [\(-0.003\), \(0.000\)] \\
L40--L47 & \(-0.43\) [\(-0.45\), \(-0.41\)] & \(+0.22\) [\(+0.19\), \(+0.25\)] & \(-0.02\) [\(-0.02\), \(-0.02\)] \\
L44--L51 & \(-0.07\) [\(-0.08\), \(-0.07\)] & \(+0.004\) [\(+0.003\), \(+0.006\)] & \(-0.01\) [\(-0.02\), \(-0.01\)] \\
L48--L55 & \(-0.04\) [\(-0.04\), \(-0.03\)] & \(+0.01\) [\(+0.01\), \(+0.01\)] & \(-0.004\) [\(-0.006\), \(-0.003\)] \\
L52--L59 & \(-0.04\) [\(-0.04\), \(-0.03\)] & \(+0.01\) [\(+0.01\), \(+0.01\)] & \(-0.003\) [\(-0.004\), \(-0.001\)] \\
L56--L63 & \(0.000\) [\(-0.002\), \(+0.002\)] & \(+0.004\) [\(+0.003\), \(+0.005\)] & \(0.000\) [\(-0.001\), \(+0.001\)] \\
\midrule
\multicolumn{4}{l}{\emph{Qwen3.8-27B, thinking}} \\
L0--L7 & \(-0.61\) [\(-0.69\), \(-0.54\)] & \(0.00\) [\(0.00\), \(0.00\)] & \(+0.02\) [\(-0.01\), \(+0.05\)] \\
L4--L11 & \(-0.65\) [\(-0.73\), \(-0.57\)] & \(0.00\) [\(0.00\), \(0.00\)] & \(0.00\) [\(-0.04\), \(+0.04\)] \\
L8--L15 & \(-0.72\) [\(-0.79\), \(-0.64\)] & \(0.00\) [\(0.00\), \(0.00\)] & \(+0.03\) [\(0.00\), \(+0.06\)] \\
L12--L19 & \(-0.67\) [\(-0.75\), \(-0.60\)] & \(+0.02\) [\(0.00\), \(+0.05\)] & \(+0.03\) [\(0.00\), \(+0.06\)] \\
L16--L23 & \(-0.72\) [\(-0.79\), \(-0.63\)] & \(0.00\) [\(0.00\), \(0.00\)] & \(+0.04\) [\(+0.01\), \(+0.07\)] \\
L20--L27 & \(-0.86\) [\(-0.91\), \(-0.79\)] & \(+0.01\) [\(0.00\), \(+0.02\)] & \(+0.01\) [\(-0.02\), \(+0.04\)] \\
L24--L31 & \(-0.95\) [\(-0.99\), \(-0.92\)] & \(+0.02\) [\(0.00\), \(+0.05\)] & \(+0.03\) [\(+0.01\), \(+0.07\)] \\
L28--L35 & \(-0.95\) [\(-0.99\), \(-0.92\)] & \(+0.57\) [\(+0.49\), \(+0.65\)] & \(-0.01\) [\(-0.05\), \(+0.03\)] \\
L32--L39 & \(-0.93\) [\(-0.97\), \(-0.89\)] & \(+0.39\) [\(+0.31\), \(+0.47\)] & \(+0.01\) [\(-0.01\), \(+0.03\)] \\
L36--L43 & \(-0.49\) [\(-0.58\), \(-0.40\)] & \(+0.13\) [\(+0.08\), \(+0.19\)] & \(+0.01\) [\(-0.02\), \(+0.04\)] \\
L40--L47 & \(-0.43\) [\(-0.52\), \(-0.34\)] & \(+0.01\) [\(0.00\), \(+0.03\)] & \(0.00\) [\(-0.03\), \(+0.03\)] \\
L44--L51 & \(-0.17\) [\(-0.24\), \(-0.09\)] & \(0.00\) [\(0.00\), \(0.00\)] & \(+0.01\) [\(-0.02\), \(+0.03\)] \\
L48--L55 & \(-0.02\) [\(-0.06\), \(+0.02\)] & \(0.00\) [\(0.00\), \(0.00\)] & \(0.00\) [\(-0.03\), \(+0.03\)] \\
L52--L59 & \(+0.02\) [\(0.00\), \(+0.05\)] & \(0.00\) [\(0.00\), \(0.00\)] & \(+0.01\) [\(-0.01\), \(+0.03\)] \\
L56--L63 & \(+0.01\) [\(-0.01\), \(+0.04\)] & \(0.00\) [\(0.00\), \(0.00\)] & \(+0.01\) [\(0.00\), \(+0.03\)] \\
\midrule
\multicolumn{4}{l}{\emph{gpt-oss-20b, thinking}} \\
L0--L7 & \(-0.73\) [\(-0.80\), \(-0.66\)] & \(+0.21\) [\(+0.15\), \(+0.28\)] & \(+0.01\) [\(-0.03\), \(+0.06\)] \\
L4--L11 & \(-0.76\) [\(-0.83\), \(-0.69\)] & \(+0.36\) [\(+0.28\), \(+0.44\)] & \(-0.02\) [\(-0.06\), \(+0.02\)] \\
L8--L15 & \(-0.75\) [\(-0.82\), \(-0.69\)] & \(+0.57\) [\(+0.49\), \(+0.65\)] & \(+0.01\) [\(-0.03\), \(+0.06\)] \\
L12--L19 & \(-0.39\) [\(-0.47\), \(-0.31\)] & \(+0.07\) [\(+0.03\), \(+0.11\)] & \(+0.01\) [\(-0.05\), \(+0.07\)] \\
L16--L23 & \(0.00\) [\(-0.07\), \(+0.07\)] & \(0.00\) [\(0.00\), \(0.00\)] & \(+0.06\) [\(+0.02\), \(+0.11\)] \\
\midrule
\multicolumn{4}{l}{\emph{gpt-oss-120b, thinking}} \\
L0--L7 & \(-0.41\) [\(-0.49\), \(-0.33\)] & \(0.00\) [\(0.00\), \(0.00\)] & \(+0.02\) [\(-0.01\), \(+0.05\)] \\
L4--L11 & \(-0.90\) [\(-0.95\), \(-0.85\)] & \(0.00\) [\(0.00\), \(0.00\)] & \(+0.01\) [\(-0.01\), \(+0.03\)] \\
L8--L15 & \(-0.93\) [\(-0.97\), \(-0.88\)] & \(+0.13\) [\(+0.08\), \(+0.19\)] & \(0.00\) [\(-0.03\), \(+0.03\)] \\
L12--L19 & \(-0.88\) [\(-0.93\), \(-0.83\)] & \(+0.88\) [\(+0.83\), \(+0.93\)] & \(0.00\) [\(-0.03\), \(+0.03\)] \\
L16--L23 & \(-0.63\) [\(-0.71\), \(-0.56\)] & \(+0.35\) [\(+0.27\), \(+0.43\)] & \(-0.01\) [\(-0.03\), \(+0.02\)] \\
L20--L27 & \(-0.12\) [\(-0.17\), \(-0.07\)] & \(0.00\) [\(0.00\), \(0.00\)] & \(0.00\) [\(-0.03\), \(+0.03\)] \\
L24--L31 & \(0.00\) [\(-0.03\), \(+0.03\)] & \(0.00\) [\(0.00\), \(0.00\)] & \(0.00\) [\(-0.03\), \(+0.03\)] \\
L28--L35 & \(-0.01\) [\(-0.03\), \(+0.02\)] & \(0.00\) [\(0.00\), \(0.00\)] & \(-0.01\) [\(-0.03\), \(+0.02\)] \\
\bottomrule
\end{tabular}
\end{table}

\FloatBarrier
\subsection{First-Order Sensitivity by Layer}
\label{app:mech_fd}

\begin{figure}[htbp]
\centering
\includegraphics[width=\linewidth]{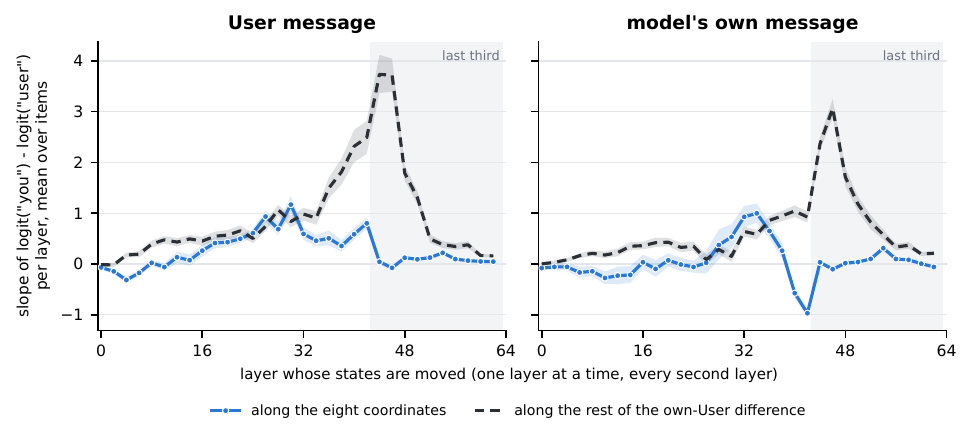}
\caption{First-order sensitivity of the attribution answer by layer, Qwen3.8-27B without thinking (150 items, every second layer). Slope of the logit difference between the letters for ``you'' and ``user'' at the response onset when the states of one layer are moved along the part of the own--User difference in the eight coordinates (blue) or along the rest of it (dark, dashed), mean over items with 95\% bootstrap intervals; left, at the User message; right, at the model's own message; shaded, the last third of the layers.}
\label{fig:self_fd}
\end{figure}

The layer windows of Table~\ref{tab:self_mechanism_windows} replace the coordinates over eight layers at a time. The first-order measure of Section~\ref{sec:res_mechanism} asks instead how strongly the answer responds to the coordinates at each layer. On each held-out item, \(d_\ell\) is the difference between the states of the model's own message and of the same text as a User message at the aligned positions of layer \(\ell\). At one layer at a time, these states are moved by \(\pm\epsilon v\), \(\epsilon=0.25\), along \(v=d_\ell\), along its part in the subspace, \(d_\ell B_\ell B_\ell^{\top}\), along the rest, and along its part in the random subspace of Table~\ref{tab:self_mechanism}; the later layers are recomputed, and the slope along \(v\) is \((f_{+}-f_{-})/2\epsilon\), where \(f\) is the logit of the letter for ``you'' minus that of the letter for ``user'' where the answer is read: at the response onset without thinking, and with thinking after the model's own reasoning, which is held fixed (teacher-forced). Only forward passes are used, with the weights and kernels of the main runs, since the MXFP4 expert kernels of gpt-oss have no backward pass; in Qwen3.8-27B every second layer is computed. The measure is taken at the User message and at the model's own message. Table~\ref{tab:self_fd} sums the slopes over the layers of a band and averages them over items; the share of the eight coordinates is the summed slope along their part divided by that along the whole difference, given where the summed slope is positive with its 95\% interval above zero. Figure~\ref{fig:self_fd} shows the layers of Qwen3.8-27B without thinking.

With thinking, the answer lies far from the point at which it would change (a median logit difference of \(+17\) to \(+25\) at the model's own message and of \(-17\) to \(-25\) at the User message), and since the reasoning is held fixed, the measure sees only the letter that follows it. At the model's own message, the slopes of different depths then differ in sign or stay near zero, so the share is defined only in some bands. In gpt-oss-120b, the slope at the User message lies mostly in the last third, far below what would change the answer; the layer windows there change the claim by at most 0.01. The slope along the part of the difference in the random subspace stays near zero throughout, as that part holds only 4--5\% of the norm of the difference (not of its square).

\begin{table}[htbp]
\centering
\caption{First-order sensitivity of the attribution answer by band of layers (150 items; 149 at the model's own message in gpt-oss-20b). Slope: the slope of the logit difference between the letters for ``you'' and ``user'' along the own--User difference, summed over the layers of the band and averaged over items. Share of the eight: the part of that slope carried by the eight coordinates, with a 95\% bootstrap interval over items, given where the summed slope is positive with its interval above zero (--: not defined). Size share: median share of the squared norm of the difference held by the eight coordinates. Installation band: the layers of the windows in which installation exceeds \(+0.10\) (Table~\ref{tab:self_mechanism_windows}). Every second layer in Qwen3.8-27B; with thinking, the reasoning is held fixed.}
\label{tab:self_fd}
\scriptsize
\setlength{\tabcolsep}{2.4pt}
\begin{tabular}{lccccc}
\toprule
& \multicolumn{2}{c}{User message} & \multicolumn{2}{c}{Model's own message} & \\
Band (layers) & Slope & Share of the eight & Slope & Share of the eight & Size share \\
\midrule
\multicolumn{6}{l}{\emph{Qwen3.8-27B, no thinking}} \\
\quad first third (L0--L20) & \(+3.94\) & \(0.14\) [\(-0.01\), \(0.25\)] & \(+2.00\) & \(-0.61\) [\(-3.03\), \(-0.01\)] & \(0.53\) \\
\quad middle third (L22--L42) & \(+22.03\) & \(0.33\) [\(0.30\), \(0.35\)] & \(+10.54\) & \(0.20\) [\(0.06\), \(0.31\)] & \(0.31\) \\
\quad last third (L44--L62) & \(+13.30\) & \(0.06\) [\(0.05\), \(0.07\)] & \(+11.22\) & \(0.05\) [\(0.03\), \(0.06\)] & \(0.39\) \\
\quad installation band (L24--L46) & \(+28.40\) & \(0.23\) [\(0.21\), \(0.26\)] & \(+15.41\) & \(0.13\) [\(0.04\), \(0.21\)] & \(0.31\) \\
\midrule
\multicolumn{6}{l}{\emph{Qwen3.8-27B, thinking}} \\
\quad first third (L0--L20) & \(+0.81\) & \(0.42\) [\(-0.90\), \(0.93\)] & \(+8.17\) & \(0.78\) [\(0.72\), \(0.82\)] & \(0.53\) \\
\quad middle third (L22--L42) & \(+4.75\) & \(0.47\) [\(0.40\), \(0.55\)] & \(-3.33\) & -- & \(0.33\) \\
\quad last third (L44--L62) & \(+2.55\) & \(0.45\) [\(0.38\), \(0.53\)] & \(-1.02\) & -- & \(0.40\) \\
\quad installation band (L28--L42) & \(+3.43\) & \(0.53\) [\(0.45\), \(0.61\)] & \(-3.38\) & -- & \(0.33\) \\
\midrule
\multicolumn{6}{l}{\emph{gpt-oss-20b, thinking}} \\
\quad first third (L0--L7) & \(+1.34\) & \(0.81\) [\(0.61\), \(1.06\)] & \(+2.38\) & \(0.91\) [\(0.76\), \(1.08\)] & \(0.55\) \\
\quad middle third (L8--L15) & \(+3.35\) & \(0.50\) [\(0.43\), \(0.59\)] & \(+1.86\) & \(0.09\) [\(-0.51\), \(0.34\)] & \(0.41\) \\
\quad last third (L16--L23) & \(+2.58\) & \(0.50\) [\(0.45\), \(0.56\)] & \(+0.44\) & -- & \(0.56\) \\
\quad installation band (L0--L15) & \(+4.69\) & \(0.59\) [\(0.51\), \(0.67\)] & \(+4.24\) & \(0.55\) [\(0.38\), \(0.67\)] & \(0.49\) \\
\midrule
\multicolumn{6}{l}{\emph{gpt-oss-120b, thinking}} \\
\quad first third (L0--L11) & \(+3.88\) & \(0.27\) [\(-0.16\), \(0.71\)] & \(-0.74\) & -- & \(0.42\) \\
\quad middle third (L12--L23) & \(+0.62\) & -- & \(-5.50\) & -- & \(0.35\) \\
\quad last third (L24--L35) & \(+7.99\) & \(0.64\) [\(0.52\), \(0.76\)] & \(+2.10\) & -- & \(0.41\) \\
\quad installation band (L8--L23) & \(+2.46\) & -- & \(-6.88\) & -- & \(0.34\) \\
\bottomrule
\end{tabular}
\end{table}

\FloatBarrier
\subsection{Random Subspaces}
\label{app:mech_random}

Table~\ref{tab:self_seeds} gives the random control of Section~\ref{sec:res_mechanism}, at every layer and position, for five random subspaces: the seed of Table~\ref{tab:self_mechanism} and the four that follow. Table~\ref{tab:self_seeds_scaled} gives the stricter version, in which the random edit of each seed is scaled up, item by item and layer by layer, to at least the size of the fitted edit of the same item, measured as the summed squared change over the edited positions relative to the summed squared state; its direction and positions are those of the random edit, and it is never scaled down. At every layer of every item, the scaled random edit is at least as large as the fitted one (smallest ratio 1.000). All conditions of the two tables share the unedited messages of one run.

\begin{table}[htbp]
\centering
\caption{Random subspaces of five seeds: the replacement of Table~\ref{tab:self_mechanism} in a random 8-dimensional subspace at every layer and position, item-paired change in ``you'' from the unedited message of Table~\ref{tab:self_question}, with a 95\% bootstrap interval over items (150 items per condition). The first seed draws the random subspace of Table~\ref{tab:self_mechanism}; the runs are those of Table~\ref{tab:self_question}, so with thinking its changes differ from Table~\ref{tab:self_mechanism} by up to 0.05.}
\label{tab:self_seeds}
\scriptsize
\setlength{\tabcolsep}{1.6pt}
\begin{tabular}{lcccc}
\toprule
& Qwen3.8-27B & Qwen3.8-27B & gpt-oss-20b & gpt-oss-120b \\
Seed & no thinking & thinking & thinking & thinking \\
\midrule
\multicolumn{5}{l}{own \(\leftarrow\) User, random} \\
\quad 20261022 & \(-0.01\) [\(-0.02\), \(-0.01\)] & \(-0.03\) [\(-0.08\), \(+0.01\)] & \(+0.05\) [\(-0.01\), \(+0.10\)] & \(+0.01\) [\(-0.01\), \(+0.03\)] \\
\quad 20261023 & \(+0.005\) [\(+0.002\), \(+0.008\)] & \(-0.05\) [\(-0.10\), \(-0.01\)] & \(0.00\) [\(-0.05\), \(+0.05\)] & \(-0.02\) [\(-0.05\), \(+0.01\)] \\
\quad 20261024 & \(+0.001\) [\(-0.002\), \(+0.004\)] & \(-0.05\) [\(-0.10\), \(-0.01\)] & \(-0.03\) [\(-0.09\), \(+0.02\)] & \(0.00\) [\(-0.02\), \(+0.02\)] \\
\quad 20261025 & \(-0.02\) [\(-0.02\), \(-0.01\)] & \(-0.07\) [\(-0.11\), \(-0.02\)] & \(-0.01\) [\(-0.07\), \(+0.04\)] & \(-0.03\) [\(-0.06\), \(0.00\)] \\
\quad 20261026 & \(+0.01\) [\(+0.01\), \(+0.02\)] & \(-0.03\) [\(-0.06\), \(+0.01\)] & \(-0.03\) [\(-0.09\), \(+0.03\)] & \(-0.05\) [\(-0.09\), \(-0.01\)] \\
\midrule
\multicolumn{5}{l}{User \(\leftarrow\) own, random} \\
\quad 20261022 & \(+0.004\) [\(+0.003\), \(+0.005\)] & \(0.00\) [\(0.00\), \(0.00\)] & \(0.00\) [\(0.00\), \(0.00\)] & \(0.00\) [\(0.00\), \(0.00\)] \\
\quad 20261023 & \(+0.003\) [\(+0.002\), \(+0.004\)] & \(0.00\) [\(0.00\), \(0.00\)] & \(0.00\) [\(0.00\), \(0.00\)] & \(+0.01\) [\(0.00\), \(+0.02\)] \\
\quad 20261024 & \(+0.002\) [\(+0.001\), \(+0.003\)] & \(0.00\) [\(0.00\), \(0.00\)] & \(0.00\) [\(0.00\), \(0.00\)] & \(0.00\) [\(0.00\), \(0.00\)] \\
\quad 20261025 & \(0.000\) [\(-0.001\), \(+0.001\)] & \(0.00\) [\(0.00\), \(0.00\)] & \(0.00\) [\(0.00\), \(0.00\)] & \(0.00\) [\(0.00\), \(0.00\)] \\
\quad 20261026 & \(+0.002\) [\(+0.001\), \(+0.003\)] & \(0.00\) [\(0.00\), \(0.00\)] & \(0.00\) [\(0.00\), \(0.00\)] & \(0.00\) [\(0.00\), \(0.00\)] \\
\bottomrule
\end{tabular}
\end{table}

\begin{table}[htbp]
\centering
\caption{Random subspaces of the five seeds of Table~\ref{tab:self_seeds}, with the random edit scaled up at every item and layer to at least the size of the fitted edit: item-paired change in ``you'' from the unedited message of Table~\ref{tab:self_question}, with a 95\% bootstrap interval over items (150 items per condition).}
\label{tab:self_seeds_scaled}
\scriptsize
\setlength{\tabcolsep}{1.6pt}
\begin{tabular}{lcccc}
\toprule
& Qwen3.8-27B & Qwen3.8-27B & gpt-oss-20b & gpt-oss-120b \\
Seed & no thinking & thinking & thinking & thinking \\
\midrule
\multicolumn{5}{l}{own \(\leftarrow\) User, random, scaled} \\
\quad 20261022 & \(-0.01\) [\(-0.02\), \(-0.01\)] & \(-0.13\) [\(-0.19\), \(-0.07\)] & \(-0.13\) [\(-0.20\), \(-0.05\)] & \(0.00\) [\(-0.03\), \(+0.03\)] \\
\quad 20261023 & \(+0.01\) [\(0.00\), \(+0.02\)] & \(-0.10\) [\(-0.15\), \(-0.05\)] & \(+0.05\) [\(-0.03\), \(+0.12\)] & \(-0.01\) [\(-0.03\), \(+0.02\)] \\
\quad 20261024 & \(-0.004\) [\(-0.009\), \(+0.001\)] & \(-0.08\) [\(-0.12\), \(-0.03\)] & \(+0.06\) [\(-0.01\), \(+0.13\)] & \(+0.01\) [\(0.00\), \(+0.03\)] \\
\quad 20261025 & \(-0.02\) [\(-0.03\), \(-0.01\)] & \(-0.08\) [\(-0.14\), \(-0.03\)] & \(-0.09\) [\(-0.15\), \(-0.05\)] & \(-0.05\) [\(-0.09\), \(-0.01\)] \\
\quad 20261026 & \(+0.02\) [\(+0.02\), \(+0.03\)] & \(-0.03\) [\(-0.07\), \(+0.01\)] & \(-0.15\) [\(-0.21\), \(-0.09\)] & \(-0.05\) [\(-0.09\), \(-0.01\)] \\
\midrule
\multicolumn{5}{l}{User \(\leftarrow\) own, random, scaled} \\
\quad 20261022 & \(+0.02\) [\(+0.02\), \(+0.03\)] & \(0.00\) [\(0.00\), \(0.00\)] & \(+0.01\) [\(0.00\), \(+0.02\)] & \(0.00\) [\(0.00\), \(0.00\)] \\
\quad 20261023 & \(+0.02\) [\(+0.01\), \(+0.02\)] & \(0.00\) [\(0.00\), \(0.00\)] & \(0.00\) [\(0.00\), \(0.00\)] & \(0.00\) [\(0.00\), \(0.00\)] \\
\quad 20261024 & \(+0.01\) [\(+0.01\), \(+0.01\)] & \(0.00\) [\(0.00\), \(0.00\)] & \(0.00\) [\(0.00\), \(0.00\)] & \(0.00\) [\(0.00\), \(0.00\)] \\
\quad 20261025 & \(+0.01\) [\(+0.01\), \(+0.01\)] & \(0.00\) [\(0.00\), \(0.00\)] & \(0.00\) [\(0.00\), \(0.00\)] & \(0.00\) [\(0.00\), \(0.00\)] \\
\quad 20261026 & \(+0.01\) [\(+0.01\), \(+0.01\)] & \(0.00\) [\(0.00\), \(0.00\)] & \(+0.01\) [\(0.00\), \(+0.02\)] & \(0.00\) [\(0.00\), \(0.00\)] \\
\bottomrule
\end{tabular}
\end{table}

\FloatBarrier
\subsection{Exchanging the Coordinates of User and System}
\label{app:mech_exchange}

Table~\ref{tab:self_exchange_full} gives the exchange of Section~\ref{sec:res_mechanism} in full: a User message receives the coordinates of the same text as a System message (User \(\leftarrow\) System), and a System message those of the same text as a User message (System \(\leftarrow\) User), with the same exchange in the random subspace of Table~\ref{tab:self_mechanism}; the subspace is the one fitted with the question, edited at every layer and position, and all conditions share the unedited messages of one run. The gpt-oss template renders a System message as a developer message. The phrase ``developer message'' appears in 108 of the 150 reasoning traces of gpt-oss-20b for the unedited System message and in all 150 of gpt-oss-120b; after the exchange, in 8 and 7; after the random exchange, in 121 and 149. For the System message with the model's own coordinates (Section~\ref{sec:res_mechanism}) it appears in 147 and 150.

\begin{table}[htbp]
\centering
\caption{The exchange of the User and System coordinates, attribution question, eight dimensions at every layer (150 items per condition). Unedited rows: probability without thinking, share of finished items with thinking, of the answers ``you'', ``user'', and ``system'', in this order. Other rows: item-paired change from the unedited message of the same role, with a 95\% bootstrap interval over items, in ``you'' and in the answer that names the role supplying the coordinates (``system'' for User \(\leftarrow\) System, ``user'' for System \(\leftarrow\) User). Random: the same exchange in the random subspace of Table~\ref{tab:self_mechanism}.}
\label{tab:self_exchange_full}
\scriptsize
\setlength{\tabcolsep}{1.6pt}
\begin{tabular}{lcccc}
\toprule
& Qwen3.8-27B & Qwen3.8-27B & gpt-oss-20b & gpt-oss-120b \\
& no thinking & thinking & thinking & thinking \\
\midrule
User, unedited & 0.04 / 0.91 / 0.04 & 0.00 / 0.99 / 0.01 & 0.00 / 1.00 / 0.00 & 0.00 / 1.00 / 0.00 \\
System, unedited & 0.07 / 0.71 / 0.21 & 0.00 / 0.37 / 0.63 & 0.00 / 0.10 / 0.90 & 0.01 / 0.11 / 0.89 \\
\midrule
\multicolumn{5}{l}{Change in ``you''} \\
\quad User \(\leftarrow\) System & \(+0.10\) [\(+0.09\), \(+0.12\)] & \(0.00\) [\(0.00\), \(0.00\)] & \(+0.01\) [\(0.00\), \(+0.02\)] & \(0.00\) [\(0.00\), \(0.00\)] \\
\quad User \(\leftarrow\) System, random & \(+0.001\) [\(0.000\), \(+0.002\)] & \(0.00\) [\(0.00\), \(0.00\)] & \(0.00\) [\(0.00\), \(0.00\)] & \(0.00\) [\(0.00\), \(0.00\)] \\
\quad System \(\leftarrow\) User & \(-0.03\) [\(-0.04\), \(-0.03\)] & \(0.00\) [\(0.00\), \(0.00\)] & \(+0.01\) [\(0.00\), \(+0.02\)] & \(-0.01\) [\(-0.02\), \(0.00\)] \\
\quad System \(\leftarrow\) User, random & \(+0.001\) [\(-0.001\), \(+0.003\)] & \(0.00\) [\(0.00\), \(0.00\)] & \(0.00\) [\(0.00\), \(0.00\)] & \(0.00\) [\(-0.02\), \(+0.02\)] \\
\midrule
\multicolumn{5}{l}{Change in ``system'' (User \(\leftarrow\) System) or ``user'' (System \(\leftarrow\) User)} \\
\quad User \(\leftarrow\) System & \(+0.07\) [\(+0.06\), \(+0.08\)] & \(+0.16\) [\(+0.10\), \(+0.23\)] & \(+0.01\) [\(0.00\), \(+0.02\)] & \(0.00\) [\(0.00\), \(0.00\)] \\
\quad User \(\leftarrow\) System, random & \(0.000\) [\(-0.001\), \(+0.001\)] & \(-0.01\) [\(-0.03\), \(0.00\)] & \(0.00\) [\(0.00\), \(0.00\)] & \(0.00\) [\(0.00\), \(0.00\)] \\
\quad System \(\leftarrow\) User & \(+0.16\) [\(+0.14\), \(+0.18\)] & \(+0.23\) [\(+0.13\), \(+0.34\)] & \(+0.88\) [\(+0.82\), \(+0.94\)] & \(+0.89\) [\(+0.84\), \(+0.93\)] \\
\quad System \(\leftarrow\) User, random & \(+0.02\) [\(+0.02\), \(+0.02\)] & \(+0.03\) [\(-0.01\), \(+0.07\)] & \(-0.01\) [\(-0.07\), \(+0.06\)] & \(0.00\) [\(-0.06\), \(+0.06\)] \\
\bottomrule
\end{tabular}
\end{table}

\FloatBarrier
\subsection{The Inverted Replacement}
\label{app:mech_rest}

Table~\ref{tab:self_rest} gives the inverted replacement of Section~\ref{sec:res_mechanism}: the recipient keeps its eight coordinates and takes the donor's state in all other directions, \(x\leftarrow y+(xB_\ell-yB_\ell)B_\ell^{\top}\), against the replacement of the eight coordinates, \(x\leftarrow x+(yB_\ell-xB_\ell)B_\ell^{\top}\), the whole donor state, \(x\leftarrow y\), and the random subspace of Table~\ref{tab:self_mechanism}, all with the subspace fitted with the question, at every layer and edited position. All conditions share the unedited messages of one run. In Qwen3.8-27B with thinking, 142 of the 150 items finish under the inverted replacement of own \(\leftarrow\) User; counting the unfinished items as not ``you'' or as ``you'' gives changes of \(-0.63\) and \(-0.58\).

\begin{table}[htbp]
\centering
\caption{The inverted replacement, attribution question, eight dimensions at every layer (150 items per condition). Unedited rows: probability of ``you'' without thinking, share of finished items answered ``you'' with thinking (in parentheses: finished of all, where not all finished). Other rows: item-paired change from the unedited message of the same role, with a 95\% bootstrap interval over items. The eight coordinates: the replacement of Table~\ref{tab:self_mechanism}; all but the eight coordinates: the recipient keeps its coordinates in the subspace and takes the donor's state outside it; the whole state: the donor's state; random subspace: the replacement in the random subspace of Table~\ref{tab:self_mechanism}.}
\label{tab:self_rest}
\scriptsize
\setlength{\tabcolsep}{1.6pt}
\begin{tabular}{lcccc}
\toprule
& Qwen3.8-27B & Qwen3.8-27B & gpt-oss-20b & gpt-oss-120b \\
Recipient \(\leftarrow\) donor & no thinking & thinking & thinking & thinking \\
\midrule
own, unedited & 0.80 & 0.97 (149/150) & 0.77 & 0.99 \\
\multicolumn{5}{l}{own \(\leftarrow\) User} \\
\quad the eight coordinates & \(-0.76\) [\(-0.77\), \(-0.75\)] & \(-0.96\) [\(-0.99\), \(-0.93\)] & \(-0.76\) [\(-0.83\), \(-0.69\)] & \(-0.96\) [\(-0.99\), \(-0.93\)] \\
\quad all but the eight coordinates & \(-0.77\) [\(-0.79\), \(-0.76\)] & \(-0.62\) [\(-0.70\), \(-0.53\)] & \(-0.14\) [\(-0.23\), \(-0.05\)] & \(-0.45\) [\(-0.53\), \(-0.36\)] \\
\quad the whole state & \(-0.76\) [\(-0.78\), \(-0.75\)] & \(-0.97\) [\(-0.99\), \(-0.95\)] & \(-0.77\) [\(-0.83\), \(-0.70\)] & \(-0.98\) [\(-1.00\), \(-0.95\)] \\
\quad random subspace & \(-0.01\) [\(-0.02\), \(-0.01\)] & \(-0.03\) [\(-0.08\), \(+0.01\)] & \(+0.05\) [\(-0.01\), \(+0.10\)] & \(+0.01\) [\(-0.01\), \(+0.03\)] \\
\midrule
User, unedited & 0.04 & 0.00 & 0.00 (149/150) & 0.00 \\
\multicolumn{5}{l}{User \(\leftarrow\) own} \\
\quad the eight coordinates & \(+0.50\) [\(+0.46\), \(+0.53\)] & \(+0.69\) [\(+0.62\), \(+0.77\)] & \(+0.46\) [\(+0.38\), \(+0.54\)] & \(+0.51\) [\(+0.43\), \(+0.59\)] \\
\quad all but the eight coordinates & \(+0.79\) [\(+0.77\), \(+0.81\)] & \(+0.03\) [\(+0.01\), \(+0.05\)] & \(+0.05\) [\(+0.01\), \(+0.08\)] & \(+0.01\) [\(0.00\), \(+0.03\)] \\
\quad the whole state & \(+0.82\) [\(+0.80\), \(+0.84\)] & \(+0.97\) [\(+0.95\), \(+0.99\)] & \(+0.74\) [\(+0.66\), \(+0.81\)] & \(+0.88\) [\(+0.83\), \(+0.93\)] \\
\quad random subspace & \(+0.004\) [\(+0.003\), \(+0.005\)] & \(0.00\) [\(0.00\), \(0.00\)] & \(0.00\) [\(0.00\), \(0.00\)] & \(0.00\) [\(0.00\), \(0.00\)] \\
\bottomrule
\end{tabular}
\end{table}

\FloatBarrier
\subsection{Wording Families and Unfinished Outputs}
\label{app:mech_cluster}

The intervals of Table~\ref{tab:self_mechanism} resample items. The 150 items of the attribution question are the 75 held-out bodies in two option orders, and the bodies form 15 wording families of five variants that differ only in their numbers, so the items of one family are not independent. Table~\ref{tab:self_cluster} gives the changes of the main condition with intervals that resample the 15 families with replacement and keep every item of a drawn family, with its option orders and its paired conditions (4,000 resamples, fixed seed). With thinking, each change is computed over the items finished in both conditions. Over all 150 items, with every unfinished output treated as unknown, the bounds of each change lie within 0.03 of it, except for the System message with the model's own coordinates in gpt-oss-20b: 125 edited and 116 unedited outputs finish, 103 items in both, and the bounds are \(-0.22\) and \(+0.17\) around a change of \(+0.01\). The multiplication is answered as unedited in every condition and readout, with family intervals of \([0.00, 0.00]\).

\begin{table}[htbp]
\centering
\caption{The causal tests with intervals over wording families, eight dimensions at every layer and position. Item-paired change from the unedited message of the same role (attribution: ``you'', probability without thinking and share of finished items with thinking, 150 items; content question: share answered correctly, 75 items), with a 95\% bootstrap interval over the 15 held-out wording families. Except for own \(\leftarrow\) own, which copies the unedited coordinates, the attribution changes are those of Table~\ref{tab:self_mechanism}, whose intervals resample items.}
\label{tab:self_cluster}
\scriptsize
\setlength{\tabcolsep}{1.6pt}
\begin{tabular}{lcccc}
\toprule
& Qwen3.8-27B & Qwen3.8-27B & gpt-oss-20b & gpt-oss-120b \\
Recipient \(\leftarrow\) donor & no thinking & thinking & thinking & thinking \\
\midrule
\multicolumn{5}{l}{\emph{Attribution: change in ``you''}} \\
own \(\leftarrow\) User & \(-0.76\) [\(-0.78\), \(-0.73\)] & \(-0.95\) [\(-0.99\), \(-0.88\)] & \(-0.75\) [\(-0.92\), \(-0.57\)] & \(-0.92\) [\(-0.99\), \(-0.83\)] \\
own \(\leftarrow\) System & \(-0.64\) [\(-0.67\), \(-0.60\)] & \(-0.86\) [\(-0.95\), \(-0.77\)] & \(-0.67\) [\(-0.83\), \(-0.49\)] & \(-0.45\) [\(-0.62\), \(-0.28\)] \\
own \(\leftarrow\) User, random & \(-0.01\) [\(-0.02\), \(-0.01\)] & \(+0.02\) [\(-0.03\), \(+0.09\)] & \(+0.05\) [\(-0.05\), \(+0.17\)] & \(-0.01\) [\(-0.05\), \(+0.01\)] \\
own \(\leftarrow\) own & \(0.00\) [\(0.00\), \(0.00\)] & \(0.00\) [\(-0.05\), \(+0.05\)] & \(0.00\) [\(-0.03\), \(+0.03\)] & \(0.00\) [\(-0.02\), \(+0.02\)] \\
User \(\leftarrow\) own & \(+0.49\) [\(+0.40\), \(+0.58\)] & \(+0.60\) [\(+0.42\), \(+0.77\)] & \(+0.51\) [\(+0.29\), \(+0.72\)] & \(+0.39\) [\(+0.21\), \(+0.59\)] \\
User \(\leftarrow\) own, random & \(+0.004\) [\(+0.001\), \(+0.006\)] & \(0.00\) [\(0.00\), \(0.00\)] & \(-0.01\) [\(-0.02\), \(0.00\)] & \(0.00\) [\(0.00\), \(0.00\)] \\
System \(\leftarrow\) own & \(+0.39\) [\(+0.32\), \(+0.45\)] & \(+0.06\) [\(0.00\), \(+0.14\)] & \(+0.01\) [\(0.00\), \(+0.03\)] & \(+0.01\) [\(0.00\), \(+0.02\)] \\
\midrule
\multicolumn{5}{l}{\emph{Content question: change in the share answered correctly}} \\
own \(\leftarrow\) User & \(0.00\) [\(0.00\), \(0.00\)] & \(0.00\) [\(0.00\), \(0.00\)] & \(-0.03\) [\(-0.08\), \(0.00\)] & \(0.00\) [\(0.00\), \(0.00\)] \\
own \(\leftarrow\) System & \(0.00\) [\(0.00\), \(0.00\)] & \(0.00\) [\(0.00\), \(0.00\)] & \(0.00\) [\(0.00\), \(0.00\)] & \(0.00\) [\(0.00\), \(0.00\)] \\
own \(\leftarrow\) User, random & \(0.00\) [\(0.00\), \(0.00\)] & \(0.00\) [\(0.00\), \(0.00\)] & \(0.00\) [\(0.00\), \(0.00\)] & \(0.00\) [\(0.00\), \(0.00\)] \\
User \(\leftarrow\) own & \(+0.03\) [\(0.00\), \(+0.08\)] & \(0.00\) [\(0.00\), \(0.00\)] & \(+0.04\) [\(0.00\), \(+0.12\)] & \(0.00\) [\(0.00\), \(0.00\)] \\
\bottomrule
\end{tabular}
\end{table}

\FloatBarrier
\subsection{Without the Question, and Stricter Tasks}
\label{app:mech_question}

Table~\ref{tab:self_question} gives the edit of Section~\ref{sec:abl_question} in the four readouts, two subspaces crossed with two sets of edited positions. The subspace of Section~\ref{sec:res_mechanism} is fitted from the first body token to the second-to-last prompt token, which includes the question; the message subspace from the first body token to the last token before the question. Both are fitted on the 75 discovery bodies in both option orders, at every layer; the mean squared cosine of the principal angles between their 8-dimensional spaces has a median of 0.77--0.88 over layers. The edit reaches every position, as in Section~\ref{sec:res_mechanism}, or only the message, the body and the end of the message. All conditions of the table share the unedited messages of one run.

\begin{table}[htbp]
\centering
\caption{Attribution without the question, eight dimensions at every layer (150 items per condition). Unedited rows: probability of ``you'' without thinking, share of finished items answered ``you'' with thinking (in parentheses: finished of all, where not all finished). Other rows: item-paired change from the unedited message of the same role, with a 95\% bootstrap interval over items. Subspace fitted with the question (as in Section~\ref{sec:res_mechanism}) or on the message alone; edited positions: all (as in Section~\ref{sec:res_mechanism}) or the message alone (body and end of the message).}
\label{tab:self_question}
\scriptsize
\setlength{\tabcolsep}{1.6pt}
\begin{tabular}{lcccc}
\toprule
& Qwen3.8-27B & Qwen3.8-27B & gpt-oss-20b & gpt-oss-120b \\
Subspace, edited positions & no thinking & thinking & thinking & thinking \\
\midrule
own, unedited & 0.80 & 0.97 (149/150) & 0.77 & 0.99 \\
\multicolumn{5}{l}{own \(\leftarrow\) User} \\
\quad with question, all & \(-0.76\) [\(-0.77\), \(-0.75\)] & \(-0.96\) [\(-0.99\), \(-0.93\)] & \(-0.76\) [\(-0.83\), \(-0.69\)] & \(-0.96\) [\(-0.99\), \(-0.93\)] \\
\quad with question, message & \(-0.76\) [\(-0.77\), \(-0.75\)] & \(-0.96\) [\(-0.99\), \(-0.93\)] & \(-0.77\) [\(-0.83\), \(-0.70\)] & \(-0.99\) [\(-1.00\), \(-0.97\)] \\
\quad message, all & \(-0.73\) [\(-0.74\), \(-0.72\)] & \(-0.96\) [\(-0.99\), \(-0.93\)] & \(-0.77\) [\(-0.83\), \(-0.70\)] & \(-0.95\) [\(-0.98\), \(-0.91\)] \\
\quad message, message & \(-0.74\) [\(-0.75\), \(-0.73\)] & \(-0.93\) [\(-0.97\), \(-0.89\)] & \(-0.76\) [\(-0.83\), \(-0.69\)] & \(-0.96\) [\(-0.99\), \(-0.93\)] \\
\midrule
User, unedited & 0.04 & 0.00 & 0.00 (149/150) & 0.00 \\
\multicolumn{5}{l}{User \(\leftarrow\) own} \\
\quad with question, all & \(+0.50\) [\(+0.46\), \(+0.53\)] & \(+0.69\) [\(+0.62\), \(+0.77\)] & \(+0.46\) [\(+0.38\), \(+0.54\)] & \(+0.51\) [\(+0.43\), \(+0.59\)] \\
\quad with question, message & \(+0.33\) [\(+0.30\), \(+0.37\)] & \(+0.49\) [\(+0.41\), \(+0.57\)] & \(+0.34\) [\(+0.27\), \(+0.42\)] & \(+0.33\) [\(+0.26\), \(+0.41\)] \\
\quad message, all & \(+0.42\) [\(+0.38\), \(+0.46\)] & \(+0.63\) [\(+0.55\), \(+0.71\)] & \(+0.39\) [\(+0.32\), \(+0.47\)] & \(+0.26\) [\(+0.19\), \(+0.33\)] \\
\quad message, message & \(+0.38\) [\(+0.34\), \(+0.42\)] & \(+0.65\) [\(+0.58\), \(+0.73\)] & \(+0.32\) [\(+0.24\), \(+0.39\)] & \(+0.11\) [\(+0.07\), \(+0.17\)] \\
\bottomrule
\end{tabular}
\end{table}

Table~\ref{tab:self_strict} gives the stricter versions of the two self-related tasks of Section~\ref{sec:abl_question}, both with the message subspace and the edit of the message alone, at every layer. The attribution question in the held-out wording reads ``Who is the author of the message above? A.~You (the AI assistant) B.~The user C.~The system. Answer with the letter only.'', in the two option orders of the causal tests; the Chinese wording will be released with the code. In the secret game, the edit reaches X, its body and the template tokens after it up to the first token of the last message; the last message and the generation prompt are left unedited. The games are those of Table~\ref{tab:self_game}. Each test has its own unedited runs.

\begin{table}[htbp]
\centering
\caption{Stricter versions of the two self-related tasks, with the subspace fitted on the message alone and only the message edited, eight dimensions at every layer. Attribution in a held-out wording (150 items per condition): unedited rows give the probability of ``you'' without thinking and the share of finished items answered ``you'' with thinking. Secret game with X alone edited (100 games per condition): unedited rows give the share of finished games in which the model acts as the setter (in parentheses: finished of all, where not all finished). Other rows: item-paired change from the unedited message of the same role, with a 95\% bootstrap interval over items. Control: share of the games in which the control question is answered correctly, lowest and highest over all conditions.}
\label{tab:self_strict}
\scriptsize
\setlength{\tabcolsep}{1.6pt}
\begin{tabular}{lcccc}
\toprule
& Qwen3.8-27B & Qwen3.8-27B & gpt-oss-20b & gpt-oss-120b \\
& no thinking & thinking & thinking & thinking \\
\midrule
\multicolumn{5}{l}{\emph{Attribution question, held-out wording}} \\
own, unedited & 0.48 & 0.63 & 0.71 & 0.99 \\
User, unedited & 0.01 & 0.00 & 0.00 & 0.00 \\
\quad own \(\leftarrow\) User & \(-0.46\) [\(-0.49\), \(-0.43\)] & \(-0.63\) [\(-0.70\), \(-0.55\)] & \(-0.71\) [\(-0.78\), \(-0.63\)] & \(-0.99\) [\(-1.00\), \(-0.97\)] \\
\quad User \(\leftarrow\) own & \(+0.12\) [\(+0.11\), \(+0.14\)] & \(+0.33\) [\(+0.25\), \(+0.41\)] & \(+0.09\) [\(+0.05\), \(+0.13\)] & \(0.00\) [\(0.00\), \(0.00\)] \\
\quad own \(\leftarrow\) User, random & \(-0.02\) [\(-0.03\), \(-0.01\)] & \(+0.03\) [\(-0.03\), \(+0.09\)] & \(0.00\) [\(-0.04\), \(+0.04\)] & \(+0.01\) [\(-0.01\), \(+0.03\)] \\
\midrule
\multicolumn{5}{l}{\emph{Secret game, X alone edited}} \\
own X, unedited & 1.00 & 1.00 & 1.00 & 1.00 \\
User X, unedited & 0.00 & 0.00 (94/100) & 0.00 & 0.00 \\
\quad own \(\leftarrow\) User & \(0.00\) [\(0.00\), \(0.00\)] & \(-0.04\) [\(-0.08\), \(-0.01\)] & \(-0.27\) [\(-0.36\), \(-0.19\)] & \(-1.00\) [\(-1.00\), \(-1.00\)] \\
\quad User \(\leftarrow\) own & \(+0.26\) [\(+0.18\), \(+0.35\)] & \(0.00\) [\(0.00\), \(0.00\)] & \(+0.59\) [\(+0.48\), \(+0.69\)] & \(+0.03\) [\(0.00\), \(+0.07\)] \\
\quad own \(\leftarrow\) User, random & \(0.00\) [\(0.00\), \(0.00\)] & \(0.00\) [\(0.00\), \(0.00\)] & \(0.00\) [\(0.00\), \(0.00\)] & \(0.00\) [\(0.00\), \(0.00\)] \\
control correct & 1.00 & 1.00 & 0.98--1.00 & 1.00 \\
\bottomrule
\end{tabular}
\end{table}

\FloatBarrier
\subsection{The Replacement under the Format Ablation}
\label{app:mech_ablation}

Table~\ref{tab:self_ablation_2x2} repeats the replacement of Section~\ref{sec:res_mechanism} with the formatting specific to the Assistant role removed (Section~\ref{sec:design_ablation}), each subspace fitted under the same ablation. Where the unedited claim remains clear, User coordinates remove it from the model's own message, and random subspaces change it far less; in Qwen3.8-27B with thinking, the random control lowers it by 0.10. Without thinking, Qwen3.8-27B claims its own messages with a probability of only 0.10 under the ablation, too little for the replacement to test. Copying the unedited coordinates is exact without thinking: every item gives the same probability of ``you'' as unedited, in the main runs and under the ablation. With thinking, the copy changes the share by 0.00 in the main runs, but it alters the generated reasoning of 71--113 of the 150 items and changes individual answers in both directions (among the items finished in both runs, 5 and 5 in Qwen3.8-27B, where one more item finishes only after the copy; 6 and 6 in gpt-oss-20b; 1 and 1 in gpt-oss-120b); under the ablation in Qwen3.8-27B, 10 answers change away from ``you'' and 5 toward it, which gives the \(-0.03\) of Table~\ref{tab:self_ablation_2x2}. Effects of this size in the thinking readouts lie within this variation.

\begin{table}[htbp]
\centering
\caption{Attribution under the format ablation, replacement of eight self-index coordinates at every layer (150 items per condition): Qwen3.8-27B with the template switch, gpt-oss with the channel tag masked and later positions recomputed. Unedited rows: probability of ``you'' without thinking, share of finished items answered ``you'' with thinking; other rows: item-paired change from the unedited message of the same role, with a 95\% bootstrap interval over items. Copy: the unedited coordinates written back. --: not run.}
\label{tab:self_ablation_2x2}
\scriptsize
\setlength{\tabcolsep}{2.5pt}
\begin{tabular}{lcccc}
\toprule
& Qwen3.8-27B & Qwen3.8-27B & gpt-oss-20b & gpt-oss-120b \\
Recipient \(\leftarrow\) donor & no thinking & thinking & thinking & thinking \\
\midrule
own, unedited & 0.10 & 0.63 & 0.61 & 0.94 \\
own \(\leftarrow\) User & \(-0.08\) [\(-0.10\), \(-0.07\)] & \(-0.63\) [\(-0.70\), \(-0.55\)] & \(-0.61\) [\(-0.69\), \(-0.53\)] & \(-0.94\) [\(-0.97\), \(-0.90\)] \\
own \(\leftarrow\) System & \(+0.02\) [\(+0.01\), \(+0.02\)] & \(-0.59\) [\(-0.67\), \(-0.51\)] & -- & -- \\
own \(\leftarrow\) User, random & \(-0.01\) [\(-0.01\), \(-0.01\)] & \(-0.10\) [\(-0.18\), \(-0.03\)] & \(-0.01\) [\(-0.08\), \(+0.07\)] & \(+0.01\) [\(-0.03\), \(+0.05\)] \\
own \(\leftarrow\) own (copy) & \(0.00\) [\(0.00\), \(0.00\)] & \(-0.03\) [\(-0.09\), \(+0.01\)] & -- & -- \\
\midrule
User, unedited & 0.04 & 0.00 & 0.01 & 0.00 \\
User \(\leftarrow\) own & \(+0.08\) [\(+0.07\), \(+0.09\)] & \(+0.15\) [\(+0.10\), \(+0.21\)] & \(+0.20\) [\(+0.13\), \(+0.27\)] & \(+0.45\) [\(+0.37\), \(+0.53\)] \\
User \(\leftarrow\) own, random & \(0.00\) [\(0.00\), \(0.00\)] & \(0.00\) [\(0.00\), \(0.00\)] & \(-0.01\) [\(-0.02\), \(0.00\)] & \(0.00\) [\(0.00\), \(0.00\)] \\
\midrule
System, unedited & 0.07 & 0.00 & -- & -- \\
System \(\leftarrow\) own & \(-0.02\) [\(-0.02\), \(-0.01\)] & \(+0.02\) [\(0.00\), \(+0.05\)] & -- & -- \\
\bottomrule
\end{tabular}
\end{table}

\FloatBarrier
\subsection{Position Compression}
\label{app:mech_positions}

Table~\ref{tab:self_positions} gives the geometry of Section~\ref{sec:res_position} with and without position compression.

\begin{table}[htbp]
\centering
\caption{Geometry with and without position compression: layers at which the model's own message is singled out in at least 80\% of the held-out bodies. Compressed: the tokens specific to the Assistant role take the position of the token before them, so that later tokens have the positions of the User rendering.}
\label{tab:self_positions}
\footnotesize
\begin{tabular}{lcccc}
\toprule
& \multicolumn{2}{c}{Message content} & \multicolumn{2}{c}{Response onset} \\
\cmidrule(lr){2-3}\cmidrule(lr){4-5}
Model and rendering & default & compressed & default & compressed \\
\midrule
Qwen3.8-27B, non-thinking rendering & 64/64 & 64/64 & 64/64 & 64/64 \\
Qwen3.8-27B, thinking rendering & 64/64 & 64/64 & 61/64 & 61/64 \\
gpt-oss-20b & 20/24 & 18/24 & 24/24 & 20/24 \\
gpt-oss-120b & 25/36 & 21/36 & 33/36 & 25/36 \\
\bottomrule
\end{tabular}
\end{table}

\clearpage
\section{A Side Question: Trust in the Model's Own Records}
\label{app:ownstep}

As a side question, we asked whether a model that recognizes its own messages builds more readily on an intermediate step when the step appears as its own than when the user wrote it.

\paragraph{Task.} We generated 120 two-step arithmetic word problems, 24 from each of five templates (stock, books, students, water, vehicles), in Chinese and, for the models tested in English, in English. A conversation consists of a User message with the problem, a message with the first step (``Step 1: \(a-b=s\)''), and a User message asking for the final answer as digits only. The step message is either an Assistant message, a step the model wrote itself, or a User message, a step the user wrote; its text is identical. The step is either correct or wrong by 1--3, and problems are chosen so that the correct answer and the answer that propagates the wrong step differ in their first digit. We record greedy answers (six generated tokens without thinking; with thinking, the answer after the reasoning) and, as a continuous measure, the \emph{adoption index} \(P(d_{\mathrm{err}})/(P(d_{\mathrm{err}})+P(d_{\mathrm{cor}}))\), where \(d\) is the first digit of the error-propagating and of the correct answer at the response onset; the ten digits carry at least 0.96 of the probability mass in every condition.

In the reasoning task, the step on which the final answer builds has the same text in both roles, so a difference in how often a wrong step is adopted comes from its source alone. With a correct step, the source makes no difference: Qwen3.8-27B answers correctly in 0.97 and 0.98 of the problems without thinking (the adoption index differs by \(+0.009\), 95\% interval \(-0.007\) to \(0.030\), in the implementation of Sections~\ref{sec:res_index}--\ref{sec:res_thinking}), and every model with self-reference that reasons in its thinking mode answers all 120 problems correctly from either source; MiMo-V2.6-Distill-Qwen-9B, which answers in that mode without written reasoning, answers 0.68 of the problems correctly after its own correct step and 0.25 after the user's. With a wrong step, every model and mode with self-reference has a preference, but the preferences have no common direction (Table~\ref{tab:self_ownstep}). Qwen3.8-27B without thinking adopts its own wrong steps more often than the user's: its mean adoption index is higher by 0.14 (95\% bootstrap interval over problems 0.08--0.20) in the cross-model implementation (Table~\ref{tab:self_ownstep}) and by 0.110 (0.059--0.168) in the implementation of Sections~\ref{sec:res_index}--\ref{sec:res_thinking}, whose greedy answers propagate the error in 0.92 against 0.75 of the problems; and so do gpt-oss-20b and gpt-oss-120b with thinking (0.20 against 0.06, and 0.11 against 0.02 of finished items) and MiMo-V2.6-Distill-Qwen-9B in the default call of its template (0.62 against 0.20). Qwen3.8-27B with thinking, EXAONE-4.0.1-32B, Qwen3.8-Flash-Next, and GLM-5.3-Flash adopt the user's wrong steps more often, the last two by a wide margin (0.01 against 0.44, and 0.03 against 0.43). The direction can differ between the two modes of one model. The source of a record thus changes how a model builds on a wrong record, in a direction that differs between models and modes.

\begin{table}[htbp]
\centering
\caption{Wrong steps adopted when they appear in Assistant history (own) and when the same steps come from the user, in every model and mode with self-reference (120 problems). Without thinking: mean adoption index (this appendix), measured with the cross-model implementation (the implementation of Sections~\ref{sec:res_index}--\ref{sec:res_thinking} gives \(+0.110\), see the text); with thinking: share of finished items whose answer propagates the error. Own minus user: item-paired difference with a 95\% bootstrap interval over problems.}
\label{tab:self_ownstep}
\small
\begin{tabular}{llccc}
\toprule
Model & Mode & own & user & own minus user \\
\midrule
Qwen3.8-27B & no thinking & 0.96 & 0.82 & \(+0.14\) [0.08, 0.20] \\
Qwen3.8-27B & thinking & 0.01 & 0.04 & \(-0.03\) [\(-0.07\), \(-0.01\)] \\
gpt-oss-20b & thinking & 0.20 & 0.06 & \(+0.14\) [0.05, 0.23] \\
gpt-oss-120b & thinking & 0.11 & 0.02 & \(+0.09\) [0.03, 0.15] \\
EXAONE-4.0.1-32B & thinking & 0.03 & 0.08 & \(-0.05\) [\(-0.09\), \(-0.02\)] \\
GLM-5.3-Flash & thinking & 0.03 & 0.43 & \(-0.40\) [\(-0.49\), \(-0.32\)] \\
Qwen3.8-Flash-Next & thinking & 0.01 & 0.44 & \(-0.43\) [\(-0.53\), \(-0.35\)] \\
MiMo-V2.6-Distill-9B & default call & 0.62 & 0.20 & \(+0.42\) [0.33, 0.53] \\
\bottomrule
\end{tabular}
\end{table}

\end{document}